\pdfoutput=1
\documentclass[11pt]{article}

\usepackage{acl} % [review]

\usepackage{times}
\usepackage{latexsym}

\usepackage{mathptmx}
\usepackage{array}
\usepackage{color}
\usepackage{mathtools}
\usepackage{enumitem} 
\usepackage{amsmath}
\usepackage{amsthm}
\usepackage{amssymb}
\usepackage{multirow}
\usepackage{booktabs}
\usepackage{xcolor}
\usepackage{subcaption} 
\usepackage{pifont}

\usepackage{fancyhdr}
\usepackage{lipsum} % For generating dummy text

\fancypagestyle{firstpage}{
  \fancyhf{} % Clear all header and footer fields
  \fancyfoot[C]{%
    \parbox{\textwidth}{\centering
      Accepted by the 63rd Annual Meeting of the Association for Computational Linguistics (ACL 2025) 
    }%
  }
  \fancyfoot[C]{\thepage\vspace{0.5em}\\ % Page number above the conference info
    \parbox{\textwidth}{\centering
      Accepted by the 63rd Annual Meeting of the Association for Computational Linguistics (ACL 2025)
    }%
  }
}

\AtBeginDocument{\thispagestyle{firstpage}}

\newtheorem{theorem}{Theorem}

\newtheorem{corollary}[theorem]{Corollary}

\DeclareMathAlphabet{\mathcal}{OMS}{cmsy}{m}{n}

\usepackage[T1]{fontenc}
\usepackage[utf8]{inputenc}

\usepackage{microtype}

\usepackage{inconsolata}

\usepackage{graphicx}

\title{Principled Understanding of Generalization for Generative Transformer Models in Arithmetic Reasoning Tasks}

\author{
 \textbf{Xingcheng Xu\textsuperscript{1}},
 \textbf{Zibo Zhao\textsuperscript{2,3}},
 \textbf{Haipeng Zhang\textsuperscript{2}\thanks{Corresponding authors.}},
 \textbf{Yanqing Yang\textsuperscript{4}\footnotemark[1]}
\\
\\
 \textsuperscript{1}Shanghai Artificial Intelligence Laboratory,
 \textsuperscript{2}ShanghaiTech University\\
 \textsuperscript{3}University of Hong Kong,
 \textsuperscript{4}Fudan University
\\
\small{
   \href{mailto:xingcheng.xu18@gmail.com}{xingcheng.xu18@gmail.com}, 
   \href{mailto:andrewzhao054@gmail.com}{andrewzhao054@gmail.com}
}
\\
\small{
   \href{mailto:zhanghp@shanghaitech.edu.cn}{zhanghp@shanghaitech.edu.cn}, 
   \href{mailto:yanqingyang@fudan.edu.cn}{yanqingyang@fudan.edu.cn}
}
}

\begin{document}
\maketitle
\begin{abstract}
Transformer-based models excel in various tasks but their generalization capabilities, especially in arithmetic reasoning, remain incompletely understood. Arithmetic tasks provide a controlled framework to explore these capabilities, yet performance anomalies persist, such as inconsistent effectiveness in multiplication and erratic generalization in modular addition (e.g., modulo 100 vs. 101). This paper develops a unified theoretical framework for understanding the generalization behaviors of transformers in arithmetic tasks, focusing on length generalization. Through detailed analysis of addition, multiplication, and modular operations, we reveal that translation invariance in addition aligns with relative positional encoding for robust generalization, while base mismatch in modular operations disrupts this alignment. Experiments across GPT-family models validate our framework, confirming its ability to predict generalization behaviors. Our work highlights the importance of task structure and training data distribution for achieving data-efficient and structure-aware training, providing a systematic approach to understanding of length generalization in transformers.
\end{abstract}

\section{Introduction}

Since the introduction of Transformer~\citep{vaswani2017attention}, Transformer-based models including large language models (LLMs) and large multimodal models (LMMs) have experienced a rapid rise,  excel in a wide range of tasks, such as natural language processing, coding, mathematical reasoning, and vision understanding~\citep{bubeck2023sparks,lu2024gpt}. However, the generalization capabilities of these transformer based foundation models are not yet fully understood in areas such as natural language understanding~\citep{bender2021dangers} and mathematical reasoning~\citep{anil2022exploring,jelassi2023length}.

The generalization capabilities are often link to models' capability to generalize beyond their training data (out-of-distribution (OOD) generalization) in NLP tasks, which is much complex and challenging. LLMs perform exceptionally well on some generalization tasks while produce factual errors or misinformation on others e.g.,~\citep{bender2021dangers,lu2024gpt}. Studies therefore try to figure out why these differences exist between generalization tasks~\citep{briakou2023searching}, what LLMs are actually learning on failed ones~\citep{xu2023ood}, and how they manage to generalize on successful tasks~\citep{jelassi2023length,mcleish2024transformers}. 

Given the complexity of next-token prediction across diverse corpora and models' opacity, mathematical tasks (e.g., $n$-digit addition / multiplication / modular operations) serve as interpretable probes for generalization analysis. \textit{In this paper we introduce a unified theoretical framework for understanding generalization behavior of transformers on arithmetic tasks. By clarifying the downward and upward Out-of-Distribution (OOD) generalization and their requirement of task and training data}, we are able to connect the mysterious discrepancies in models' generalization capability found in the literature summarized as the following:

(1) certain tasks (e.g., addition) succeed in unseen generalization with certain positional encodings (e.g., relative) but not other tasks (e.g., multiplication), and (2) there is a significant generalization difference between very close moduli in modular operations (e.g., modulo 100 and 101). Specifically, previous studies have observed that when training models with absolute positional embeddings (APE) on \( n \)-digit operations (e.g., addition), where both input operands are no longer than \( n \)-digit in length such as \( 1234+5678 \) for \( n=4 \), the models successfully generalize on unseen \( n \)-digit inputs such as \( 4321+8765 \) (termed in-distribution (ID) generalization). However, they fail on longer unseen cases such as \( 91234+15678 \) (termed OOD generalization) as shown by \citet{anil2022exploring}, \citet{jelassi2023length}, \citet{lee2023teaching}, and \citet{xu2023ood}. Besides, models with relative positional embeddings (RPE) can generalize to longer unseen inputs for addition tasks but struggle with multiplication tasks, according to \citet{jelassi2023length} and \citet{mcleish2024transformers}. Additionally, models trained on modular operations with specific moduli such as 100 can perfectly generalize to any longer unseen inputs with either absolute or relative positional embeddings. However, they fail to generalize to longer unseen inputs for other very close moduli such as 101, as noted by \citet{jelassi2023length}. These OOD generalization mysteries are cataloged in Table~\ref{table:length_generalize_literature}.

\begin{table}[ht]
\centering
\resizebox{0.45\textwidth}{!}{%
\begin{tabular}{lccccc}
\toprule
\multicolumn{1}{c}{} & \multicolumn{1}{c}{\multirow{2}{*}{Addition}} & \multicolumn{1}{c}{\multirow{2}{*}{Multiplication}} & \multicolumn{2}{c}{Modular Operations} \\
\cmidrule(lr){4-5}
 &  &  & $p=100$ & $p=101$\\
\midrule
APE & \ding{55} & \ding{55} & \ding{51} & \ding{55} \\
RPE & \ding{51} & \ding{55} & \ding{51} & \ding{55} \\
\bottomrule
\end{tabular}}
\caption{Length Generalization of Transformers  with APE and RPE on Arithmetic Tasks}
\label{table:length_generalize_literature}
\end{table}

As we can summarize, these previous efforts address generalization issues in specific tasks, modifying components of individual models, such as altering positional encodings~\citep{jelassi2023length,mcleish2024transformers} or attention mechanisms~\citep{dubois2019location}. Their failure in figuring out the underneath mechanism calls for a reflective examination -- we believe the field has overlooked the differences in task properties (e.g., addition v.s. multiplication, modulo $10^2$ v.s. modulo $10^2+1)$ that may drive the difference in generalization property among tasks. The perspective of mechanistic interpretability~\citep{hernandez2022scaling,liu2022towards} offers an angle in this direction. This data-driven and experimentally-based analytical approach has helped identify and interpret phenomena such as "grokking"~\citep{liu2022towards} and analyze the impact of repeated data on the performance of LLMs~\citep{hernandez2022scaling}. 

%In this paper, we develop a unified theoretical framework that builds on the principles of language modeling, the model's ability for universal approximation, and detailed task property analysis in various arithmetic scenarios. 
In this paper, we present a unified theoretical framework integrating properties of autogressive transformer model, universal approximation capabilities, and task-specific property analysis across diverse arithmetic tasks.
It assumes that generalization behaviors depend on task properties once the model converges on the training data. For example, digital addition is translation invariant with a large probability, yielding consistent results despite digit shifts, aligning with RPE's preservation of positional relationships, unlike multiplication. This leads to well generalization of addition with a large probability to unseen longer domains under RPE but not for multiplication. The modulo (e.g. 100, 101) discrepancy stems from base alignment: modulo 100 matches base 10, discarding higher digits \( 11234+15678 \equiv 1234+5678 \equiv 34+78 \pmod{100} \), whereas modulo 101 requires them.

We then perform more extensive generalization analyzes assuming that transformer models are trained in \( n \)-digit operations with at least one operand having a length of \( n \) such as \( 1234+567 \) for \( n=4 \). This differs from the literature where the length of both operands is no longer than \( n \). We categorize generalization into two types: \emph{downward OOD generalization} and \emph{upward OOD generalization}. Downward OOD generalization\footnote{As a note, the downward OOD domain generalization is not trivial. If a model is trained on a smaller domain with a significant gap from the desired training dataset, such as training on \( n \)-digit addition with both operands having a length of \( n \) and the highest digits of both operands being greater than, for example, 5, the model fails to generalize to the downward OOD domain.} involves generalizing to downward domains, such as \( 120+235 \) or \( 11+32 \), while upward OOD generalization involves generalizing to upward domains, such as \( 12035+235 \) or \( 123456+323456 \). The core conclusions of our theoretical analysis are as follows: (1) For addition, under APE, Transformer models can generalize to the downward (downward) OOD domain, but not to the upward (upward) OOD domain. However, under RPE, the models can generalize to both downward and upward OOD domains, benefiting from the translation invariance of digit addition. (2) For multiplication, even RPE has limited effectiveness in the upward OOD domain due to the lack of translation invariance property. (3) For modular operations, if the modulus \(p\) divides \(10^n\), models can generalize to both downward and upward OOD domains regardless of the positional encoding, due to the compatibility with base 10 such that the information at higher-digit positions of the operands do not affect the result. When the modulus \(p\) does not divide \(10^n\), models can only generalize to the downward OOD domain. For upward OOD domains, we have derived a theoretical accuracy formula based on the information loss and identification of the model's final learned function.

The challenge in understand the generalization capacity of LLM has significant implications for LLM training, alignment, and application ~\citep{ji2023aialignment}. Our analysis highlights the importance of training data distribution. If the data excluded from the training dataset does not affect the desired ground truth support set, such as when the downward OOD domain is excluded during training, the model can still learn to generalize to the excluded downward OOD domain. However, if a significant amount of data is omitted, or a large number of training samples are mapped to the same answer, as shown in our counterexample above, the downward OOD domain generalization fails. Therefore, when our goal is to align the model to generalize certain OOD domains as expected, precise analysis of the task nature and careful control of the training data are necessary.

To validate our theoretical framework, we experiment on various transformer that shares the same architecture with many popular autoregressive transformer-based language models, including models of various sizes ~\citep{karpathy2023nanogpt}, and our tasks involving \( n \)-digit addition, multiplication, and modular operations. We further perform robustness analysis across different model scales, dataset sizes, and training data schemes. 

Our main contributions are as follows: 

\textbf{1. Establishing a unified theoretical framework for understanding generalization behavior of transformers on arithmetic tasks:} 
Our framework is the first to address task differences in transformer models' generalization ability in arithmetic tasks. We also provided comprehensive experimental evidences validate our theoretical predictions in Section~\ref{sec_experiment} and Appendix~\ref{appendix_results}. We make our code public to facilitate future research\footnote{We opensource our code at \url{https://github.com/xingchengxu/ArithmeticLLM} under the MIT license.}.

\textbf{2. Clarifying the downward and upward OOD generalization and their requirement of task and training data with an arithmetic setting.} We introduce the concepts of downward and upward generalization, which more clearly delineates the differences between generalization to downward and upward domains.

\section{Related Work}\label{sec:related_work}

\paragraph{Generalization of Transformers and LLMs on Arithmetic.} 
Numerous studies have examined the performance of Transformer-based language models in tasks involving arithmetic operations and mathematical reasoning. \citet{brown2020language}, \citet{bubeck2023sparks} and \citet{lu2024gpt} investigated various LLMs, such as GPT-3, GPT-4, and Gemini, in performing basic arithmetic and mathematical reasoning. \citet{nogueira2021investigating} explored the limitations of Transformers in learning arithmetic, highlighting the significant influence of surface representation on model accuracy and the need for improved tokenization and positional encoding strategies. Subsequent research such as \citet{qian2022limitations}, \citet{anil2022exploring}, \citet{jelassi2023length}, \citet{lee2023teaching},  \citet{xu2023ood}, \citet{mcleish2024transformers} and \citet{duan2024interpolation}. \citet{abbe2023generalization} examined generalization on unseen logical functions.
While previous studies have mainly focused on evaluating or improving generalization capabilities, our work develops a unified theoretical framework to analyze OOD generalization behaviors in Transformer models trained on arithmetic operations, bridging the gap between empirical observations and theoretical understanding.
\paragraph{Mechanistic Interpretability and General Understanding.} 
Many studies have focused on understanding and interpreting the working dynamics of neural networks and Transformer models~\citep{zhang2021survey,hernandez2022scaling,elhage2022toy,bills2023language,templeton2024scaling}. From the perspective of universal approximation, \citet{yun2019transformers} and \citet{alberti2023sumformer} demonstrated that Transformer models equipped with trainable positional encodings can act as universal approximators for continuous functions in a compact domain under the \(L^p\) norm or the supremum norm.

From a mechanistic viewpoint, \citet{hernandez2022scaling} investigated the impact of repeated data on the performance of LLMs, highlighting significant performance degradation when a small fraction of data is repeated multiple times. \citet{liu2022towards} addressed the phenomenon of delayed generalization or "grokking" using addition and modular addition tasks, and \citet{zhong2023clock} utilized modular addition to mechanistically explain algorithm discovery in neural networks.

Our work contributes to this growing field of mechanistic interpretability by providing a macroscopic explanation specifically for Transformer models. We systematically identify systematic biases and understand model behaviors in arithmetic reasoning scenarios.

\section{Theoretical Analysis on Generalization for Arithmetic Reasoning}\label{sec:addition}

We first review the Transformer model and the universal approximation theorem, and then conduct theoretical analyses of the downward and upward OOD generalization capabilities of the Transformer in solving tasks related to addition, modular addition, multiplication, and modular multiplication.

\subsection{Preliminaries on Transformer and Universal Approximation}
A Transformer model~\citep{vaswani2017attention} predicts the next token based on the preceding tokens within the input sequence. Its output is subsequently used as input for the next prediction. For a target token $x_{i}$ at position $i$
in the sequence, the model generates a probability distribution over
the vocabulary of the next potential tokens. To be precise, let $x=x_{1}x_{2}\ldots x_{T}\in\mathcal{V}^{T}$
denote the input sequence of tokens. The probability of observing
this sequence with respect to a Transformer model is given as follows: 
\begin{equation*}
\mathbf{P}_{\theta}(x)=\prod_{i=1}^{T}\mathbf{P}_{\theta}(x_{i}|x_{1},x_{2},...,x_{i-1})=\prod_{i=1}^{T}\mathbf{P}_{\theta}(x_{i}|x_{<i}).
\end{equation*}
The conditional probability $\mathbf{P}_{\theta}(x_{i}|x_{<i})$ is computed using softmax applied to the last hidden state. 

\paragraph{Universal approximation theorem for Transformer models:} Transformer models have the capacity to universally approximate any arbitrary continuous sequence-to-sequence function within a compact domain. \citet{yun2019transformers} and \citet{alberti2023sumformer} have shown that, when equipped with trainable positional encodings, Transformers can serve as universal approximators for continuous functions in a compact domain under the $L^p$ norm or the supremum norm. These characterizations highlight the representation power of fixed-width Transformer networks, despite the intrinsic parameter sharing and permutation equivariance. 
To facilitate the reader's navigation of the subsequent mathematical analysis, we first establish an intuitive understanding grounded in the Universal Approximation Theorem (UAT). The UAT demonstrates that Transformers, provided with appropriate conditions such as trainable positional encodings, are theoretically capable of approximating any continuous function within a compact domain. This inherent representational power leads to a critical insight regarding generalization: a Transformer's inability to generalize to an upward out-of-distribution (OOD) domain (e.g., longer-digit sequences) does not stem from a fundamental representational deficit. Rather, it signifies that the model, trained on a limited support (e.g., inputs up to length n), has learned a function whose approximation is effectively truncated at the boundary of its observed data. We leverage the UAT as a contrastive lens to argue that, when a Transformer fails in OOD scenarios despite its capacity to learn the correct function, the root causes lie in the interplay of task structure, the specific nature of positional encoding (e.g., Absolute Positional Encoding), and the characteristics of the training distribution.

\subsection{Theoretical Analysis on Addition}

Consider two natural numbers $a=\sum_{i=1}^{n}a_{i}\times 10^{i-1}=(a_1,a_2,\cdots,a_{n})$ and $b=\sum_{i=1}^{n}b_i\times 10^{i-1}=(b_1,b_2,\cdots,b_{n})$. The addition of these $n$-digit numbers, denoted as $f(a,b)=a+b$, is expressed by $c=\sum_{i=1}^{n+1}c_i\times 10^{i-1}=(c_1,c_2,\cdots,c_{n},c_{n+1})$.

Let the dataset $\mathcal{D}_{n}:=\{(a,b)\in\mathbb{N}^2:a_n \vee b_n\geq 1, a_i=b_i\equiv 0,\forall i>n\}$. For notation simplicity, assume $(0,0)\in\mathcal{D}_{1}$. Here, $a_n \vee b_n=\max\{a_n,b_n\}$. Note that $\mathcal{D}_{n}\cap \mathcal{D}_{m}=\emptyset$ for $n\neq m$ and $\mathbb{N}^2=\bigcup_{n=1}^{\infty}\mathcal{D}_{n}$. Denote the downward (downward) domain $\mathcal{D}_{<n}:=\bigcup_{m=1}^{n-1}\mathcal{D}_{m}$ and  the upward domain $\mathcal{D}_{>n}:=\bigcup_{m=n+1}^{\infty}\mathcal{D}_{m}$.

\begin{theorem}\label{thm:addition-ape}(\textbf{Informal})
    Assume a Transformer model with absolute positional embedding (APE) is trained on a multi-digit addition dataset for the operands $(a,b)\in \mathcal{D}_{n}$ ($n\geq 2$) with infinite training computation, then the learned model can perfectly generalize for the downward OOD domain $\mathcal{D}_{<n}$, but fail for the upward OOD domain $\mathcal{D}_{>n}$. 
\end{theorem}
\paragraph{Proof Sketch.}
Assume a Transformer model is trained on this dataset $\mathcal{D}_{n}$ using absolute positional embeddings (APE). The model is trained to approximate the function that computes the sum digit by digit, with carries propagated as follows:
\[
c_i = \zeta(a_i + b_i + c_{i-1}^\chi),
\]
where \(c_{i-1}^\chi\) is the carry from the previous position, and $\zeta$ is a function taking the units of the input.

\noindent\textbf{Case I: Downward OOD Domain (\(\mathcal{D}_{<n}\))}

For positions \(i \leq n\), the model can generalize well to the downward OOD domain \(\mathcal{D}_{<n}\) by \textit{universal approximation theorem for Transformer models}. Since the model has seen all possible carry combinations during training, it can correctly predict the digit sums at positions \(i = 1, 2, \cdots, n\). For position \(i = n+1\), the model predicts the carry \(c_{n+1} = c_n^\chi \in \{0, 1\}\) for all pairs where \(a_n \vee b_n \geq 1\), and when both \(a_n = b_n = 0\), the model learns \(c_{n+1} = 0\). For positions \(i > n+1\), the model predicts zero, since the input digits \(a_i\) and \(b_i\) are zero beyond the \(n\)-th position. Thus, the model perfectly generalizes to \(\mathcal{D}_{<n}\).

\noindent\textbf{Case II: Upward OOD Domain (\(\mathcal{D}_{>n}\))}

For positions \(i \leq n\), the model behaves similarly to the downward OOD case. However, when \(i = n+1\), the model is unable to predict the correct sum. The probability distribution learned by the model at this position only supports values in \(\{0, 1\}\), but for the model to correctly predict the carry, the support must include \(\{0, 1, \cdots, 9\}\). Since the model has never seen pairs where both \(a_{n+1}\) and \(b_{n+1}\) are non-zero, it cannot generalize correctly to the upward OOD domain. Beyond position \(n+1\), the model will predict zeros, as \(a_i = b_i = 0\) for all \(i > n\). Thus, the model fails to generalize to \(\mathcal{D}_{>n}\).
\qed

Based on the analysis above, we can immediately draw the following conclusion, which provides an explanation for the findings by~\citet{xu2023ood}.
\begin{corollary}\label{corollary:addition}(\textbf{Informal})
    The learned Transformer model with APE approximates the function $\hat{f}(a,b)=(a\ \operatorname{mod}\ 10^n)+(b\ \operatorname{mod}\ 10^n)$. The OOD generalization error is zero for the downward OOD domain $\mathcal{D}_{<n}$, but not less than $10^n$ for every point in the upward OOD domain $\mathcal{D}_{>n}$.
\end{corollary}

We are curious about the conditions under which a Transformer model can learn to perform addition operations. With APE, the model successfully generalizes downward, but fails to generalize upward. What would be the conclusion under RPE? Through theoretical and experimental analysis, we have arrived at the following conclusions.

\begin{theorem}\label{thm:addition-rpe}(\textbf{Informal})
    Assume a Transformer model with relative/abacus positional embedding (RPE) is trained on a multi-digit addition dataset for the operands $(a,b)\in \mathcal{D}_{n}$ ($n\geq 2$) with infinite training computation, then the learned model can perfectly generalize for the downward OOD domain $\mathcal{D}_{<n}$ and generalize well for the upward OOD domain $\mathcal{D}_{>n}$, with a probability of failure in the upward domain being less than $1/10^{n-1}$. 
\end{theorem}
\paragraph{Proof Sketch.}

A Transformer model with relative positional embeddings (RPE) has a key property of \textit{translation invariance}. This means the model’s predictions at any position \( i \) depend only on the relative distances between positions, not their absolute locations.

\noindent\textbf{Special Case: Translation Invariance}

Translation invariance can be expressed as:
\[
\mathbf{P}_{\theta}(c_i \mid a_{\leq i}, b_{\leq i}) = \mathbf{P}_{\theta}(c_i \mid a_{i-1}, a_i, b_{i-1}, b_i),
\]
ensuring that the carry at each position is determined by the preceding digits \( a_{i-1}, b_{i-1} \), not their absolute positions. Thus, the sum at position \( i \) is:
\[
c_i = \zeta(a_i + b_i + c_{i-1}^\chi),
\]
where \( c_{i-1}^\chi = \chi(a_{i-1} + b_{i-1}) \), as long as \( a_{i-1} + b_{i-1} \neq 9 \).

\noindent\textbf{General Case: Extended Translation Invariance}

For longer sequences, the prediction for \( c_i \) depends on the relative positions \( a_{i-n+1}, \cdots, a_i, b_{i-n+1},\cdots, b_i \). The translation invariance fails when carry propagation extends past the \( n \)-th digit, which happens if \( a_{i-k} + b_{i-k} = 9 \) for all \( k = 1, \cdots, n-1 \). The probability of this failure is small, less than \( 1/10^{n-1} \). Thus, the model effectively handles longer sequences by mapping them to shorter ones with similar relative distances, with the failure probability in the upward domain being less than \( 1/10^{n-1} \).
\qed

\subsection{Theoretical Analysis on Modular Addition}
\label{sec:mod_addition_theory}

Consider the function for modular addition with a modulus $p$, expressed as $f(a, b) = (a + b) \mod p$, which will be the focus of our analysis in the following section. Subsequently, we will also represent modular addition using the notation $\overline{c}^p = \overline{a + b}^p$. For simplicity, we will omit the superscript $p$ when it is clear from the context.

\paragraph{Scenarios on Divisibility of 10's Power by Modulus}

\begin{theorem}\label{thm:modular-addition-div}(\textbf{Informal})
    Assume a Transformer model with either absolute or relative/abacus positional embedding is trained on a multi-digit modular addition dataset with a modulus $p$ that divides $10^m$ for the operands $(a,b)\in \mathcal{D}_{n}$ ($n\geq 2$ and $m\leq n$) with infinite training computation, then the learned model can perfectly generalize both for the downward OOD domain $\mathcal{D}_{<n}$ and the upward OOD domain $\mathcal{D}_{>n}$. 
\end{theorem}
\paragraph{Scenarios on Non-Divisibility of 10's Power by Modulus}

\begin{theorem}\label{thm:modular-addition-nondiv}(\textbf{Informal})
    (1) Assuming a Transformer model equipped with absolute positional embeddings is trained on a multi-digit modular addition dataset $\mathcal{D}_{n}$ ($n \geq 2$) where the modulus $p$ neither divides $10^n$ nor exceeds $10^n$, and provided that infinite training computation is allocated, then the resulting trained model is capable of perfect generalization to the downward OOD domain $\mathcal{D}_{<n}$, while encountering difficulties in generalizing to the upward OOD domain $\mathcal{D}_{>n}$.     
    (2) The function that the model has learned is $\hat{f}^p(a,b)=\overline{\overline{a}^{10^n}+\overline{b}^{10^n}}^p$.     
    (3) Furthermore, the test accuracy on $\widetilde{\mathcal{D}}_{n_{\text{test}}}$ ($n_{\text{test}}>n$) is given by $\operatorname{Acc}(p,n,n_{\text{test}}) \approx \frac{\gcd(p, 10^n)}{p}$ if $n_{\text{test}}\geq n+\log_{10}(p'/2+1)$, otherwise $\operatorname{Acc}(p,n,n_{\text{test}}) =0$, where $\gcd(p, 10^n)$ represents the greatest common divisor of $p$ and $10^n$, and $p'=p/\gcd(p, 10^n)$.% and $0\leq \epsilon_{p,n,n_{\text{test}}}<1/(10^{n_{\text{test}}-n} - 1)$.
\end{theorem}

\subsection{Theoretical Analysis on Multiplication}
\label{sec:multiplication}

\begin{theorem}\label{thm:multiplication-ape}(\textbf{Informal})
    (1) Assuming a Transformer model equipped with absolute positional embeddings is trained on a multi-digit multiplication dataset $\mathcal{D}_{n}$ ($n \geq 2$), and provided that infinite training computation is allocated, then the resulting trained model is capable of perfect generalization to the downward OOD domain $\mathcal{D}_{<n}$, while it cannot generalize to the upward OOD domain $\mathcal{D}_{>n}$.    
    (2) The function that the model has learned is $\hat{f}(a,b)=\overline{a}^{10^n}\times \overline{b}^{10^n}$. 
\end{theorem}

\subsection{Theoretical Analysis on Modular Multiplication}
\label{sec:modular_multiplication}

\begin{theorem}\label{thm:modular-multip}(\textbf{Informal})
    (1) Assume that a Transformer model with absolute or relative/abacus positional embedding is trained on a multidigit modular multiplication dataset with a modulus $p$ that divides $10^m$ for operands $(a,b)\in \mathcal{D}_{n}$ ($n\geq 2$ and $m\leq n$) with infinite training computation, then the learned model can perfectly generalize both for the downward OOD domain $\mathcal{D}_{<n}$ and the upward OOD domain $\mathcal{D}_{>n}$.     
    (2) If the modulus $p$ neither divides $10^n$ nor exceeds $10^n$, and provided that infinite training computation is allocated, then the resulting trained model is capable of perfect generalization to the downward OOD domain $\mathcal{D}_{<n}$, while encountering difficulties in generalizing to the upward OOD domain $\mathcal{D}_{>n}$. 
    The function that the model with APE has learned is $\hat{f}^p(a,b)=\overline{\overline{a}^{10^n} \times \overline{b}^{10^n}}^p$. 
\end{theorem}

% \textbf{Remark}: As the modular addition with a modulus $p$ that does not divide $10^n$, the modular multiplication with such a modulus also does not exhibit the property of translation invariance. Therefore, there is no evident advantage to be gained from exploiting this characteristic and relative/abacus positional embedding.

\section{Experiments}\label{sec_experiment}

In this section, we describe our experiment design with result outcome validating the prediction make using our theoretical framework. We also conducted additional experiment providing detailed investigation into the the learning mechanism as well as checking for robustness of our result with different model scale, data digits, and with yet-to-converged models, provided in Appendix~\ref{appendix_results_addition}.

\subsection{Experimental Design}

\paragraph{Model Description:}
In line with most LLMs, we utilize a decoder-only architecture consisting of multiple layers and multi-head attentions. Our models are trained from scratch with varying model scale\footnote{The models architecture are in respect NanoGPT, MicroGPT, and MiniGPT \citep{karpathy2023nanogpt}} in \ref{table:model_size}. Detailed configuration of training and architecture are provided in Table \ref{table:hyperparameters} in Appendix \ref{training_hyper_config}.

\begin{table}[ht]
\small
\begin{centering}
\begin{tabular}{>{\centering}m{2.0cm}>{\centering}m{1.5cm}>{\centering}m{1.5cm}>{\centering}m{1.5cm}}
\toprule 
\centering{}\textbf{Hyperparameter} & \centering{}\textbf{NanoGPT} & \centering{}\textbf{MicroGPT} & \textbf{MiniGPT}\tabularnewline
\midrule
num layer & 3 & 4 & 6\tabularnewline
num head & 3 & 4 & 6\tabularnewline
dim embd  & 48 & 128 & 384\tabularnewline
vocab size & 16 & 16 & 16\tabularnewline
context window & 256 & 256 & 256\tabularnewline 
\bottomrule
\end{tabular}
\par\end{centering}
\caption{Model Scale}
\label{table:model_size}
\end{table}

\paragraph{Data Description:}
We employ four primary arithmetic operations with different symmetric property as well as difficulty in term of how much a digit can have impact in term of upward/downward generalization, which are described here:
\begin{itemize}[nosep,leftmargin=*,itemsep=0pt,parsep=0pt]
    \item \textbf{Addition:} \( c = a + b \)
    \item \textbf{Modular addition:} \( c \equiv a + b \pmod{p} \)
    \item \textbf{Multiplication:} \( c = a \times b \)
    \item \textbf{Modular multiplication:} \( c \equiv a \times b \pmod{p} \)
\end{itemize}

We randomly generate datasets for each arithmetic task. Following ~\citep{lee2023teaching,xu2023ood} we organize our training data as a sequence of operand pairs in natural order, with the results of the operations in reversed order with character-level tokenization\footnote{After the tokenization, ";", "[bos]", and "[eos]", a "line break" token are added to the beginning and the end of each line of data, resulting in a vocabulary size of 16. When the context window exceeds the required size for $n$-digit arithmetic operations, we pad zeros before the numbers "$a$", "$b$", and "$c$".}, which has been shown to be more effective for learning in next-token prediction models in arithmetic tasks\footnote{For example, consider an $n$-digit addition $a + b = c$, represented in standard format as "$a_{n} \cdots a_{2}a_{1} + b_{n} \cdots b_{2}b_{1} = c_{n+1} \cdots c_{2}c_{1}$". By reversing the order of the output "$c$", we obtain the reversed data format "$a_{n} \cdots a_{2}a_{1} + b_{n} \cdots b_{2}b_{1} = c_{1} \cdots c_{n}c_{n+1}$".}. An example input would look like "$[\text{bos}]a_{n} \cdots a_{2}a_{1} + b_{n} \cdots b_{2}b_{1} = $" for addition, and the model output would be in the format "$c_{1} \cdots c_{n}c_{n+1}[\text{eos}];\text{\textbackslash n}$", where $a_1, \cdots, a_n, b_1, \cdots, b_n, c_1 \cdots c_n$ are single digit integers and "[bos]", "[eos]", ";", and a special "line break token" are special tokens. The data format that can be used to train other arithmetic tasks can be obtained respectively.\footnote{We also provided code and data for generate training data and training the models in our repository at \url{https://github.com/xingchengxu/ArithmeticLLM}.}

\begin{table}[ht]
\small
\begin{centering}
\begin{tabular}{>{\centering}m{1.0cm}>{\centering}m{2.0cm}>{\centering}m{1.5cm}}
\toprule 
\centering{}\textbf{DataSet} & \centering{}\textbf{Description} & \centering{}\textbf{Obs.Num} \tabularnewline
\midrule
$\mathcal{D}_{4}$ & $n=4, m=4$ & 100,000 \tabularnewline
$\mathcal{D}_{5}$ & $n=5, m=5$ & 100,000 \tabularnewline
$\mathcal{D}_{4,5}$ & 0.5$D_4$+0.5$D_5$ & 100,000 \tabularnewline
\bottomrule
\end{tabular}
\par\end{centering}
\caption{Data Scale: Training Data}
\caption*{\textit{Note:} We provided a more detailed DataSet description table in Table \ref{table:data_size_full} at Appendix \ref{appendix_data}.}
\label{table:data_size}
\end{table}

We control the length of arithmetic operations $n$ and randomly generate datasets from \(\mathcal{D}_n\) for different lengths $n$. These datasets for each arithmetic task are categorized into three distinct subsets: a training set, an in-distribution (ID) test set, and additional out-of-distribution (OOD) test sets which we further break down by the upper bound digit for upward generalization, sampled from $m$-digit operations with $m \neq n$. The case where $m < n$ is referred to as the downward (downward) OOD domain, and the case where $m > n$ is termed the upward (upward) OOD domain. We also construct numerous combination sets of samples from different domains \(\mathcal{D}_n\), such as \(\mathcal{D}_{n-1,n}\), to be used as training and ID test datasets. In the demonstrative example, the OOD test sets are  sampled from \(\mathcal{D}_m\) with $m \neq n-1$ and $n$. The test accuracy is measured using maximum probability sampling. For the modular addition tasks and modular multiplication tasks, we selected moduli with varying relationships to $10^{n}$, namely varied by the divisibilily by $10^n$,  coprime relationship to $10^n$, and whether do the moduli has a greatest common divisor with $10^n$ that is neither $1$ nor $p$ (the modulus). The choice of $p$ is made to demonstrate the relationship  between the modulus $p$ in modular arithmetic and the maximum length $n$ of the training set.

\begin{figure}[ht]
\centering
\begin{minipage}{0.5\textwidth}
\centering
\begin{subfigure}{0.46\textwidth}
  \includegraphics[width=\textwidth,height=0.75\textwidth]{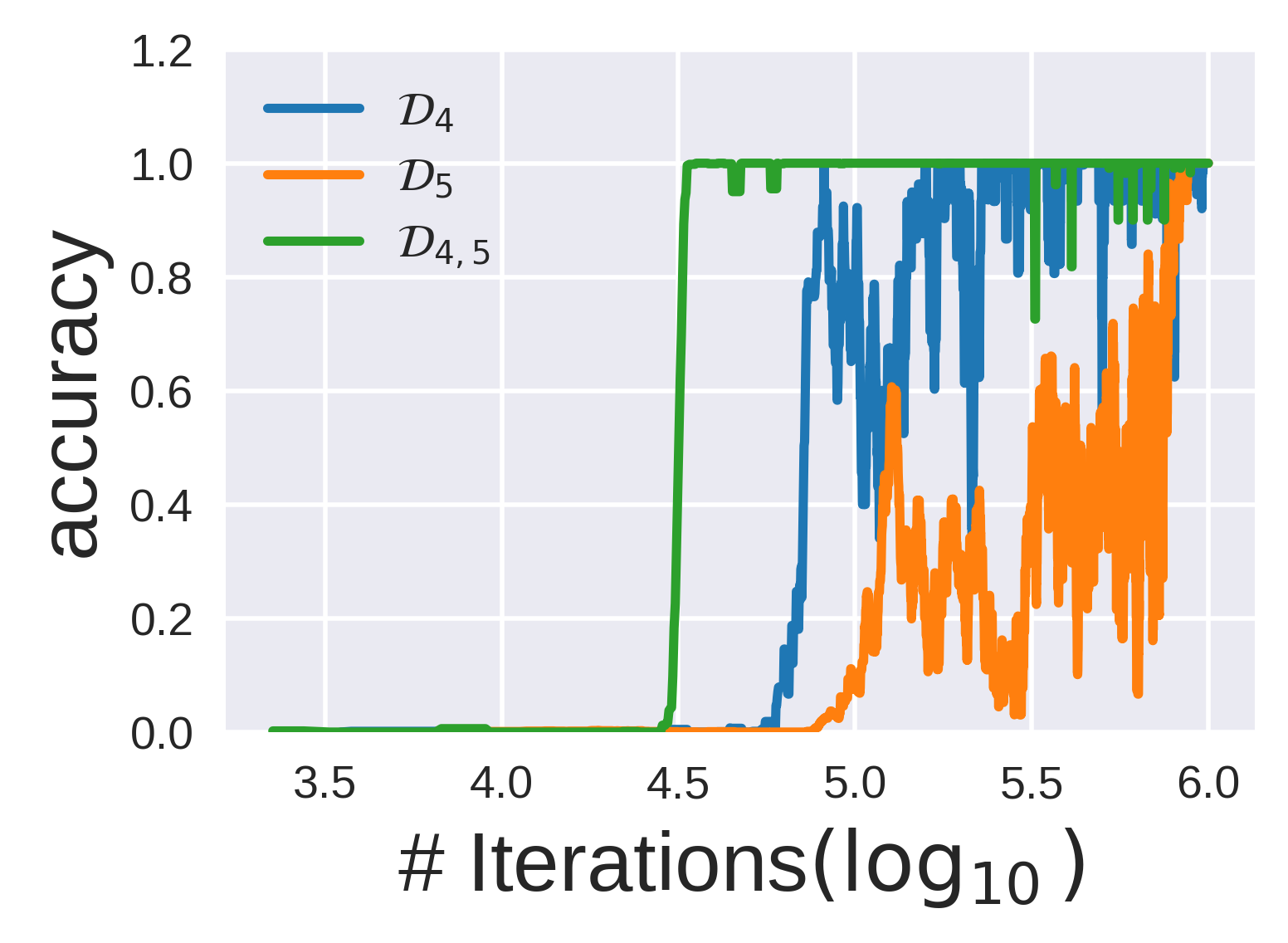}
  \caption*{{\footnotesize $\mathcal{D}_1$ Task}}
\end{subfigure}
\hfill
\begin{subfigure}{0.46\textwidth}
  \includegraphics[width=\textwidth,height=0.75\textwidth]{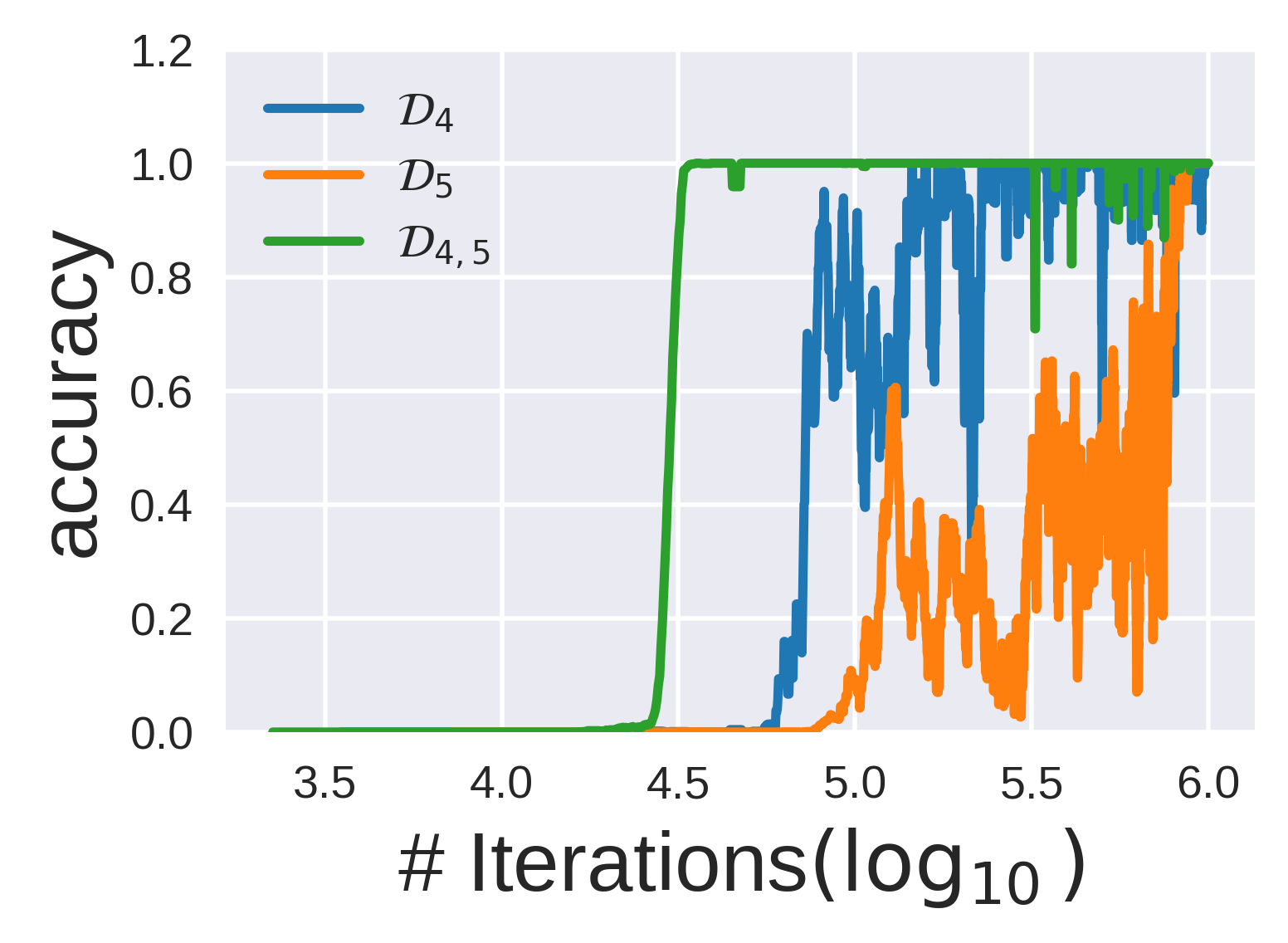}
  \caption*{\footnotesize $\mathcal{D}_2$ Task}
\end{subfigure}
\\
\begin{subfigure}{0.46\textwidth}
  \includegraphics[width=\textwidth,height=0.75\textwidth]{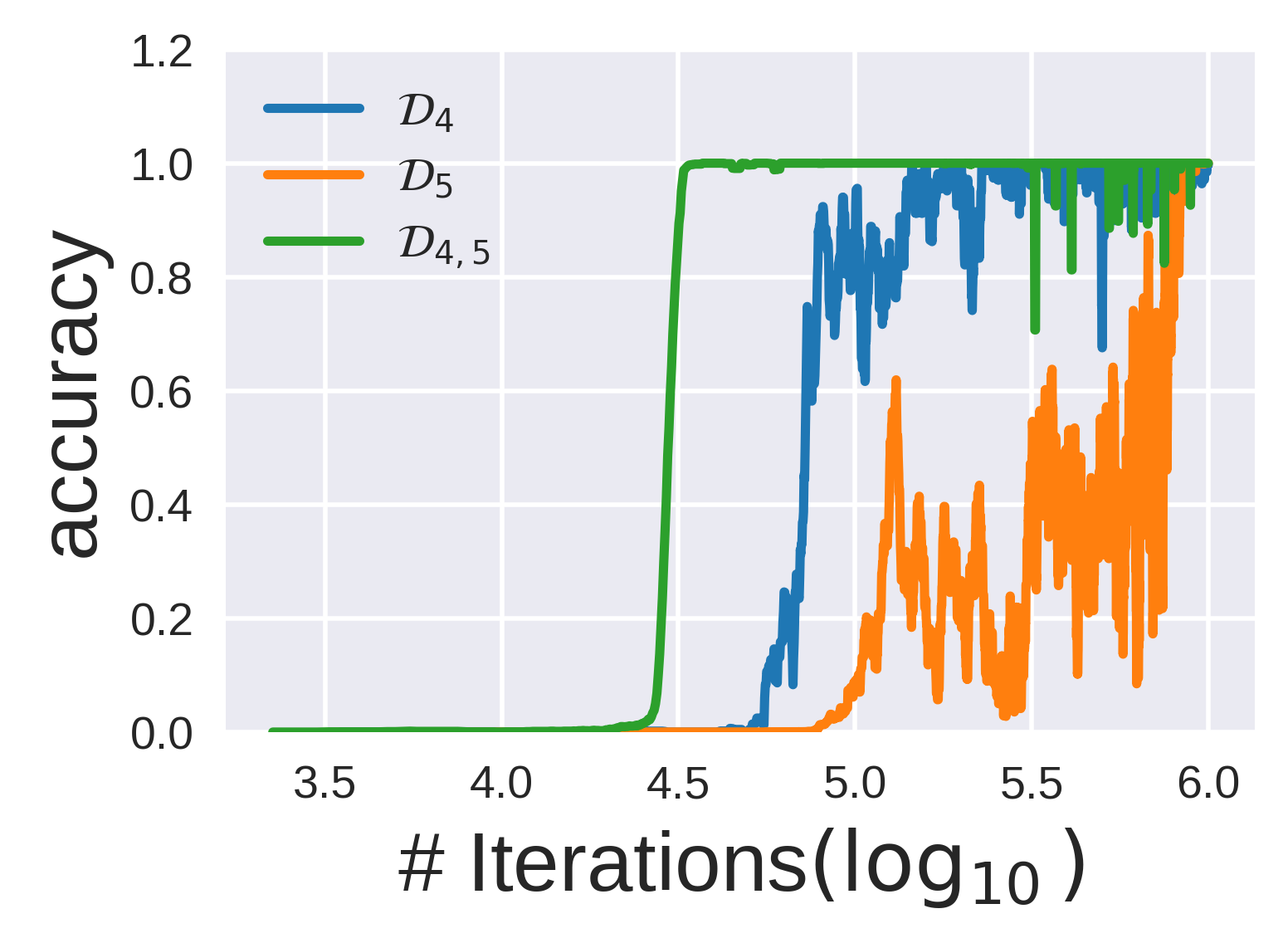}
  \caption*{\footnotesize $\mathcal{D}_3$ Task}
\end{subfigure}
\hfill
\begin{subfigure}{0.46\textwidth}
  \includegraphics[width=\textwidth,height=0.75\textwidth]{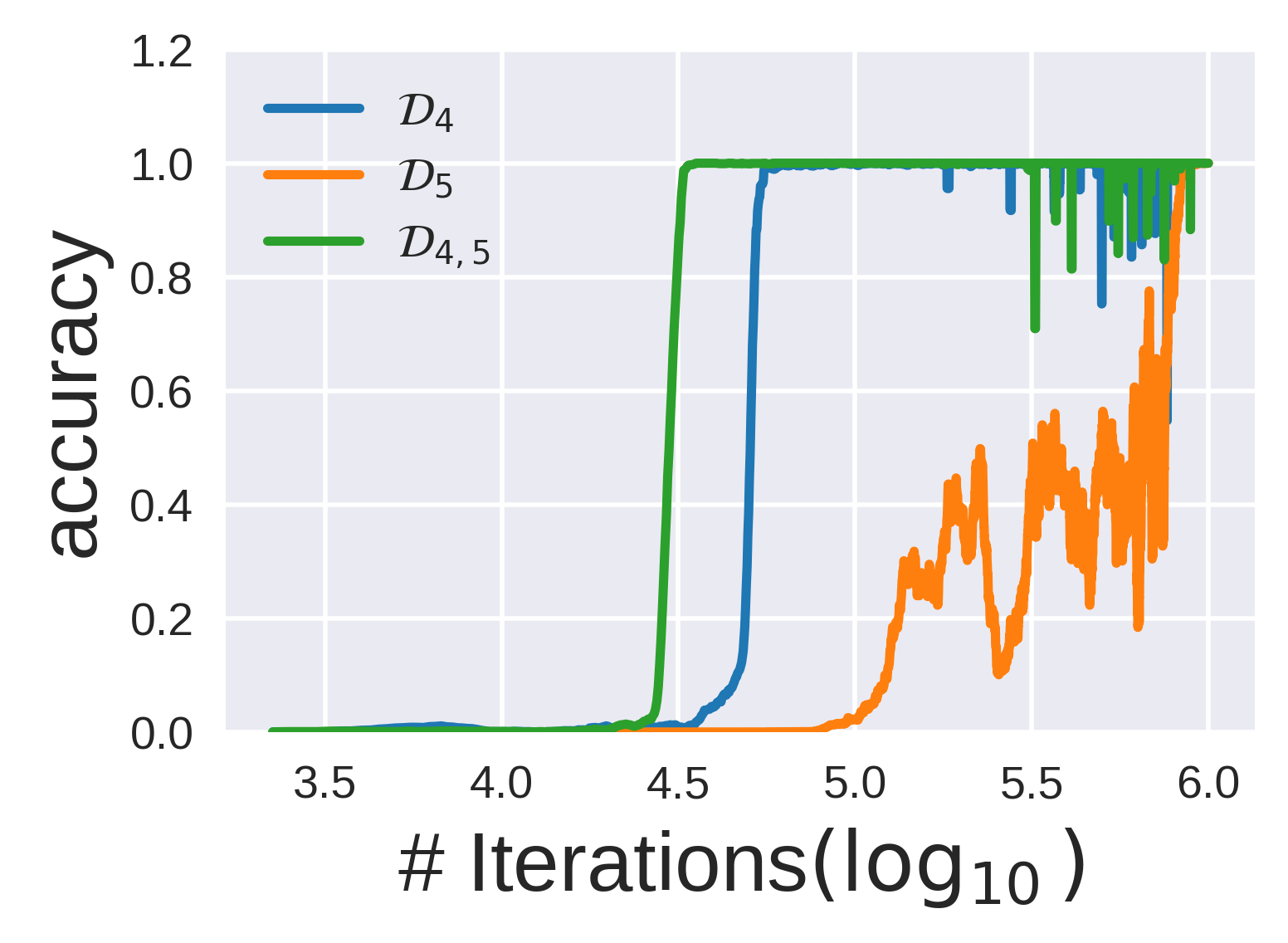}
  \caption*{\footnotesize $\mathcal{D}_4$ Task}
\end{subfigure}
\\
\begin{subfigure}{0.46\textwidth}
  \includegraphics[width=\textwidth,height=0.75\textwidth]{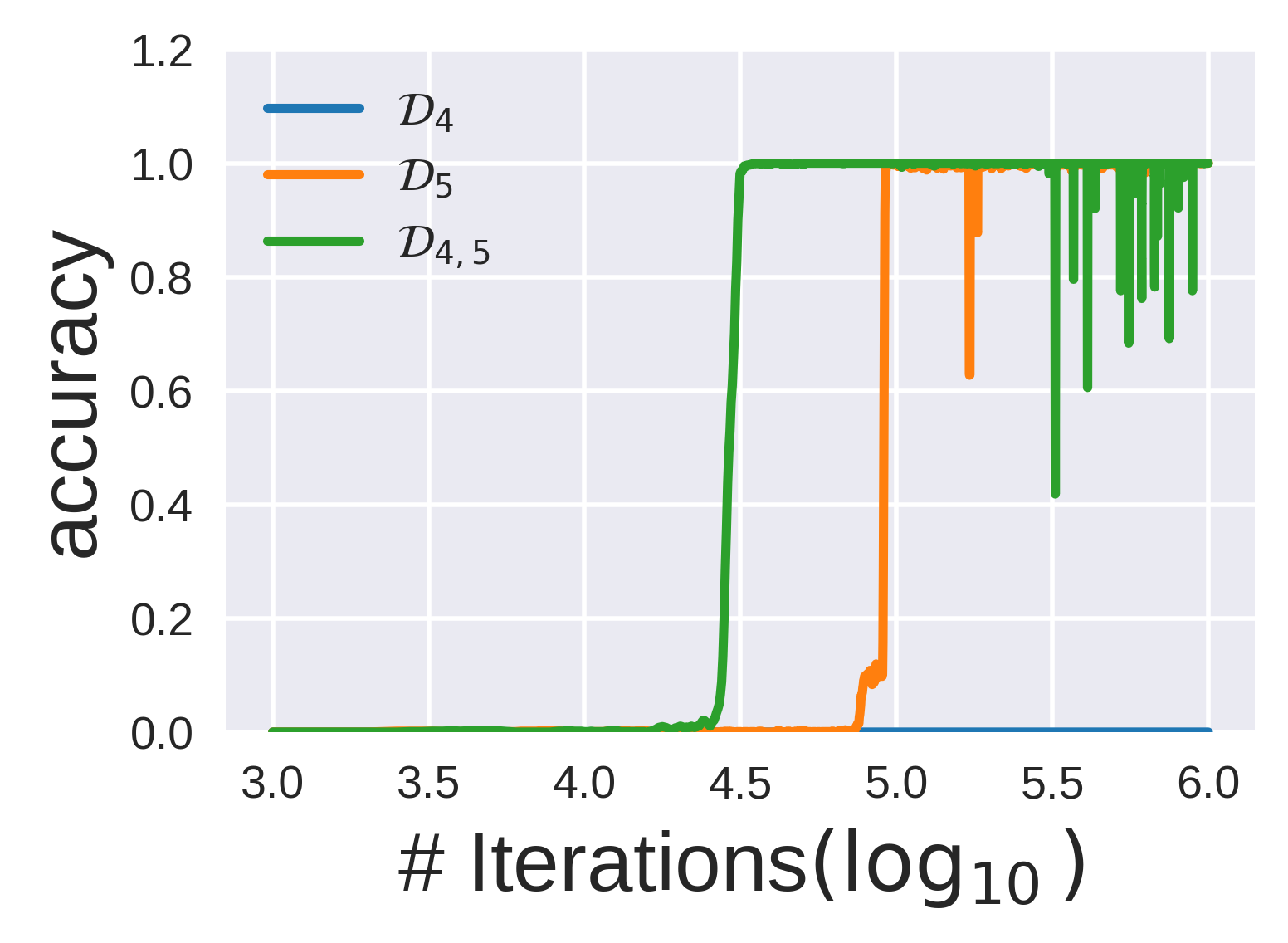}
  \caption*{\footnotesize $\mathcal{D}_5$ Task}
\end{subfigure}
\hfill
\begin{subfigure}{0.46\textwidth}
  \includegraphics[width=\textwidth,height=0.75\textwidth]{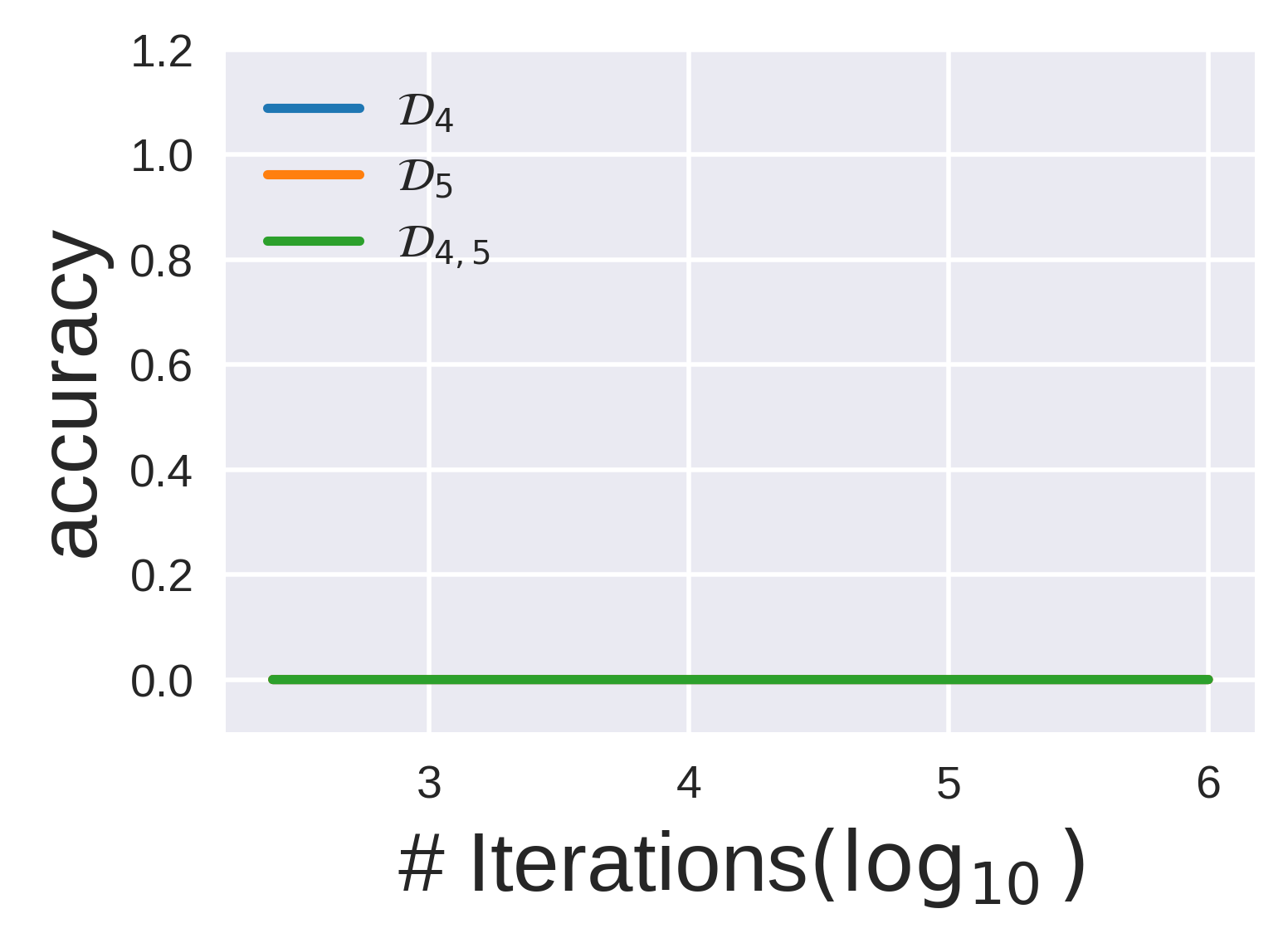}
  \caption*{\footnotesize $\mathcal{D}_6,\cdots,\mathcal{D}_9$ Tasks}
\end{subfigure}
\end{minipage}

\caption{Test Accuracy of Transformer Models with APE for Different Multi-digit Addition Tasks}
\caption*{\textit{Note:} This figure presents results from three experiments using different training datasets with the MiniGPT model and a learned APE. The labels $\mathcal{D}_4$, $\mathcal{D}_5$, and $\mathcal{D}_{4,5}$ indicate training on random samples from $\mathcal{D}_4$, $\mathcal{D}_5$, and a combined subset of both, respectively. Each subfigure shows test accuracy across different domains $\mathcal{D}_{i}$ during training.}
\label{fig:addition_acc_tasks}
\end{figure}

% Modular Addition: Ground Truth Accuracy (with colors)
\begin{table*}[ht]
\centering
\begin{tabular}{lrrrrrrrrr|r}
\toprule
\multicolumn{1}{c}{} & \multicolumn{9}{c}{Test Accuracy (\%) w.r.t. the Ground Truth on the Domain $\widetilde{\mathcal{D}}_i$} & \multicolumn{1}{c}{Theory} \\
Modulus & 1 & 2 & 3 & 4 & 5 & 6 & 7 & 8 & 9 & $1/p'$\\
\midrule
$p=50$ & 100 & 100 & 100 & 100 & \textcolor{red}{99.3} & \textcolor{red}{92.0} & \textcolor{red}{93.1} & \textcolor{red}{95.2} & \textcolor{red}{91.4} & 100\\
$p=51$ & 100 & 98.5 & 99.9 & 99.3 & \textcolor{gray}{0.3} & \textcolor{gray}{1.8} & \textcolor{gray}{1.9} & \textcolor{gray}{1.9} & \textcolor{gray}{1.6} & 1.96\\
$p=100$ & 100 & 100 & 100 & 100 & \textcolor{red}{100} & \textcolor{red}{100} & \textcolor{red}{100} & \textcolor{red}{100} & \textcolor{red}{100} & 100\\
$p=101$ & 100 & 100 & 100 & 100 & \textcolor{gray}{0.0} & \textcolor{gray}{1.2} & \textcolor{gray}{0.9} & \textcolor{gray}{1.1} & \textcolor{gray}{1.0} & 0.99\\
$p=150$ & 100 & 100 & 100 & 100 & \textcolor{blue}{33.2} & \textcolor{blue}{33.6} & \textcolor{blue}{32.3} & \textcolor{blue}{33.0} & \textcolor{blue}{33.7} & 33.3\\
$p=151$ & 100 & 99.9 & 99.9 & 100 & \textcolor{gray}{0.0} & \textcolor{gray}{0.6} & \textcolor{gray}{0.7} & \textcolor{gray}{0.7} & \textcolor{gray}{0.6} & 0.66\\
$p=200$ & 100 & 100 & 100 & 100 & \textcolor{red}{99.8} & \textcolor{red}{98.9} & \textcolor{red}{93.7} & \textcolor{red}{94.1} & \textcolor{red}{93.5} & 100\\
$p=201$ & 100 & 100 & 99.9 & 99.9 & \textcolor{gray}{0.0} & \textcolor{gray}{0.0} & \textcolor{gray}{0.5} & \textcolor{gray}{0.4} & \textcolor{gray}{0.5} & 0.50\\
\bottomrule
\end{tabular}
\caption{Modular Addition: Test Accuracy w.r.t. the Ground Truth $f^p(a,b)=\overline{a+b}^p$ on $\widetilde{\mathcal{D}}_i$}
\caption*{\textit{Note:} All the Transformer models in above experiments are instances of MiniGPT with a learned APE, which have been trained on a random sample drawn from $\mathcal{D}_4$ with 100,000 random training sample (except $p=150$). The accuracy is tested on 10,000 random test samples (when \( n > 2 \)), otherwise on the entire dataset. The outputs of models are generated using maximum probability sampling.}
\label{table:modular_addition_true_acc}
\end{table*}

\subsection{Experiments on Addition}

In this subsection, we trained multiple models on different datasets (e.g. $\mathcal{D}_4$, $\mathcal{D}_5$, $\mathcal{D}_{4,5}$) and tracked the changes in their accuracy. Additionally, we demonstrated how the models learn each digit during the training process.

\subsubsection{Generalization for Different Digit Tasks}

In Figure~\ref{fig:addition_acc_tasks}, we present the results of three different experiments using distinct training datasets (i.e., $\mathcal{D}_4$, $\mathcal{D}_5$, $\mathcal{D}_{4,5}$). For all experiments, we employ the MiniGPT model equipped with a learned APE. Each subfigure illustrates the test accuracy on different test domains $\mathcal{D}_{i}$ for these models throughout the training process. Figure~\ref{fig:addition_acc_tasks} verifies our Theorem~\ref{thm:addition-ape}. It demonstrates that models incorporating APE are unable to generalize to longer digits than those they are trained on but can succeed with lower digits. Additionally, the model trained on $\mathcal{D}_5$ has a much more challenging training process compared to the model trained on $\mathcal{D}_4$, while the model trained on $\mathcal{D}_{4,5}$ experiences the easiest and smoothest training process among the three models. The reason, as explained in Theorem~\ref{thm:addition-ape}, is that for $\mathcal{D}_{4,5}$, the model learns addition tasks on lower digits directly from the training data. In contrast, $\mathcal{D}_{4}$ and $\mathcal{D}_{5}$ require OOD generalization for the edge positions. 

More results can be found in Table~\ref{table:addition_true_acc} and Table~\ref{table:addition_modular_acc}. We test the final trained model on datasets with varying digit lengths. While the models do not learn the addition of higher digits, they successfully learn the operation $\hat{f}(a,b)=\overline{a}^{10^n}+\overline{b}^{10^n}$, supporting our Corollary~\ref{corollary:addition}.

We also conduct extensive experiments using various training datasets, model scales, and data scales. The results of these experiments are robust, and presented in Appendix \ref{appendix_results}.

\subsubsection{Learning Dynamics for Each Digit Position}

The models and training datasets are identical to those described in Figure~\ref{fig:addition_acc_tasks}. We have assembled a comprehensive test dataset that contains a random sample from $\mathcal{D}_1$ to $\mathcal{D}_9$. Our objective is to demonstrate how these Transformer models equipped with APE learn each digit at every position throughout the training phase. The digit-wise test accuracy is defined as the accuracy of the prediction for each position in the result $c$. 

The plots in Figure~\ref{fig:addition_acc_digits} (see Appendix) %~\ref{appendix_results_addition}) 
visually represent whether these models are capable of accurately predicting the digits $c_i$ at all positions. These graphs effectively illustrate the learning dynamics for each token in the context of addition tasks. The models exhibit high accuracy for the first four or five digits, with accuracy approaching 1.0 as training progresses, for datasets \(\mathcal{D}_4\), or \(\mathcal{D}_5\), and \(\mathcal{D}_{4,5}\), respectively. However, accuracy sharply declines for the 5th or 6th digits and remains near zero for the 7th, 8th, and 9th digits. These findings illustrate that while the models can effectively learn and predict lower-position digits, they struggle significantly with higher-position digits. This aligns with the theorem that Transformer models with APE can generalize well for downward OOD domains but fail for upward OOD domains.

\subsubsection{Generalization Under Relative/Abacus Positional Embeddings}

\citet{mcleish2024transformers} conducted experiments using a 16-layer Transformer (decoder only) model with abacus positional embedding, trained on a random sample from \( \mathcal{D}_{\leq 20} \). It can generalize on 100-digit addition problems (Figure~\ref{fig:addition-abacus} in Appendix \ref{appendix_results}) 
\footnote{Code to reproduce the results can be found on GitHub: \url{https://github.com/mcleish7/arithmetic}.}. Additionally, \citet{jelassi2023length} demonstrated that relative positional embeddings enable length generalization in addition tasks. In their work, models such as Transformer and Universal Transformer (encoder only) trained to add 5-digit numbers could generalize to 20-digit operands.

These results provide empirical evidence validating our Theorem~\ref{thm:addition-rpe} for upward OOD generalization. The findings are clear, and we will not replicate the procedures here. Instead, we reference these studies in the present context.

\subsection{Experiments on Modular Addition}

The results in Table~\ref{table:modular_addition_true_acc} validate Theorem~\ref{thm:modular-addition-div}, which states that Transformer models with absolute positional embeddings trained on multi-digit modular addition datasets exhibit distinct generalization capabilities based on the modulus \( p \). For moduli such as \( p = 50, 100, 200 \) that divide \( 10^n \), the models achieve perfect test accuracy across all digit domains, demonstrating their ability to generalize flawlessly to both downward and upward OOD domains. In contrast, for moduli such as \( p = 51, 101, 150, 151, 201 \) that do not divide \( 10^n \), the models maintain high accuracy for lower digit domains but show significant performance degradation for higher digit positions\footnote{The task of performing addition modulo 150 requires an extended training duration in our experiment. To facilitate this, we prime the training process with samples that have downward additions.}.

The OOD test accuracy in Table~\ref{table:modular_addition_true_acc} for high-order digits can be completely expected using Theorem~\ref{thm:modular-addition-nondiv}, which states that the  test accuracy on $\widetilde{\mathcal{D}}_{n_{\text{test}}}$ ($n_{\text{test}}>n$) is given by $\operatorname{Acc}(p,n,n_{\text{test}}) \approx 1/p'$ if $n_{\text{test}}\geq n+\log_{10}(p'/2+1)$, otherwise $\operatorname{Acc}(p,n,n_{\text{test}}) =0$. These observations align well with the theoretical expectations outlined in Theorem~\ref{thm:modular-addition-div} and Theorem~\ref{thm:modular-addition-nondiv}, also explaining the experimental results found in the literature (see, e.g., \citet{jelassi2023length}) in handling modular addition tasks with different moduli. 

Furthermore, the results in Table~\ref{table:modular_addition_modular_acc} (see Appendix) support Theorem~\ref{thm:modular-addition-nondiv}, indicating that Transformer models with absolute positional embeddings trained on multi-digit modular addition datasets learns the function $\hat{f}^p(a,b)=\overline{\overline{a}^{10^n}+\overline{b}^{10^n}}^p$ for any modulus $p$. These findings fully align with the theoretical predictions.

\subsection{Experiments on Multiplication and Modular Multiplication}

We also conducted extensive experimental analyses for multiplication and modular multiplication tasks, examining the performance and generalization capabilities of Transformer models. These experiments are designed to test various configurations, including different positional encodings, model size and training data schemes. Detailed results and additional analyses are available in Appendix.%~\ref{appendix_results_multiplication} and \ref{appendix_results_modular_multiplication}. 
The experimental outcomes consistently support our theoretical framework, demonstrating the robustness of our approach and providing further insights into the behavior of Transformer models in arithmetic reasoning tasks.

\section{Discussion}\label{sec:discussion}

Our study sheds light on the \textit{mechanistic interpretability} of Transformer models. Understanding the learning mechanisms is crucial for ensuring the meaningfulness of learned representations. % Our theoretical framework provides a pathway for interpreting how these models generalize from training data to unseen tasks. This understanding is essential for aligning models with human-defined objectives, and reducing the risk of unintended behaviors.

% We also observed phenomena akin to the \textit{satori phenomenon and emergence}, where models suddenly exhibit a leap in understanding or capability once a critical threshold in training or data complexity is reached. This emergent behavior underscores the non-linear nature of model learning and highlights the need for further research into the conditions that trigger such phenomena.

Additionally, our work identifies challenges associated with different training data schemes, such as concatenation training without padding\footnote{e.g. "$123+45=168;267+1=268;$" as input.} and line-by-line padding training\footnote{e.g. "$123+45=168;$[pad][pad][pad]" as input.}. These approaches can significantly impact model performance and generalization. Further understanding on these problems is essential for refining training strategies to improve model robustness and generalization.

\section{Conclusion}\label{sec:conclusion}

In this paper, we developed a unified theoretical framework to explain OOD generalization in Transformer models trained on arithmetic operations, categorizing generalization into downward OOD (downward domains) and upward OOD (upward domains). Our analysis highlights the interactions among task properties, training data coverage, and model characteristics. Experiments with NanoGPT, MicroGPT, and MiniGPT validate our predictions, highlighting the framework's robustness. This work clarifies generalization mechanisms and provides insights for efficient model training and AI alignment. Future research should extend this framework to more complex tasks and factors influencing OOD generalization.

\section{Limitation}\label{sec:limitation}
This paper presents a unified theoretical framework for understanding generalization in transformers applied to arithmetic tasks. However, there are notable limitations to our analysis. Firstly, our focus on length generalization may overlook other critical aspects of out-of-distribution (OOD) generalization, as the representations learned for different tasks can exhibit varying relationships with length.

We selected arithmetic tasks for this study due to their clarity in distinguishing between downward and upward OOD generalization, as well as our ability to control the training data distribution effectively. Nonetheless, our framework's predictions are predicated on the assumption that the model has converged on the training data, which may not always hold true in practice, particularly given that many LLMs remain undertrained.

Additionally, while our findings provide insights into generalization behaviors, they may not fully encompass the complexities involved in more intricate mathematical reasoning or other types of sequence-to-sequence tasks. Future work should explore these broader contexts to enhance our understanding of transformer generalization.

\section*{Acknowledgments}
This work is supported by Shanghai Artificial Intelligence Laboratory.

% Bibliography entries for the entire Anthology, followed by custom entries
%\bibliography{anthology,custom}
% Custom bibliography entries only
\bibliography{custom}

\newpage
\appendix

\section{Appendix on Transformer}
A Transformer model~\citep{vaswani2017attention} predicts the next token based on the preceding tokens within the input sequence. Its output is subsequently used as input for the next prediction. For a target token $x_{i}$ at position $i$
in the sequence, the model generates a probability distribution over
the vocabulary of potential next tokens. To be precise, let $x=x_{1}x_{2}\ldots x_{T}\in\mathcal{V}^{T}$
denote the input sequence of tokens. The probability of observing
this sequence with respect to a Transformer model is given as follows: 
\begin{equation*}
\mathbf{P}_{\theta}(x)=\prod_{i=1}^{T}\mathbf{P}_{\theta}(x_{i}|x_{1},x_{2},...,x_{i-1})=\prod_{i=1}^{T}\mathbf{P}_{\theta}(x_{i}|x_{<i}).
\end{equation*}
The conditional probability $\mathbf{P}_{\theta}(x_{i}|x_{<i})$ is computed using the softmax function applied to the last hidden state. One way to design
this model (see e.g. \citet{karpathy2023nanogpt}, \citet{brown2020language}) is as follows:
\begin{equation*}
\begin{aligned}a^{\ell-1} & =h^{\ell-1}+\mathrm{MHA}_{\ell}(\mathrm{LN}_{\ell}^{A}(h^{\ell-1}))\\
h^{\ell} & =a^{\ell-1}+\mathrm{MLP}_{\ell}(\mathrm{LN}_{\ell}^{F}(a^{\ell-1}))
\end{aligned}
\end{equation*}
for $\ell=1,2,\ldots,L$, with the initial embedding $h^{0}=e_{tok}+e_{pos}$,
where $e_{tok}$ represents the initial token embedding and $e_{pos}$
represents the positional embedding. In the context of GPT-series LLMs, $\mathrm{MHA}_{\ell}$
refers to the masked multi-head attention of the $\ell$-th layer,
$\mathrm{MLP}_{\ell}$ is a multi-layer perception with one hidden
layer, and $\mathrm{LN}$ represents layer normalization. Define $f_{\ell}$
such that $h^{\ell}=f_{\ell}(h^{\ell-1})$. Consequently, the final
hidden state of this LLM is 
\begin{equation*}
\ensuremath{h^{L}=f_{L}\circ\ldots\circ f_{2}\circ f_{1}(h^{0})\in\mathbb{R}^{d_{m}\times T}},
\end{equation*}
where $d_{m}$ is the embedding dimension.

Let $X=\mathrm{LN}(h^{L})=[X_{1},X_{2},\ldots,X_{T}]$. The final
output conditional probability matrix
\begin{equation*}
\begin{split}
    \mathbf{P}_{\theta} & =\mathrm{softmax}(WX) \\ 
    & =\left(\frac{\exp(WX_{i})}{\sum_{j=1}^{N}\exp(WX_{i})_{j}}\right)_{i=1,2,\cdots,T}\in[0,1]^{N_V\times T},
\end{split}
\end{equation*}
where $W\in\mathbb{R}^{N_V\times d_{m}}$ is a weight matrix. The $i$-th
column of the matrix $\mathbf{P}_{\theta}$ represents the conditional probability
$\mathbf{P}_{\theta}(\tilde{x}_{i}|x_{<i})$ for any $\tilde{x}_{i}\in\mathcal{V}$. By training on a large corpus of language texts, the LLMs provide the estimated probabilities.

\section{Proofs of Theorems}

\subsection{Proof of Theorem~\ref{thm:addition-ape}}
Define the functions 
$$\chi(x):=\lfloor x/10 \rfloor \text{ and } \zeta(x):=x \ \operatorname{mod}\ 10, \text{ for } x\in \mathbb{N}.$$
Then $c_i=\zeta(a_i+b_i+c_{i-1}^\chi),\forall i$, and the carry $c_i^{\chi}=\chi(a_i+b_i+c_{i-1}^\chi)$. 
For simplicity, assume $a_0=b_0=0$.

We define three forms of approximation:
\begin{itemize}
    \item \emph{Strong form}: If \( \mathbf{P}_{\theta}(\tilde{c} = c_i \mid a + b = c_{<i}) = 1 \) for any \( i \geq 1 \). This means the model \( \mathbf{P}_{\theta}(\cdot \mid a + b = c_{<i}) \) can perfectly learn the function \( c_i = \zeta(a_i+b_i+c_{i-1}^\chi), \forall i \).
    
    \item \emph{Standard form}: If \( c_i = \arg\max_{\tilde{c}} \mathbf{P}_{\theta}(\tilde{c} \mid a + b = c_{<i}) \) for any \( i \geq 1 \). This means the model \( \mathbf{P}_{\theta}(\cdot \mid a + b = c_{<i}) \) can approximate the function \( c_i = \zeta(a_i+b_i+c_{i-1}^\chi), \forall i \) with the highest probability.
    
    \item \emph{Weak form}: If \( \mathbf{P}_{\theta}(\tilde{c} = c_i \mid a + b = c_{<i}) > 0 \) for any \( i \geq 1 \). This means the model \( \mathbf{P}_{\theta}(\cdot \mid a + b = c_{<i}) \) can approximate the function \( c_i = \zeta(a_i+b_i+c_{i-1}^\chi), \forall i \) with a non-zero probability.  
\end{itemize}

In the following, we will use the standard form to demonstrate out-of-distribution (OOD) generalization. When training a Transformer model on $\mathcal{D}_n$-addition using absolute positional embedding (APE), the learned model approximates the function at each position of $c$:
$$\mathbf{P}_{\theta}(c_i \mid a_{\leq i},b_{\leq i})\to c_i = \zeta(a_i+b_i+c_{i-1}^\chi).$$ %\mathbf{P}_{\theta}(c_i \mid a + b = c_{<i})=

\noindent\textbf{Case I: Downward OOD Domain} 

Let us consider the Downward OOD domain $\mathcal{D}_{<n}$ case. If $i<n$, the model trained on a sample dataset in $\mathcal{D}_n$ can at least approximate the function $c_i$ in the standard form. If $i=n$, 
$$\mathbf{P}_{\theta}(c_n \mid a_{\leq n},b_{\leq n})\to c_n = \zeta(a_n+b_n+c_{n-1}^\chi)$$
for every $a_n\vee b_n\geq 1$ except the case $a_n=b_n=0$ simultaneously. If $i=n+1$, $$\mathbf{P}_{\theta}(c_{n+1} \mid a_{\leq n+1},b_{\leq n+1})\to c_{n+1} =c_{n}^\chi\in \{0,1\}$$ % = \zeta(a_{n+1} + b_{n+1} + c_{n}^\chi)
for every pair $(a_n,b_n)$ with $a_n\vee b_n\geq 1$ and $a_{n+1}=b_{n+1}=0$. In the case where $a_n=b_n=0$, the conditions for both $i=n$ and $i=n+1$ necessitate OOD generalization. Since the model has been trained to approximate \(c_n\) accurately for \(a_n \vee b_n \geq 1\), it has learned the function for the carry-over mechanism properly. When \(a_n = b_n = 0\), the digit \(c_n\) purely depends on the carry from the previous position. For \(i = n+1\), the carry \(c_{n}^\chi\) is correctly learned such that it maps \(\{0, 1\}\) depending on whether there was a carry from the \(n\)-th digit. With \(a_n = b_n = 0\), the model correctly sets \(c_{n+1} = 0\). The training on \(\mathcal{D}_n\) includes all possible carry scenarios and digit summations for \(a_n, b_n \in \{0, \ldots, 9\}\). The zero cases are naturally included in the learned patterns\footnote{If the training dataset has significant gaps, such as when a model is trained on \(n\)-digit addition but only with \(a_n, b_n \geq n_0\) (e.g., \(a_n, b_n \geq 6\)), it means the model never encounters pairs where both \(a_n < 6\) and \(b_n < 6\). While the digit-wise addition and carry mechanisms for positions 1 through \(n-1\) are learned correctly, since these positions involve a full range of digit pairs during training, the model fails to learn proper behavior for the \(n\)-th and \((n+1)\)-th positions. Specifically, for these positions, the model will not encounter any pairs where both digits are simultaneously less than 6. In this scenario, \(\zeta(a_n + b_n) \in \{2, 3, \ldots, 8\}\) (missing the digits 0, 1, 9), and \(c_{n}^\chi \equiv 1\) (missing the digit 0). Consequently, the training dataset lacks complete coverage of all possible carry scenarios and digit summations. This substantial gap negatively affects the model's ability to handle these edge situations. Thus, the final learned model cannot generalize to the OOD domain \(\mathcal{D}_{<n}\). Specifically, you will observe that the \((n+1)\)-th position value \(c_{n+1} \equiv 1\) for all samples in \(\mathcal{D}_{<n}\).}.
For $i\geq n+2$, 
$$\mathbf{P}_{\theta}(c_i \mid a_{\leq i},b_{\leq i})\to c_i = \zeta(a_i+b_i+c_{i-1}^\chi)\equiv 0,$$
since $a_i=b_i\equiv 0$ for any $(a,b)\in \mathcal{D}_n$ with $i\geq n+1$.
Thus, the model $\mathbf{P}_{\theta}$ can approximate the function of $c$ at every position for the downward OOD domain $\mathcal{D}_{<n}$. 

\noindent\textbf{Case II: Upward OOD Domain}  

Consider the Upward OOD domain $\mathcal{D}_{>n}$ case. If $i\leq n$, the analysis remains the same as above. The learned model $\mathbf{P}_{\theta}$ can predict the correct numbers at these positions. However, when $i=n+1$, 
$$\mathbf{P}_{\theta}(c_{n+1} \mid a_{\leq n+1},b_{\leq n+1})\to c_{n+1} =c_{n}^\chi\in \{0,1\}$$  % = \zeta(a_{n+1} + b_{n+1} + c_{n}^\chi)
for every pair $(a_n,b_n)$ with $a_n\vee b_n\geq 1$ and $a_{n+1}=b_{n+1}=0$. Note that for inference in the OOD domain $\mathcal{D}_{>n}$, the model needs to predict each sample with $(a_{n+1},b_{n+1})$ at least for every $a_{n+1}\vee b_{n+1}\geq 1$. However, the support of probability measure learned by the model $\mathbf{P}_{\theta}$ is $\operatorname{supp}\mathbf{P}_{\theta}=\{0,1\}$. For the model to predict $c_{n+1}$ correctly even in the weak form, the support should be $\operatorname{supp}\mathbf{P}_{\theta}=\{0,1,\cdots,9\}$. This indicates that the model $\mathbf{P}_{\theta}$ cannot predict the number at position $n+1$. Additionally, the learned probability $\mathbf{P}_{\theta}(c_{n+1} \mid a_{\leq n+1},b_{\leq n+1})$ is actually independent of $(a_{n+1},b_{n+1})$. 
For $i\geq n+2$, 
$$\mathbf{P}_{\theta}(c_i \mid a_{\leq i},b_{\leq i})\to c_i \equiv 0,$$
since $a_i=b_i\equiv 0$ for any $(a,b)\in \mathcal{D}_n$ with $i\geq n+1$. This means that the learned model maps all inputs to zeros for positions $i\geq n+2$. If the model could predict the numbers at positions $i\geq n+2$, the requirement even in the weak form is that at least $\{0,1\}\subset \operatorname{supp}\mathbf{P}_{\theta}(c_i\mid \cdots)$. This contradicts $\operatorname{supp}\mathbf{P}_{\theta}(c_i\mid \cdots)=\{0\}$. Combining the above analysis, we conclude that the learned model $\mathbf{P}_{\theta}$ cannot solve the problems in the OOD domain $\mathcal{D}_{>n}$ but instead outputs the result $(a\ \operatorname{mod}\ 10^n)+(b\ \operatorname{mod}\ 10^n)$ for every sample in $\mathcal{D}_{>n}$.
\qed

\subsection{Proof of Theorem~\ref{thm:addition-rpe}.}

We begin by noting the key property that under the assumption of relative positional embedding (RPE), the Transformer model possesses a form of \textit{translation invariance}. This property implies that the prediction at any position \( i \) is invariant to the shift of the entire sequence, as long as the relative distances between positions remain unchanged.

\noindent\textbf{Special Case:}

The translation invariance property is mathematically expressed as:
\begin{equation*}
    \begin{split}
        \mathbf{P}_{\theta}(c_i & \mid a_{\leq i}, b_{\leq i}) = \mathbf{P}_{\theta}(c_i \mid a_{i-1}, a_i, b_{i-1}, b_i) \\
        & = \mathbf{P}_{\theta}(c_{i+j} \mid a_{i+j-1}, a_{i+j}, b_{i+j-1}, b_{i+j}),
    \end{split}
\end{equation*}
for any \( i, j \in \mathbb{N} \), provided that \( a_{i-1} + b_{i-1} \neq 9 \).

This translation invariance arises when the carry \( c_{i-1}^\chi \) is determined by the previous digits \( a_{i-1} \) and \( b_{i-1} \), and thus does not depend on any global position or the absolute positions of the digits in the sequence. In fact, we have:
\[
c_i = \zeta(a_i + b_i + c_{i-1}^\chi),
\]
where \( c_{i-1}^\chi = \chi(a_{i-1} + b_{i-1}) \), provided that \( a_{i-1} + b_{i-1} \neq 9 \).

\noindent\textbf{General Case:}

The failure of the above translation invariance property occurs when the carry \( c_{i-1}^\chi \) is influenced by more digits beyond \( a_{i-1} \) and \( b_{i-1} \). A generalized translation invariance property should be used, i.e., 
\begin{equation*}
    \begin{split}
        &\quad \mathbf{P}_{\theta}(c_i \mid a_{\leq i}, b_{\leq i}) \\
        &= \mathbf{P}_{\theta}(c_i \mid a_{i-n+1}, \cdots, a_i, b_{i-n+1}, \cdots, b_i) \\
        & = \mathbf{P}_{\theta}(c_{i+j} \mid a_{i+j-n+1}, \cdots, a_{i+j}, b_{i+j-n+1}, \cdots, b_{i+j}).
    \end{split}
\end{equation*}
The failure for above formula happens when carry propagation extends beyond the maximum length \( n \) seen during training, i.e., when the carry is influenced by positions greater than \( n \). The case only happens when \( a_{i-k} + b_{i-k} = 9 \) for all $k=1,\cdots,n-1$.

The probability of this failure is quite small. Specifically, it is less than \( 1/10^{n-1} \), because the probability of the carry propagating beyond the maximum digit position \( n \) (in a dataset where all digits are restricted to the range 0-9) diminishes exponentially as the length of the sequence increases. This ensures that such failures are rare, especially for large \( n \).

For the upward OOD domain \(\mathcal{D}_{>n}\), the model faces the challenge of predicting the carry propagation for positions \( i > n \). However, since the model and addition satisfies translation invariance, this ensures that the model can handle longer sequences by effectively "folding" them into smaller, equivalent-length sequences with the same relative distances between digits, with only a probability of failure in the upward domain being less than $1/10^{n-1}$.
\qed

\textbf{Remarks on APE and RPE}: APE encodes positional information based on the absolute positions of tokens in a sequence. This approach can limit a model's ability to generalize to sequences of different lengths or to handle out-of-distribution scenarios effectively. In contrast, RPE captures translation-invariant positional dependencies by encoding the relative distances between tokens. This method allows the model to focus on the relationships between tokens regardless of their absolute positions, enhancing its ability to generalize across varying sequence lengths and to better understand contextual relationships. Consequently, RPE is more robust and adaptable in the addition context compared to APE. Our theoretical framework can explain the addition-based experimental findings reported in the following references: \citet{jelassi2023length}, \citet{xu2023ood}, \citet{duan2024interpolation}, and \citet{mcleish2024transformers}.

\subsection{Proof Sketch of Theorem~\ref{thm:modular-addition-div}.} We will initially focus on the scenario where $p = 10^m$, and subsequently explore the general case where $p$ is a divisor of $10^m$.

\textbf{Case I}: Let us revisit the equation for modular addition, which states that $\overline{c}^p = \overline{a + b}^p = \overline{\overline{a}^p + \overline{b}^p}^p$. The above equation shows that for the case $p = 10^m$, the digits in positions higher than $m$ in numbers $a$ and $b$ do not affect the result $\overline{c}^p$; only the digits in positions $m$ and lower have an impact. Furthermore, we have $\overline{c}^p=(\overline{c}^p_{1},\overline{c}^p_{2},\cdots,\overline{c}^p_{m})=(c_1,c_2,\cdots,c_m)$, where $c=a+b$. A model trained on $\mathcal{D}_{n}$ is capable of approximating the digits at positions ranging from $1$ to $m$. This can be expressed as:
$$\mathbf{P}_{\theta}(\overline{c}^p_{i} \mid a_{\leq i},b_{\leq i})\to \overline{c}^p_{i} = \zeta(a_i+b_i+c_{i-1}^\chi),$$
for $i=1,\cdots,m.$ All these functions are learned directly from the training data without the need for out-of-distribution (OOD) generalization if $m<n$, while $m=n$, only the $n$-th term $\overline{c}^p_{n}$ need OOD generalization. For $i>m$, the probability $\mathbf{P}_{\theta}(\overline{c}^p_{i} \mid \cdot)\equiv 0$. The aforementioned conclusions apply to both domains $\mathcal{D}_{<n}$ and $\mathcal{D}_{>n}$.

\textbf{Case II}: Consider the case where $p$ is a divisor of $10^m$. Since we have $\overline{c}^p = \overline{a + b}^p = \overline{\overline{a+b}^{10^m}}^p$,  the result $\overline{c}^p$ is indeed not influenced by the digits in positions higher than $m$ in numbers $a$ and $b$. If let $m$ be the minimum number which the $m$-th power of 10 can be divided by the modulus $p$, i.e. $m=\arg\min\{\tilde{m}: p\mid 10^{\tilde{m}}\}$, the model approximates the function at each position $i$:
$$\mathbf{P}_{\theta}(\overline{c}^p_{i} \mid a_{\leq m},b_{\leq m})\to \overline{c}^p_{i} = f_i^p(a_{\leq m},b_{\leq m}),$$
for $i=1,\cdots,m$, where $f_i^p$ is the function for $\overline{c}^p_{i}$ at the position $i$. As an aside, it is worth noting that in the case described above, the function is more intricate than standard addition or modular addition with a modulus that divides a power of 10. These functions generally rely on the digits at all positions of the numbers $a$ and $b$, from position $1$ through $m$. All these functions can be learned directly from the training data without the need for OOD generalization when training on $\mathcal{D}_{n}$ ($n\geq m$) except the term $\overline{c}^p_{n}$. 
\qed

\subsection{Proof Sketch of Theorem~\ref{thm:modular-addition-nondiv}.} In this case, the model approximates the function for each position $i$ as follows when training on $\mathcal{D}_n$:
$$\mathbf{P}_{\theta}(\overline{c}^p_{i} \mid a_{\leq n},b_{\leq n})\to \overline{c}^p_{i} = f_i^p(a_{\leq n},b_{\leq n}),$$
for $i=1,\cdots,n$, where $f_i^p$ represents the function for $\overline{c}^p_{i}$ at position $i$. Generally, the function $f^p(a,b)=(a+b)-\lfloor (a+b)/p \rfloor p$. Each digit $f^p_i$ depends on all positions of $a$ and $b$. If the model is trained on $\mathcal{D}_{n}$, the aforementioned probabilities have been trained exclusively on scenarios where $a_n \vee b_n \geq 1$. The case where $a_n = b_n = 0$ requires OOD generalization for samples on the downward domain $\mathcal{D}_{<n}$. This can be addressed by aligning with the model trained on the domain containing $\mathcal{D}_{n-1,n}$. If the model is trained on the dataset $\mathcal{D}_{n-1,n}$, which includes the case where $a_n = b_n = 0$, it learns the relevant patterns directly from the training data without the need for OOD generalization on the domain $\mathcal{D}_{<n}$. However, the model typically struggles to generalize to the upward domain $\mathcal{D}_{>n}$. This is because the model is expected to approximate the functions $f^p(a,b)=\overline{a+b}^p$, which consider all digits of $a$ and $b$. Since the model is trained on $\mathcal{D}_{n}$, it learns the function $\hat{f}^p(a,b)=\overline{\overline{a}^{10^n}+\overline{b}^{10^n}}^p$, which is independent of the positions $i>n$ of the numbers $a$ and $b$. 

\paragraph{OOD Test Accuracy Analysis for Longer Length.} 
For the model's output to be correct, it must satisfy the condition $\overline{a+b}^p=\overline{\overline{a}^{10^n}+\overline{b}^{10^n}}^p$. This requirement also provides us with a method to estimate the OOD test accuracy on the upward domain $\mathcal{D}_{>n}$.

Let $H_n=\overline{a}^{10^n}+\overline{b}^{10^n}$, and $R_n=(a+b)-H_n$. The OOD generalization error is then 
\begin{equation*}
f^p(a,b)-\hat{f}^p(a,b) = R_n - \left(\left\lfloor (a+b)/p\right\rfloor - \left\lfloor H_n/p \right\rfloor \right)p.
\end{equation*}
Denote $\varepsilon_n^{R}:=\frac{R_n}{p}-\lfloor \frac{R_n}{p} \rfloor\in [0,1)$ and $\varepsilon_n^{H}:=\frac{H_n}{p}-\lfloor \frac{H_n}{p} \rfloor\in [0,1)$. Then 
\begin{equation*}
\begin{split}
& \quad f^p(a,b)-\hat{f}^p(a,b) \\
& = (R_n/p - \left\lfloor (R_n+H_n)/p\right\rfloor + \left\lfloor H_n/p \right\rfloor)p \\
& = (\varepsilon_n^{R} - \lfloor\varepsilon_n^{R}+\varepsilon_n^{H} \rfloor)p.
\end{split}
\end{equation*}
That is, 
\begin{equation*}
\begin{split}
    & \quad f^p(a,b)-\hat{f}^p(a,b)\\
    & =
        \begin{cases}
        \varepsilon_n^{R} p\geq 0, &\text{if } \varepsilon_n^{R}+\varepsilon_n^{H}\in [0,1)\\
        (\varepsilon_n^{R}-1) p<0, &\text{if } \varepsilon_n^{R}+\varepsilon_n^{H}\in [1,2)
        \end{cases}.
\end{split}
\end{equation*}
For the special case where $\varepsilon_n^{R}=0$ (i.e. $R_n$ is divisible by $p$), we have $\hat{f}^p(a,b)=f^p(a,b)$.
This implies that the OOD test accuracy for a finite OOD test dataset may be greater than 0. 

The OOD test accuracy on the domain (denote as $\widetilde{\mathcal{D}}_{n_{test}}$ and $n_{test}>n$) in which the length of $a,b$ are both $n_{test}$ is $\operatorname{Acc}(p,n,n_{\text{test}})=\frac{\#\{(a,b)\in \widetilde{\mathcal{D}}_{n_{test}}:\varepsilon_n^R=0\}}{\# \widetilde{\mathcal{D}}_{n_{test}}}$. This can be calculated by counting the number of $R_n$ divisible by $p$ in this domain. 
The theoretical test accuracy on $\widetilde{\mathcal{D}}_{n_{\text{test}}}$ is given by
$\operatorname{Acc}(p,n,n_{\text{test}}) \approx \frac{1}{p'}$ if $n_{\text{test}}\geq n+\log_{10}(p'/2+1)$, otherwise 0. The proof can be found in the following section on test accuracy analysis.
\qed

Let's consider some examples. For $p=151$ and $n=4$, since $\gcd(151,10^n)\equiv 1$, the test accuracy is $\operatorname{Acc}(151,4,n_{\text{test}}) = \frac{1}{151}\approx 0.66\%$ if $n_{\text{test}}\geq 6$, but $0$ when $n_{\text{test}}=5$. For $p=201$ and $n=4$, the test accuracy is $\operatorname{Acc}(201,4,n_{\text{test}}) = \frac{1}{201}\approx 0.5\%$ if $n_{\text{test}}\geq 7$, but $0$ when $n_{\text{test}}=5,6$. Another example is $p=150$ and $n=4$, where the greatest common divisor is $\gcd(150, 10^4)=50$ and $p'=3$, resulting in a test accuracy of $\operatorname{Acc}(150,4,n_{\text{test}}) = \frac{50}{150}\approx 33.3\%$ for all $n_{\text{test}}\geq 5$. In the extreme case where $p$ is a divisor of $10^n$, the test accuracy $\operatorname{Acc}(p,n,n_{\text{test}})\equiv 100\%$. This aligns with the results for the scenarios on the divisibility of a power of 10 by the modulus. All these findings are confirmed by our experimental analysis (see Table~\ref{table:modular_addition_true_acc} and Table~\ref{table:modular_addition_modular_acc}).

\subsection{Proof Sketch of Theorem~\ref{thm:multiplication-ape}.} 
Given two natural numbers \(a\) and \(b\), each represented by \(n\)-digit sequences \((a_1, a_2, \ldots, a_n)\) and \((b_1, b_2, \ldots, b_n)\), respectively, the product \(ab\) is expressed as a \(2n\)-digit number \(c = (c_1, c_2, \ldots, c_{2n})\).

To express each digit \(c_i\) of the product \(c\) in terms of the digits of \(a\) and \(b\), we need to understand the multiplication task and how the digits interact. The product \(ab\) can be represented as:
\begin{equation*}
\begin{split}
ab & = \left(\sum_{i=1}^{n} a_i \cdot 10^{i-1}\right) \left(\sum_{j=1}^{n} b_j \cdot 10^{j-1}\right) \\
& = \sum_{i=1}^{n} \sum_{j=1}^{n} a_i b_j \cdot 10^{(i-1)+(j-1)}.
\end{split}
\end{equation*}
This gives us a double sum where each term \(a_i b_j\) contributes to a specific power of 10. To express the digit \(c_k\) (where \(1 \leq k \leq 2n\)) of the product, we need to collect all terms from the expansion that contribute to the \(10^{k-1}\) place.

For \(c_k\), we consider all pairs \((i, j)\) such that \(i + j - 2 = k - 1\), which simplifies to \(i + j = k + 1\). Define that the raw sum \(c_k^{R}\) at the $k$-th position as follows: 
\[ c_k^{R} = \sum_{\substack{1\leq i, j\leq n \\ i + j = k + 1}} a_i b_j.\]
However, since this is a digital product and carries might affect higher places, the correct formulation needs to account for carries from previous steps. The process of digit-wise calculation and adjustment with carries are as follows:

1. Initialize carry \(c_0^{\chi} = 0\).

2. Calculate the sum for each digit place:
   \[ S_i = c_i^{R} + c_{i-1}^{\chi} =\sum_{\substack{1 \leq i', j' \leq n \\ i' + j' = i + 1}} a_{i'} b_{j'} + c_{i-1}^{\chi}, \]
   where \(a_{i'}\) and \(b_{j'}\) are zeros if their indices are out of bounds.
   
3. Determine the digit and carry:
   \[ c_i = \zeta(S_i),\quad c_i^{\chi} = \chi(S_i). \]

Here, $\zeta(x):=x \ \operatorname{mod}\ 10$ and $\chi(x):=\lfloor x/10 \rfloor$, for $x\in \mathbb{N}$. This recursive formula provides the digits of the product considering the carries correctly. Denote that $c_i=f_i(a_1,\cdots,a_{i\wedge n},b_1,\cdots,b_{i\wedge n})$ for $i=1,2,\cdots,2n$. A Transformer model $\mathbf{P}_{\theta}(c_i\mid a\times b=c_1\cdots c_{i-1})=\mathbf{P}_{\theta}(c_i\mid a_1,\cdots,a_{i\wedge n},b_1,\cdots,b_{i\wedge n})$ will learn to approximate these functions $f_i$ when given enough data and computation power. 

Consider the longer length OOD domain $(a,b) \in \mathcal{D}_{>n}$. Let $\overline{a} = \overline{a}^{10^n}$ and $\overline{b} = \overline{b}^{10^n}$. The function learned by a Transformer model with absolute positional embeddings (APE) when trained with $(a,b) \in \mathcal{D}_{n-1,n}$ is then
\begin{equation*}
\hat{f}(a,b) = \overline{a}^{10^n} \cdot \overline{b}^{10^n} = \overline{c} = (\overline{c}_1,\overline{c}_2,\cdots,\overline{c}_{2n},0,\cdots,0)
\end{equation*}
with $\overline{c}_i = f_i(a_1,\cdots,a_{i \wedge n},b_1,\cdots,b_{i \wedge n}),\ 1 \leq i \leq 2n$, as all terms related to $a_i, b_i$ for $i > n$ are discarded during the training process. If the true value of $ab$ is $c$, then $\overline{c}_i = c_i$ for $1 \leq i \leq n$, but generally differs from $c_i$ when $i > n$ since $\overline{c}_i$ neglects the contribution of higher terms (greater than $n$) of $a$ and $b$.

Note that when a Transformer model is trained on domain $\mathcal{D}_n$, if $i<n$, the model learns the function $f_i(a_1,\cdots,a_{i \wedge n},b_1,\cdots,b_{i \wedge n})$ directly from the training data. However, when $i\geq n$, the model learns the function $f_i(a_1,\cdots,a_{n},b_1,\cdots,b_{n})$ only for the case where $a_n\vee b_n\geq 1$. In the scenario where $a_n=b_n=0$, the model requires OOD generalization. The training on \(\mathcal{D}_n\) includes all possible carry scenarios and digit summations (here, we only need consider the units and tens digits of $c^R_i$ and $c^{\chi}_{i-1}$) for \(a_n, b_n \in \{0, \ldots, 9\}\). The zero cases where $a_n=b_n=0$ are naturally included in the learned patterns.
\qed

\subsection{Proof Sketch of Theorem~\ref{thm:modular-multip}.} 

The proof resembles the process for modular addition. Suppose $\overline{c}^p = \overline{ab}^p$. When $p$ is a divisor of $10^m$, we have $\overline{c}^p = \overline{\overline{ab}^{10^m}}^p$. The value of $\overline{c}^p$ remains unaffected by the digits in positions beyond $m$ in the numbers $a$ and $b$. Now, let $m$ be the smallest number such that the $m$-th power of 10 is divisible by the modulus $p$, i.e., $m=\arg\min\{\tilde{m}: p\mid 10^{\tilde{m}}\}$. The model approximates the function for each position $i$ as follows:
$$\mathbf{P}_{\theta}(\overline{c}^p_{i} \mid a_{\leq m},b_{\leq m})\to \overline{c}^p_{i} = f_i^p(a_{\leq m},b_{\leq m}),$$
for $i=1,\cdots,m$, where $f_i^p$ represents the function for the $i$-th digit of $\overline{c}^p$. All these functions can be learned directly from the training data without the need for OOD generalization when training on $\mathcal{D}_{n}$ ($n\geq m$) except the term $\overline{c}^p_{n}$.

When $p$ is not a divisor of $10^n$ and $p<10^n$, the model approximates the function $\hat{f}^p(a,b)=\overline{\overline{a}^{10^n}\times \overline{b}^{10^n}}^p$ at each position $i$.This is because the model has been trained on $\mathcal{D}_{n}$, which is agnostic to the digits in positions $i>n$ of the numbers $a$ and $b$.
\qed

\section{Remarks}

\paragraph{Remarks on Theorem~\ref{thm:addition-ape}:} The challenging aspect of model prediction in the downward OOD domain $\mathcal{D}_{<n}$ arises from the need to generalize the $n$-th and $(n+1)$-th positions in the result $c$ when trained on $\mathcal{D}_{n}$. Specifically, these positions must be generalized to the scenario where $a_{n}=b_{n}=0$. Through our experimental analysis, we confirmed that the positions $n$ and $n+1$ are the last to be learned during the training process. An additional observation is that if the model is trained on the domain $\mathcal{D}_{n-1,n}:=\mathcal{D}_{n-1}\cup \mathcal{D}_{n}$, the previously mentioned challenge is mitigated. This is because the case with $a_{n}=b_{n}=0$ is already incorporated into the training dataset. Consequently, the positions $n$ and $n+1$ do not require OOD generalization; instead, they are learned directly from the training data. We have also conducted experiments based on this training scheme and found that learning on the domain that includes $\mathcal{D}_{n-1,n}$ is significantly easier than learning on $\mathcal{D}_{n}$ alone.

\paragraph{Remark on Transformer models based on relative/abacus positional embedding:} The standard addition benefits from the property of translation invariance, whereas modular addition or modular multiplication with a modulus $p$ that does not divide $10^n$ lacks this property. Consequently, there is no apparent advantage to be gained from leveraging this characteristic.

\section{Difficulty for Learning Multiplication}

\paragraph{Transition Invariance Property in Multiplication.}
The transition invariance property for multiplication refers to the idea that the position of digits in the multiplication process can be shifted or "transitioned" in a systematic way that still respects the overall structure of multiplication. In the context of digit-wise multiplication, each digit \(c_i\) should be adjusted by the previous carry. This process is transition invariant because each digit's place calculation transitions in a smooth and systematic way from one digit place to the next, maintaining the structure of the multiplication.

Transformers can utilize properties like transition invariance to learn multiplication using proper positional embeddings such as relative or abacus PE. In fact, the structured nature of multiplication, especially when broken down into steps that involve digit-by-digit operations and carry propagation, aligns well with the capabilities of Transformer models to capture sequential dependencies and patterns. However, the most challenging aspect is computing  the raw sums $c_i^R$ at each position. Each \(c_i^{R}\) results from a sum of specific pairs of digits from the input sequences $a$ and $b$. For a given \(c_i^{R}\), the valid pairs \((i', j')\) must satisfy \(i' + j' = i + 1\). Identifying these pairs involves that (1) ensuring \(1 \leq i', j' \leq n\), i.e., the indices must be within the bounds of the sequences. (2) For each \(i\), determining which pairs contribute to \(c_i^{R}\) involves iterating through potential values of \(i'\) and \(j'\) and checking if their sum equals \(i + 1\). Digit multiplication depends on the positional significance of digits. Misalignment in positions can lead to incorrect contributions to the product. Therefore, positional encoding and accurate handling of positional values are necessary to ensure correct multiplication results. There are also efficiency considerations. Multiplication of large numbers involves many such sums. For large \(n\), directly computing \(c_i^{R}\) for each \(i\) involves nested loops or checks, leading to a time complexity of \(O(n^2)\) in the worst case. This poses a great difficulty for computing the raw sum $c_i^R$.

This challenge can be understood through the following analysis. Suppose the model is provided with Chain-of-Thought (CoT) style intermediate steps of multiplication as part of the training data. The CoT-like training data format is:
\[ a \times b \to (c^{R}, c^{\chi}) \to c. \]
In digit-wise format, this is:
\begin{equation*}
\begin{split}
& \quad (a_1, \cdots, a_n) \times (b_1, \cdots, b_n) \\
& \to (c_1^{R}, c_1^{\chi}, \cdots, c_{2n-1}^{R}, c_{2n-1}^{\chi}) \\
& \to (c_1, \cdots, c_{2n}).
\end{split}
\end{equation*}
The conditional probability equation is then given by:
\begin{equation*}
\begin{split}
&\quad \mathbf{P}_{\theta}(c_i \mid a_1, \cdots, a_{i \wedge n}, b_1, \cdots, b_{i \wedge n}) \\
& = \mathbf{P}_{\theta}^{\chi}(c_{i-1}^{\chi} \mid a_1, \cdots, a_{(i-1) \wedge n}, b_1, \cdots, b_{(i-1) \wedge n}) \\
& \ \times \mathbf{P}_{\theta}^{R}(c_i^{R} \mid a_1, \cdots, a_{i \wedge n}, b_1, \cdots, b_{i \wedge n}) \\
& \ \times \mathbf{P}_{\theta}(c_i \mid c_i^{R}, c_{i-1}^{\chi}),
\end{split}
\end{equation*}
and
\begin{equation*}
\begin{split}
& \quad \mathbf{P}_{\theta}^{\chi}(c_i^{\chi} \mid a_1, \cdots, a_{i \wedge n}, b_1, \cdots, b_{i \wedge n}) \\
& = \mathbf{P}_{\theta}^{\chi}(c_{i-1}^{\chi} \mid a_1, \cdots, a_{(i-1) \wedge n}, b_1, \cdots, b_{(i-1) \wedge n}) \\
& \ \times \mathbf{P}_{\theta}^{R}(c_i^{R} \mid a_1, \cdots, a_{i \wedge n}, b_1, \cdots, b_{i \wedge n}) \\
& \ \times \mathbf{P}_{\theta}^{\chi}(c_i^{\chi} \mid c_i^{R}, c_{i-1}^{\chi}).
\end{split}
\end{equation*}
For the carry at the $i$-th position, we then have that 
\begin{equation*}
\begin{split}
&\quad \mathbf{P}_{\theta}^{\chi}(c_i^{\chi} \mid a_1, \cdots, a_{i \wedge n}, b_1, \cdots, b_{i \wedge n}) \\
&= \prod_{j=1}^{i} \mathbf{P}_{\theta}^{R}(c_j^{R} \mid a_1, \cdots, a_{j \wedge n}, b_1, \cdots, b_{j \wedge n}) \\
&\ \times \mathbf{P}_{\theta}^{\chi}(c_j^{\chi} \mid c_j^{R}, c_{j-1}^{\chi}).
\end{split}
\end{equation*}
Note that \(\mathbf{P}_{\theta}(c_i \mid c_i^{R}, c_{i-1}^{\chi})\) and \(\mathbf{P}_{\theta}^{\chi}(c_i^{\chi} \mid c_i^{R}, c_{i-1}^{\chi})\) exhibit transition invariance. This could be handled by relative or abacus positional embedding. The difficulty lies in the computation of the raw sums \(\mathbf{P}_{\theta}^{R}(c_i^{R} \mid a_1, \cdots, a_{i \wedge n}, b_1, \cdots, b_{i \wedge n})\) even when using relative or abacus positional embedding.

Experiments on Transformer models using relative or abacus positional embeddings to learn multiplication have been presented in the literature. \citet{jelassi2023length} and \citet{mcleish2024transformers} show that addition can successfully generalize to OOD regions with higher numerical digits, but multiplication has largely not succeeded. Our analysis provides insights into the difficulties behind generalizing to higher numerical digits, which helps us understand the reasons for the failure in learning multiplication.

\section{Theoretical OOD Test Accuracy for Modular Arithmetic}\label{appendix_accuracy}

\subsection{Theoretical OOD Test Accuracy for Modular Addition Learning}\label{appendix_accuracy_mod_add}

To derive an accurate analytic formula (in Theorem~\ref{thm:modular-addition-nondiv}) for the OOD test accuracy on $\widetilde{\mathcal{D}}_{m}$ with $m>n$ when a Transformer model is trained on the domain $\mathcal{D}_n$, we must carefully count the valid pairs \((a, b)\in \widetilde{\mathcal{D}}_{m}\) that satisfy $\overline{a+b}^p=\overline{\overline{a}^{10^n}+\overline{b}^{10^n}}^p$.

Let \(a = A \cdot 10^n + a_0\) and \(b = B \cdot 10^n + b_0\), where \(A, B\) range from 1 to \(10^{m-n} - 1\) and \(a_0, b_0\) range from 0 to \(10^n - 1\). We require \(a + b \equiv (a \mod 10^n + b \mod 10^n) \pmod{p}\), which simplifies to that \[(A + B) \cdot 10^n \equiv 0 \pmod{p}.\] Let \(p' = \frac{p}{\gcd(p, 10^n)}\). We are then left with the condition \((A + B) \equiv 0 \pmod{p'}\).

The number of such pairs is determined by the frequency of multiples of \(p'\) in the valid range. The total number of pairs \((A, B)\) is \((10^{m-n} - 1)^2\). There are \((10^{m-n} - 1)\) valid values for \(A\). For each \(A\), the number of valid \(B\) values is determined by the number of multiples of \(p'\) in the range. That is, for each \(A\), the number of valid \(B\) values is about \( (10^{m-n} - 1) / p' \). The test accuracy is the ratio of valid pairs, i.e. the number of valid pairs divided by the total number of pairs. 

Note that for \(m \geq n + \log_{10}(p'/2 + 1)\), the range \(1 \leq A, B < 10^{m-n}\) must include at least one complete cycle of \(p'\) to ensure some pairs \((A, B)\) satisfy \(A + B \equiv 0 \pmod{p'}\). This condition ensures that the number of digits in \(A\) and \(B\) is large enough to cover a full period of \(p'\). Otherwise, there exists no pair $(A,B)$ for which \(A + B \equiv 0 \pmod{p'}\).

The ultimate formula is as follows: 
\begin{equation*}
\begin{split}
\operatorname{Acc}(p,n,m) & = \frac{\text{Number of Valid Pairs}}{\text{Total Number of Pairs}} \\
& \approx \frac{(10^{m-n} - 1) \cdot \left( \frac{10^{m-n} - 1}{p'} \right)}{(10^{m-n} - 1)^2} = \frac{1}{p'}
\end{split}
\end{equation*}
for \( m \geq n + \log_{10}(p'/2 + 1) \), otherwise 0. 

Given that \(p' = \frac{p}{\gcd(p, 10^n)}\), we have that 
\begin{equation*}
\begin{split}
&\quad \operatorname{Acc}(p,n,m) \\
& \approx 
\begin{cases} 
\frac{\gcd(p, 10^n)}{p}, & \text{if } m \geq n + \log_{10}(p'/2 + 1) \\
0, & \text{otherwise}
\end{cases}.
\end{split}
\end{equation*}

\subsection{Theoretical OOD Test Accuracy for Modular Multiplication Learning}\label{appendix_accuracy_mod_multip}

To count the valid pairs \((a, b) \in \widetilde{\mathcal{D}}_{m}\) that satisfy \(a \times b \equiv ((a \mod 10^n) \times (b \mod 10^n)) \pmod{p}\), 
denote \(a\) and \(b\) can be written as \(a = A \cdot 10^n + a_0\) and \(b = B \cdot 10^n + b_0\), where \(A, B\) are the upper \((m-n)\)-digit parts and \(a_0, b_0\) are the lower \(n\)-digit parts. \(A, B\) range from 1 to \(10^{m-n} - 1\) (since they are non-zero leading digits). \(a_0, b_0\) range from 0 to \(10^n - 1\). We need 
\[ (A \cdot 10^n + a_0) \times (B \cdot 10^n + b_0) \equiv (a_0 \times b_0) \pmod{p}. \]
This simplifies to that
\[ A \cdot B \cdot 10^{2n} + (A \cdot b_0 + B \cdot a_0) \cdot 10^n \equiv 0 \pmod{p}. \]
This further simplifies to that 
\[ A \cdot B \cdot 10^n + A \cdot b_0 + B \cdot a_0 \equiv 0 \pmod{p'}, \]
\[ p'=\frac{p}{\gcd(p, 10^n)}. \]

The theoretical closed expression for this problem is challenging to derive, but the numerical solution can be computed through an algorithmic program for small-scale cases.

\section{Model and Training Hyperparameters}\label{training_hyper_config}

We follow the architecture at \url{https://github.com/karpathy/nanoGPT} and \url{https://github.com/karpathy/minGPT} for building our research code, which is compliant with the MIT license that the github repositories are under.

Detailed hyperparameters of the models and training are provided in Table~\ref{table:hyperparameters}.

\begin{table}[ht]
\small
\begin{centering}
\begin{tabular}{>{\centering}m{2.2cm}>{\centering}m{1.5cm}>{\centering}m{1.5cm}>{\centering}m{1.5cm}}
\toprule 
\centering{}\textbf{Hyperparameter} & \centering{}\textbf{NanoGPT} & \centering{}\textbf{MicroGPT} & \textbf{MiniGPT}\tabularnewline
\midrule
num layer & 3 & 4 & 6\tabularnewline
num head & 3 & 4 & 6\tabularnewline
dim embd  & 48 & 128 & 384\tabularnewline
vocab size & 16 & 16 & 16\tabularnewline
context window & 256 & 256 & 256\tabularnewline
dropout prob & 0.2 & 0.2 & 0.2\tabularnewline
optimizer & AdamW & AdamW & AdamW\tabularnewline
learning rate & 0.001 & 0.001 & 0.001\tabularnewline
betas & (0.9, 0.99) & (0.9, 0.99) & (0.9, 0.99)\tabularnewline
weight decay & True & True & True \tabularnewline
grad norm clip & 1.0 & 1.0 & 1.0\tabularnewline
\bottomrule
\end{tabular}
\par\end{centering}
\caption{Hyperparameter for Arithmetic Operations Training}
\label{table:hyperparameters}
\end{table}

\section{Training DataSet Parameter}\label{appendix_data}

\begin{table}[ht]
\small
\begin{centering}
\begin{tabular}{>{\centering}m{2.0cm}>{\centering}m{2.0cm}>{\centering}m{2.0cm}}
\toprule 
\centering{}\textbf{DataSet} & \centering{}\textbf{Description} & \centering{}\textbf{Obs.Num} \tabularnewline
\midrule
$\mathcal{D}_{4}$ & $n=4, m=4$ & 100,000 \tabularnewline
$\mathcal{D}_{5}$ & $n=5, m=5$ & 100,000 \tabularnewline
$\mathcal{D}_{6}$ & $n=6, m=6$ & 100,000 \tabularnewline
$\mathcal{D}_{7}$ & $n=7, m=7$ & 100,000 \tabularnewline
$\mathcal{D}_{4,5}$ & .5$D_4$+.5$D_5$ & 100,000 \tabularnewline
$\mathcal{D}_{5,6}$ & .5$D_5$+.5$D_6$ & 100,000 \tabularnewline
$\mathcal{D}_{6,7}$ & .6$D_6$+.6$D_7$ & 120,000 \tabularnewline
\bottomrule
\end{tabular}
\par\end{centering}
\caption{Data Scale: Training Data (Cont.)}
\label{table:data_size_full}
\end{table}

\onecolumn

\section{Further Results}\label{appendix_results}

\subsection{Further Results on Addition}\label{appendix_results_addition}

% Standard Addition
\begin{figure}[htbp]
\centering
\begin{minipage}{0.85\textwidth}
\centering
\begin{subfigure}{0.45\textwidth}
  \includegraphics[width=\textwidth,height=0.75\textwidth]{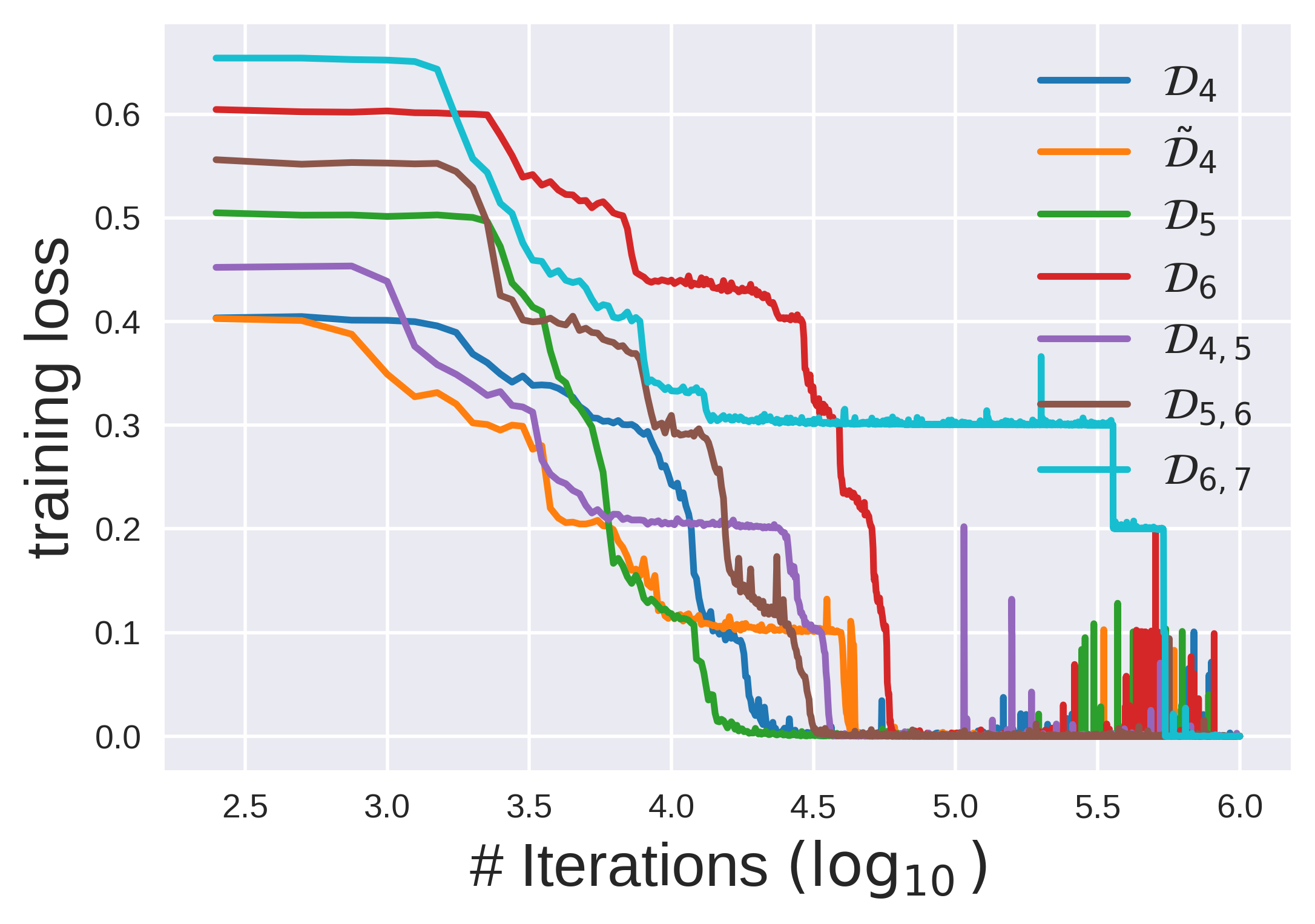}
  \caption*{Training Loss}
\end{subfigure}
\hfill
\begin{subfigure}{0.45\textwidth}
  \includegraphics[width=\textwidth,height=0.75\textwidth]{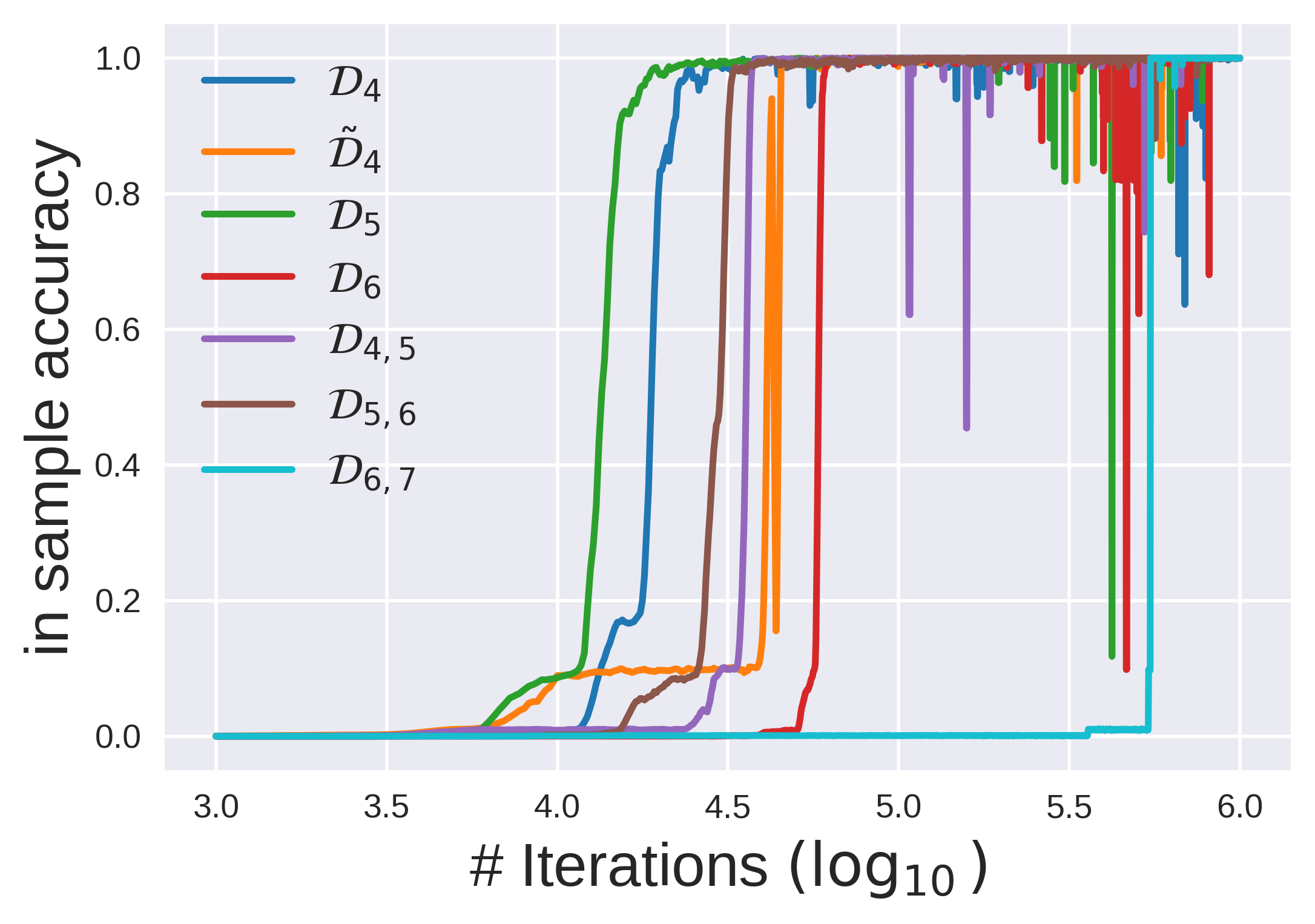}
  \caption*{In Sample Accuracy}
\end{subfigure}
\end{minipage}
\caption{Training Loss \& Out of Sample In-Distribution Test Accuracy on Addition}
\caption*{\textit{Note:} $\mathcal{D}_i$ is trained on two number addition task with at least one number to be a $i$-digit number, $\mathcal{D}_{i,j}$ is trained on the combined training dataset of $\mathcal{D}_i$ and $\mathcal{D}_j$.}
\end{figure}

% Robustness Studies
\begin{figure}[htbp]
\centering
\begin{minipage}{0.85\textwidth}
\centering
\begin{subfigure}{0.45\textwidth}
  \includegraphics[width=\textwidth,height=0.75\textwidth]{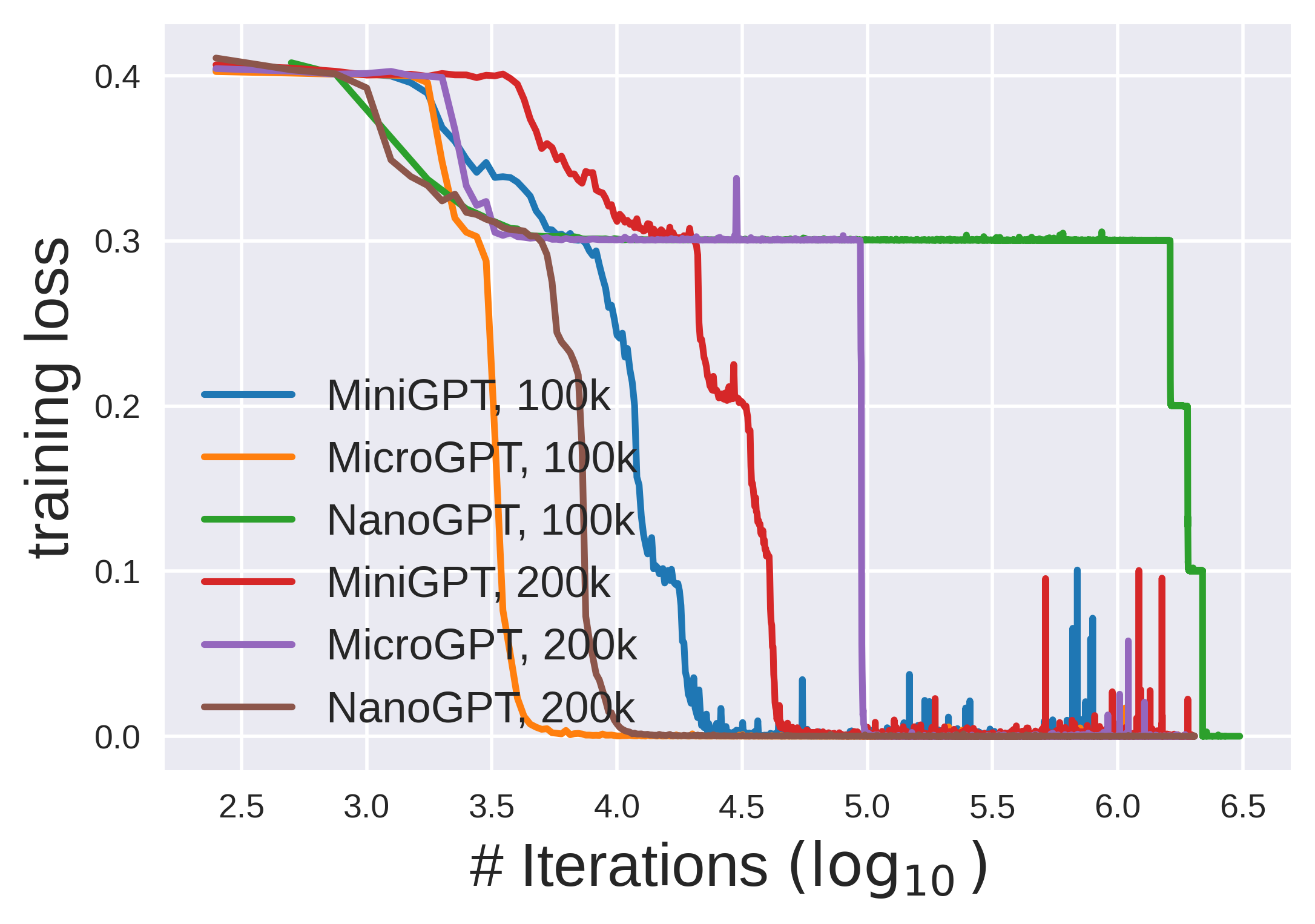}
  \caption*{Training Loss}
\end{subfigure}
\hfill
\begin{subfigure}{0.45\textwidth}
  \includegraphics[width=\textwidth,height=0.75\textwidth]{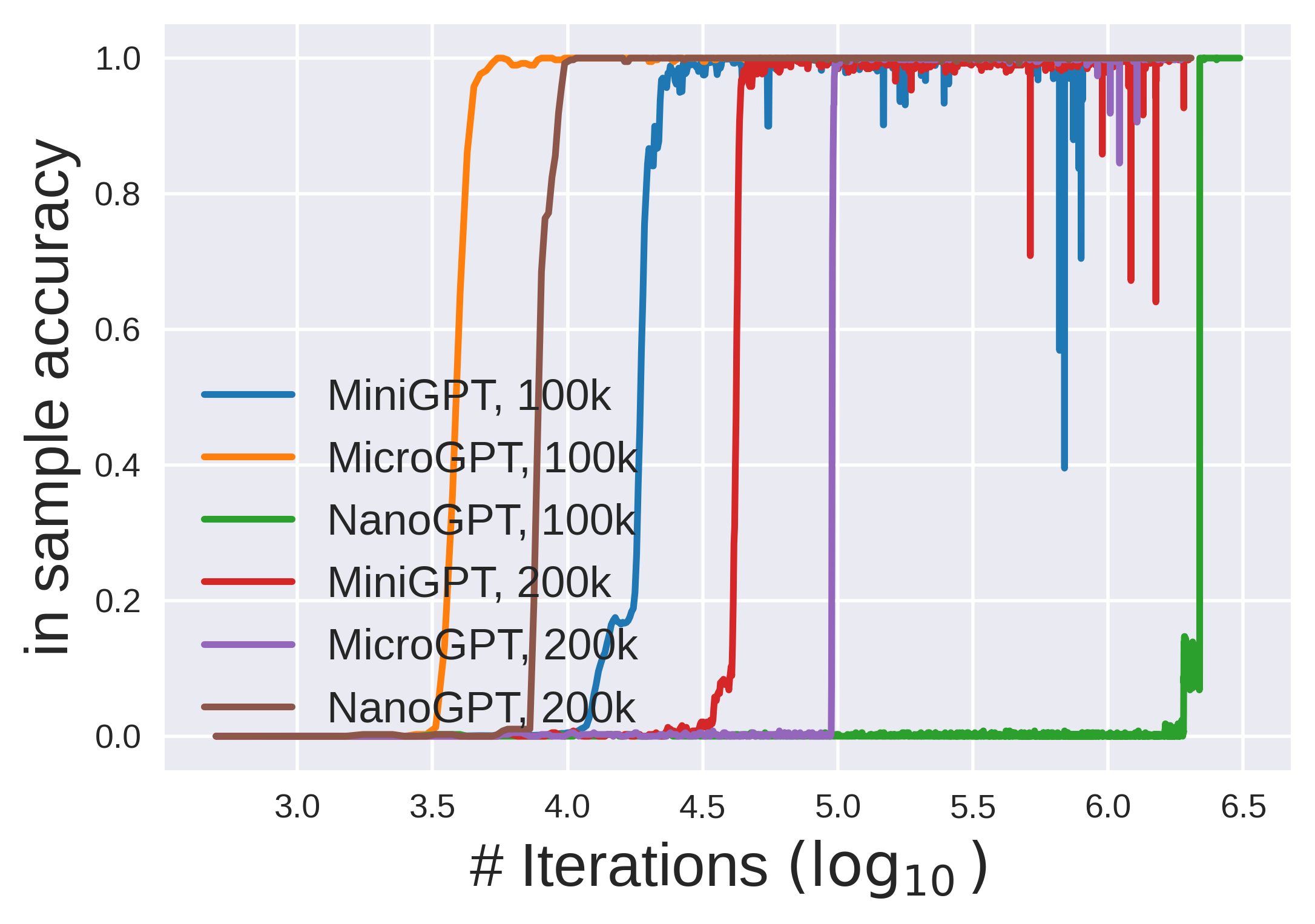}
  \caption*{In Sample Accuracy}
\end{subfigure}
\end{minipage}
\caption{Training Loss \& Out of Sample In-Distribution Test Accuracy on Addition}
\caption*{\textit{Note:} Robustness study on model and data scales. All models are trained on $\mathcal{D}_4$ where $a$ and $b$ are at least one to be a $4$-digit number. NanoGPT represents the smallest model, with MicroGPT being of medium size and MiniGPT the largest. The designations "100k" and "200k" indicate that the training sets are 90\% the size of 100,000 or 200,000, respectively.
}
\end{figure}

% Standard Addition: Ground Truth Accuracy
\begin{table*}[ht]
\centering
\begin{tabular}{lrrrrrrrrr}
\toprule
\multicolumn{1}{c}{} & \multicolumn{9}{c}{Test Accuracy (\%) w.r.t. the Ground Truth on the Domain $\mathcal{D}_i$} \\
Training Data & 1 & 2 & 3 & 4 & 5 & 6 & 7 & 8 & 9\\
\midrule
$\mathcal{D}_{4}$ & 100 & 100 & 100 & 100 & 0 & 0 & 0 & 0 & 0 \\
$\widetilde{\mathcal{D}}_{4}$ & 100 & 100 & 72.6 & 100 & 0 & 0 & 0 & 0 & 0 \\
$\mathcal{D}_{5}$ & 100 & 100 & 100 & 100 & 100 & 0 & 0 & 0 & 0 \\
$\mathcal{D}_{6}$ & 100 & 100 & 100 & 100 & 100 & 100 & 0 & 0 & 0 \\
$\mathcal{D}_{4,5}$ & 100 & 100 & 100 & 100 & 100 & 0 & 0 & 0 & 0 \\
$\mathcal{D}_{5,6}$ & 100 & 100 & 100 & 100 & 100 & 100 & 0 & 0 & 0 \\
$\mathcal{D}_{6,7}$ & 100 & 100 & 100 & 100 & 100 & 100 & 100 & 0 & 0 \\
\bottomrule
\end{tabular}
\caption{Standard Addition: Test Accuracy w.r.t. the Ground Truth $f(a,b)=a+b$ on the Domain $\mathcal{D}_i$ for $i=1,2\cdots,9$. All models are instances of MiniGPT. The accuracy is tested on 10,000 random test samples (when \( n > 2 \)), otherwise on the entire dataset. The outputs of models are generated using maximum probability sampling.}
\label{table:addition_true_acc}
\end{table*}

% Standard Addition: Modular Truth Accuracy
\begin{table*}[ht]
\centering
\begin{tabular}{lrrrrrrrrr}
\toprule
\multicolumn{1}{c}{} & \multicolumn{9}{c}{Test Accuracy (\%) w.r.t. the Modular Truth on the Domain $\mathcal{D}_i$} \\
Training Data & 1 & 2 & 3 & 4 & 5 & 6 & 7 & 8 & 9\\
\midrule
$\mathcal{D}_{4}$ & 100 & 100 & 100 & 100 & 100 & 100 & 100 & 100 & 100 \\
$\widetilde{\mathcal{D}}_{4}$ & 100 & 99.9 & 72.3 & 100 & 99.7 & 99.7 & 99.6 & 99.7 & 99.5 \\
$\mathcal{D}_{5}$ & 100 & 100 & 100 & 100 & 100 & 100 & 100 & 100 & 100 \\
$\mathcal{D}_{6}$ & 100 & 100 & 100 & 100 & 100 & 100 & 100 & 100 & 100 \\
$\mathcal{D}_{4,5}$ & 100 & 100 & 100 & 100 & 100 & 100 & 100 & 100 & 100 \\
$\mathcal{D}_{5,6}$ & 100 & 100 & 100 & 100 & 100 & 100 & 100 & 100 & 100 \\
$\mathcal{D}_{6,7}$ & 100 & 100 & 100 & 100 & 100 & 100 & 100 & 100 & 100 \\
\bottomrule
\end{tabular}
\caption{Standard Addition: Test Accuracy w.r.t. the Modular Truth $\hat{f}(a,b)=\overline{a}^{10^n}+\overline{b}^{10^n}$ on the Domain $\mathcal{D}_i$ for $i=1,2\cdots,9$. All models are instances of MiniGPT, and test methods are indicated as above.}
\label{table:addition_modular_acc}
\end{table*}

% Digit-wise accuracy of Addition
\begin{figure*}[htbp]
\centering
\begin{minipage}{0.85\textwidth}
\centering
\begin{subfigure}{0.3\textwidth}
  \includegraphics[width=\textwidth,height=0.75\textwidth]{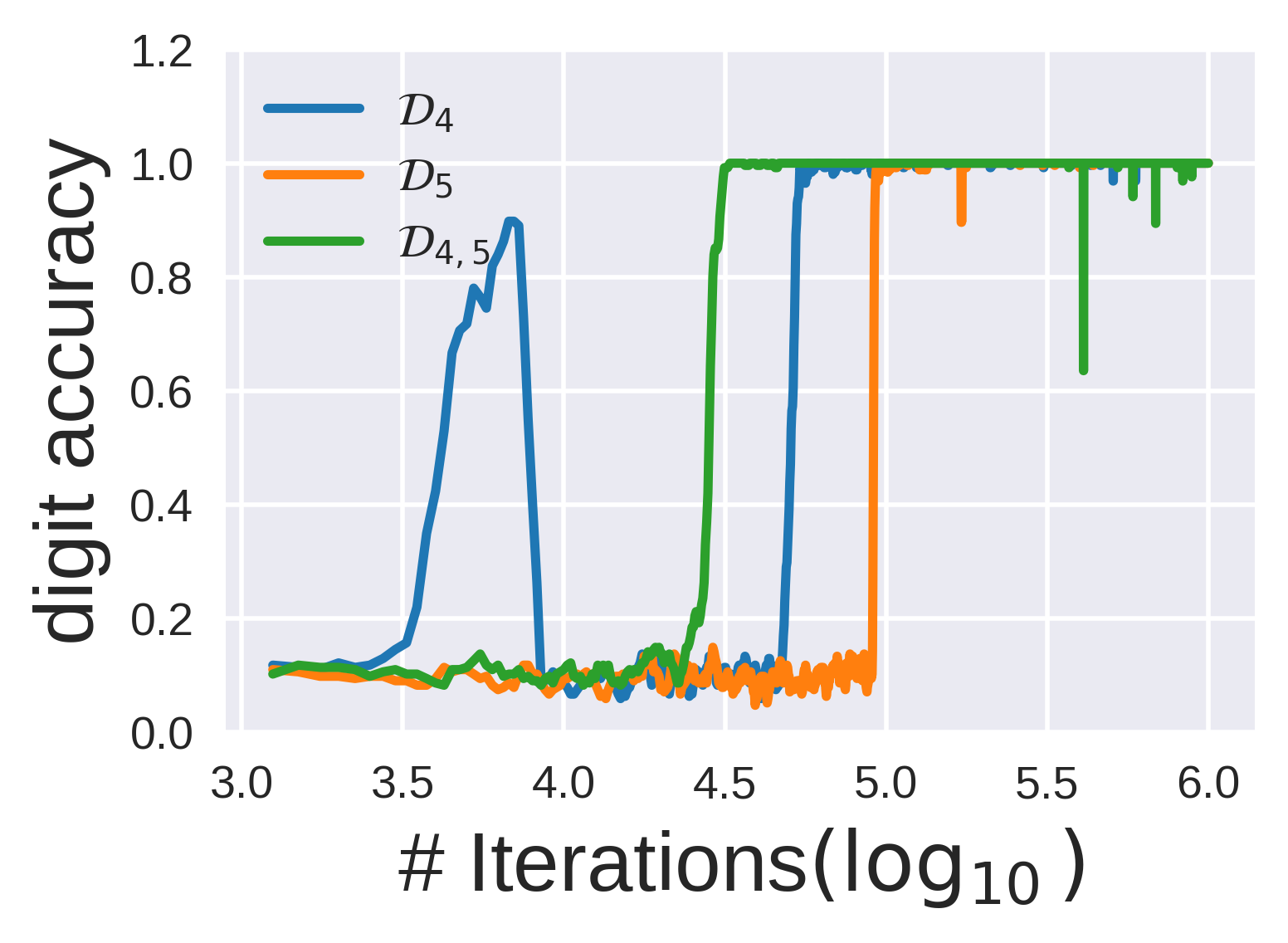}
  \caption*{1st digit}
\end{subfigure}
\hfill
\begin{subfigure}{0.3\textwidth}
  \includegraphics[width=\textwidth,height=0.75\textwidth]{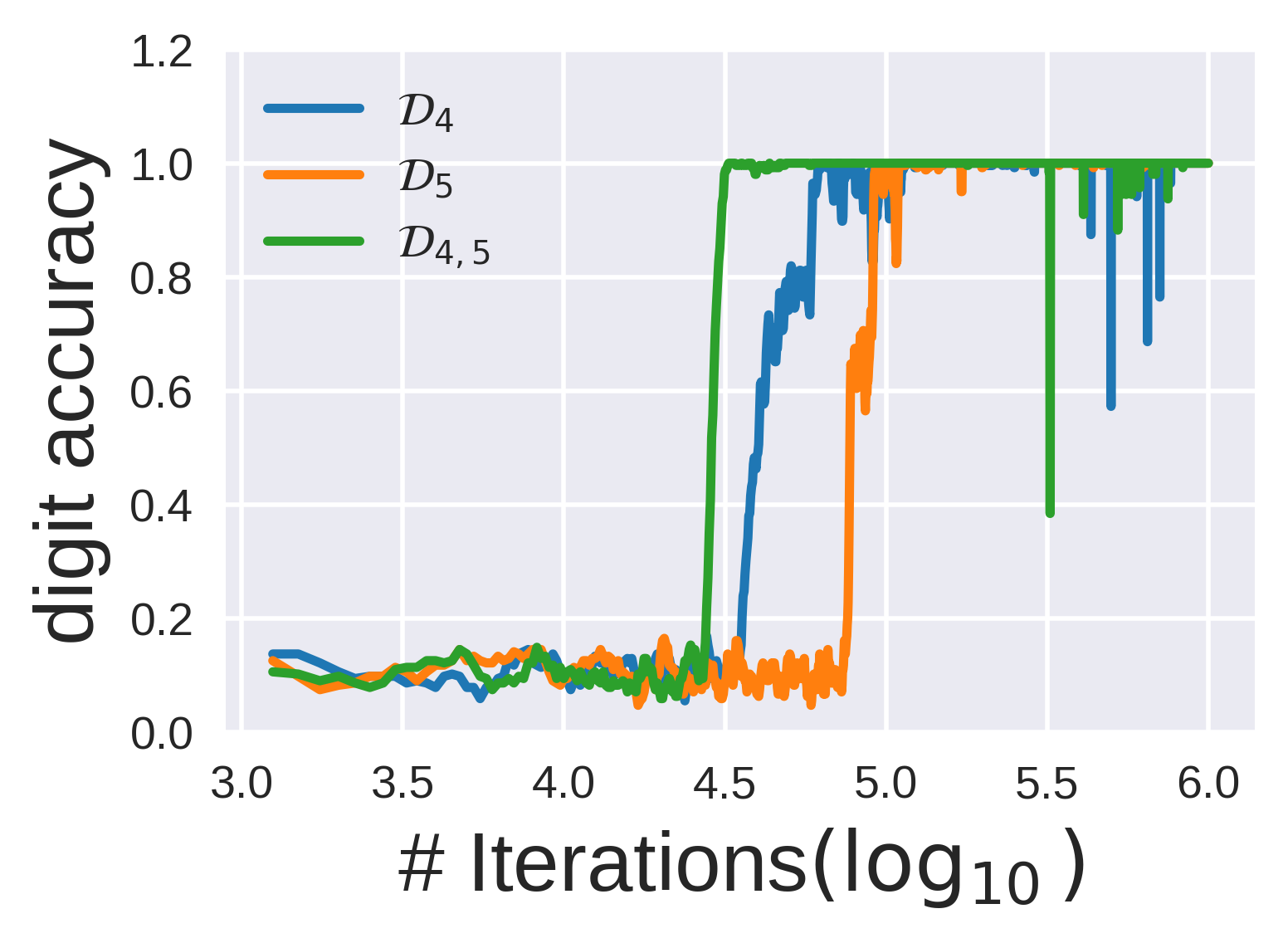}
  \caption*{2nd digit}
\end{subfigure}
\hfill
\begin{subfigure}{0.3\textwidth}
  \includegraphics[width=\textwidth,height=0.75\textwidth]{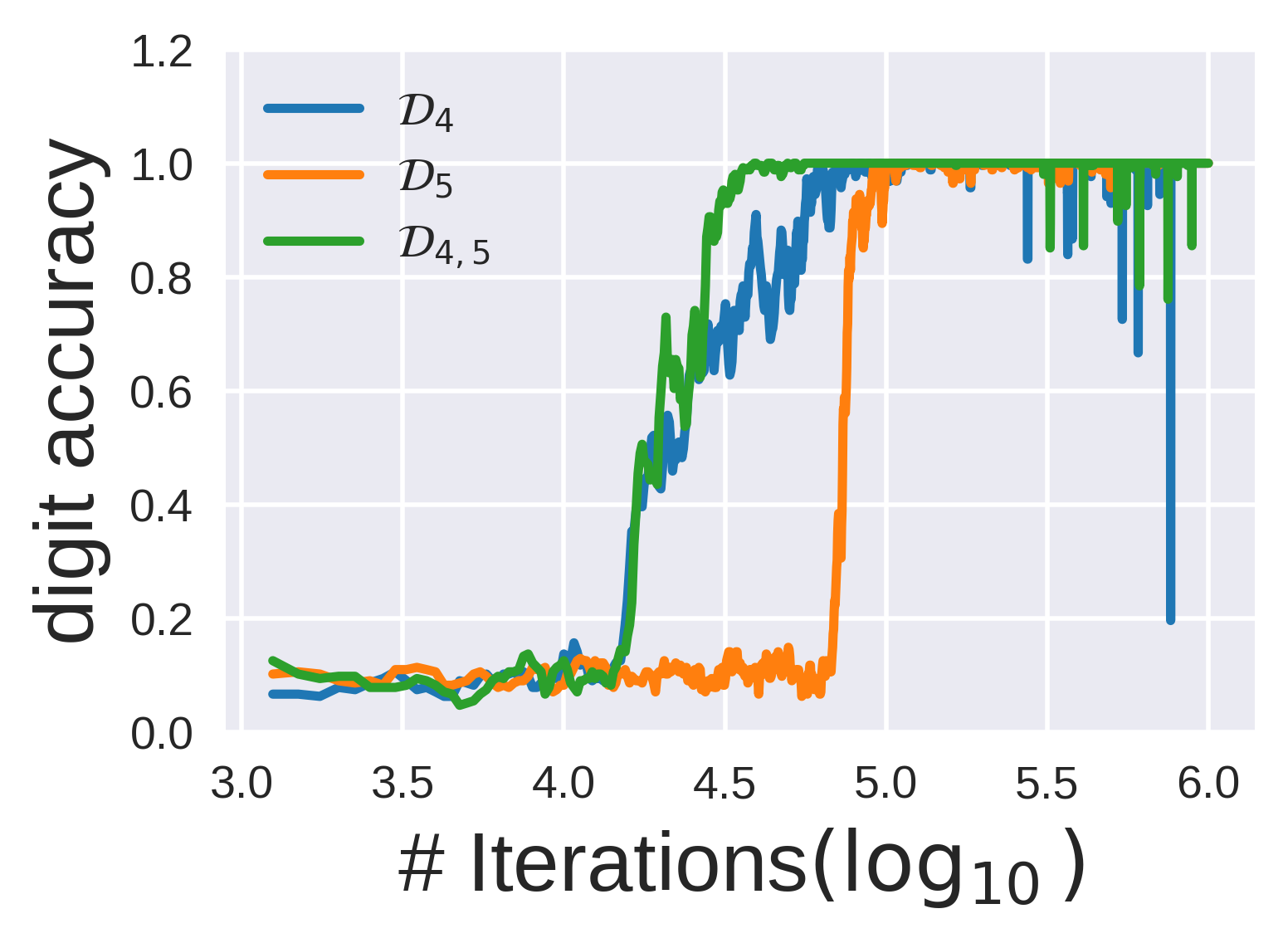}
  \caption*{3rd digit}
\end{subfigure}
\\
\begin{subfigure}{0.3\textwidth}
  \includegraphics[width=\textwidth,height=0.75\textwidth]{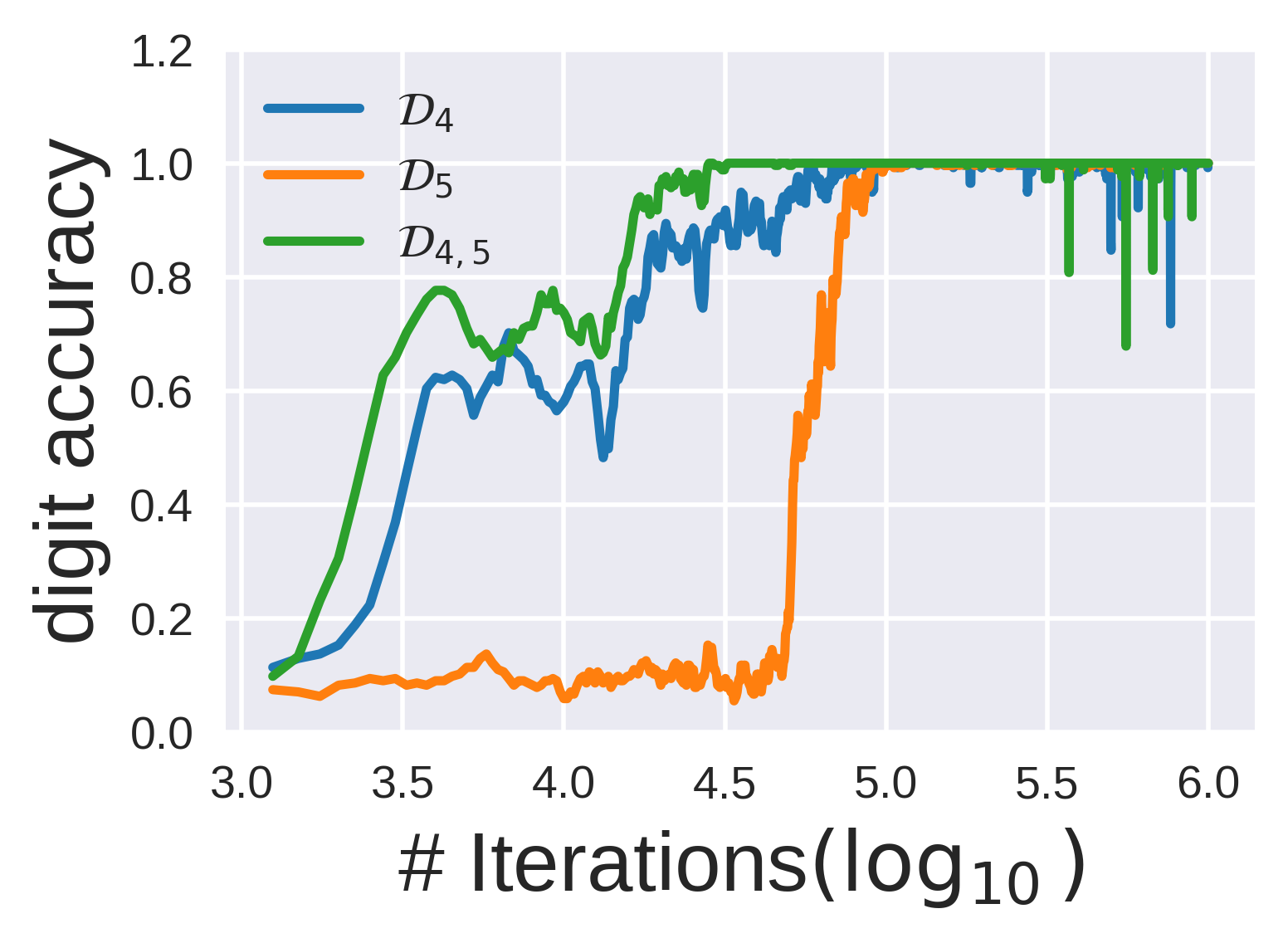}
  \caption*{4th digit}
\end{subfigure}
\hfill
\begin{subfigure}{0.3\textwidth}
  \includegraphics[width=\textwidth,height=0.75\textwidth]{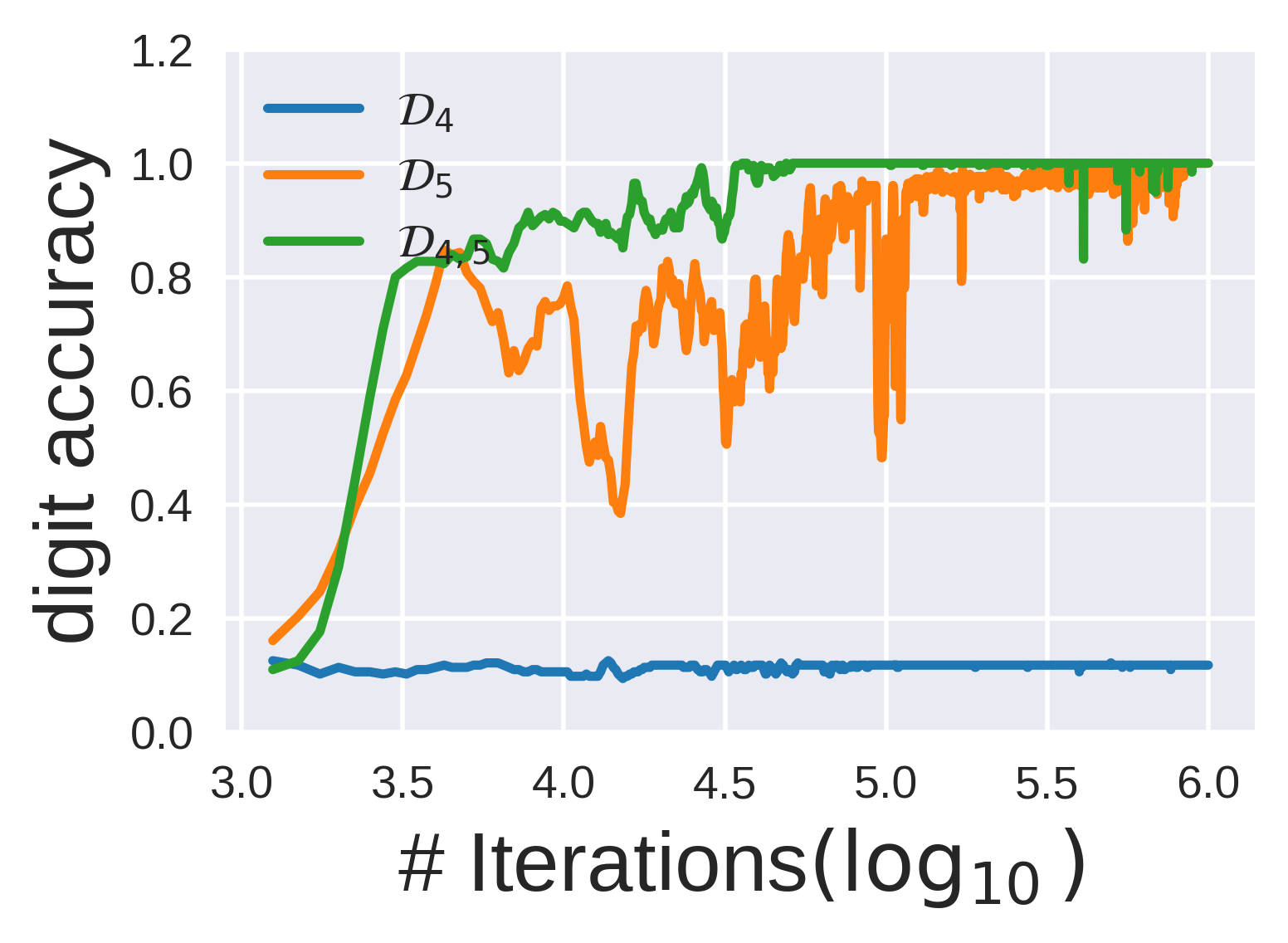}
  \caption*{5th digit}
\end{subfigure}
\hfill
\begin{subfigure}{0.3\textwidth}
  \includegraphics[width=\textwidth,height=0.75\textwidth]{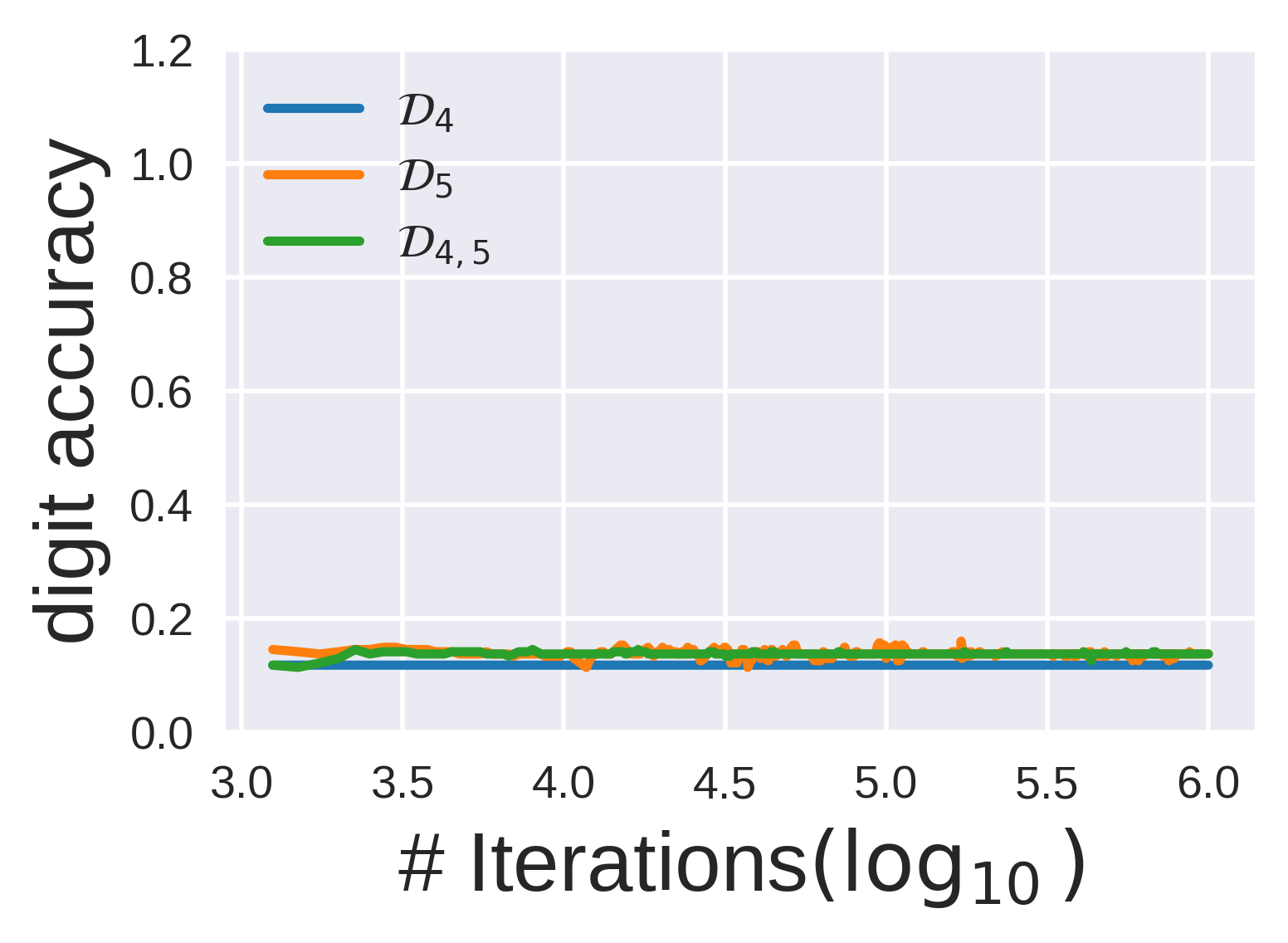}
  \caption*{6th digit}
\end{subfigure}
\\
\begin{subfigure}{0.3\textwidth}
  \includegraphics[width=\textwidth,height=0.75\textwidth]{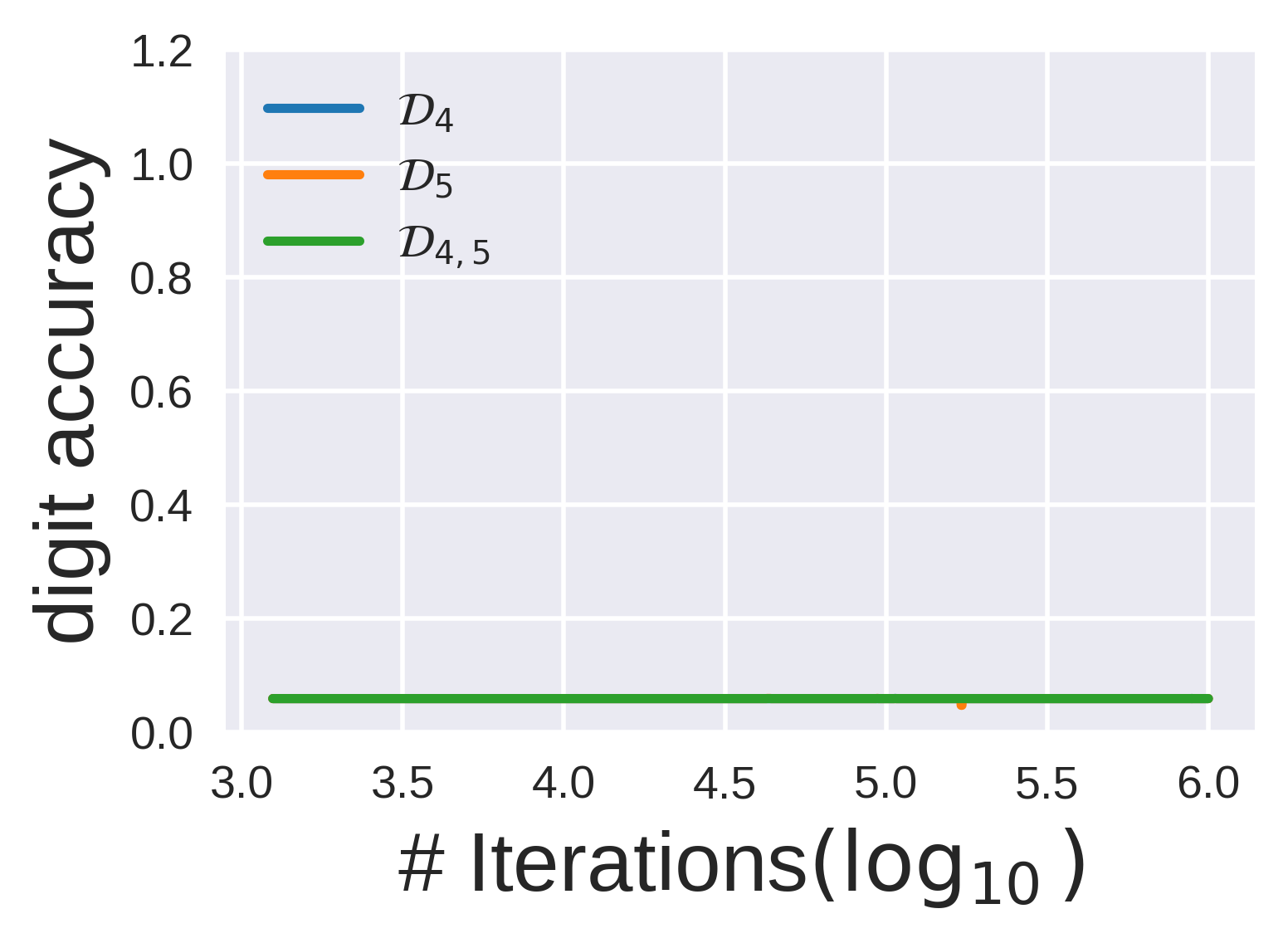}
  \caption*{7th digit}
\end{subfigure}
\hfill
\begin{subfigure}{0.3\textwidth}
  \includegraphics[width=\textwidth,height=0.75\textwidth]{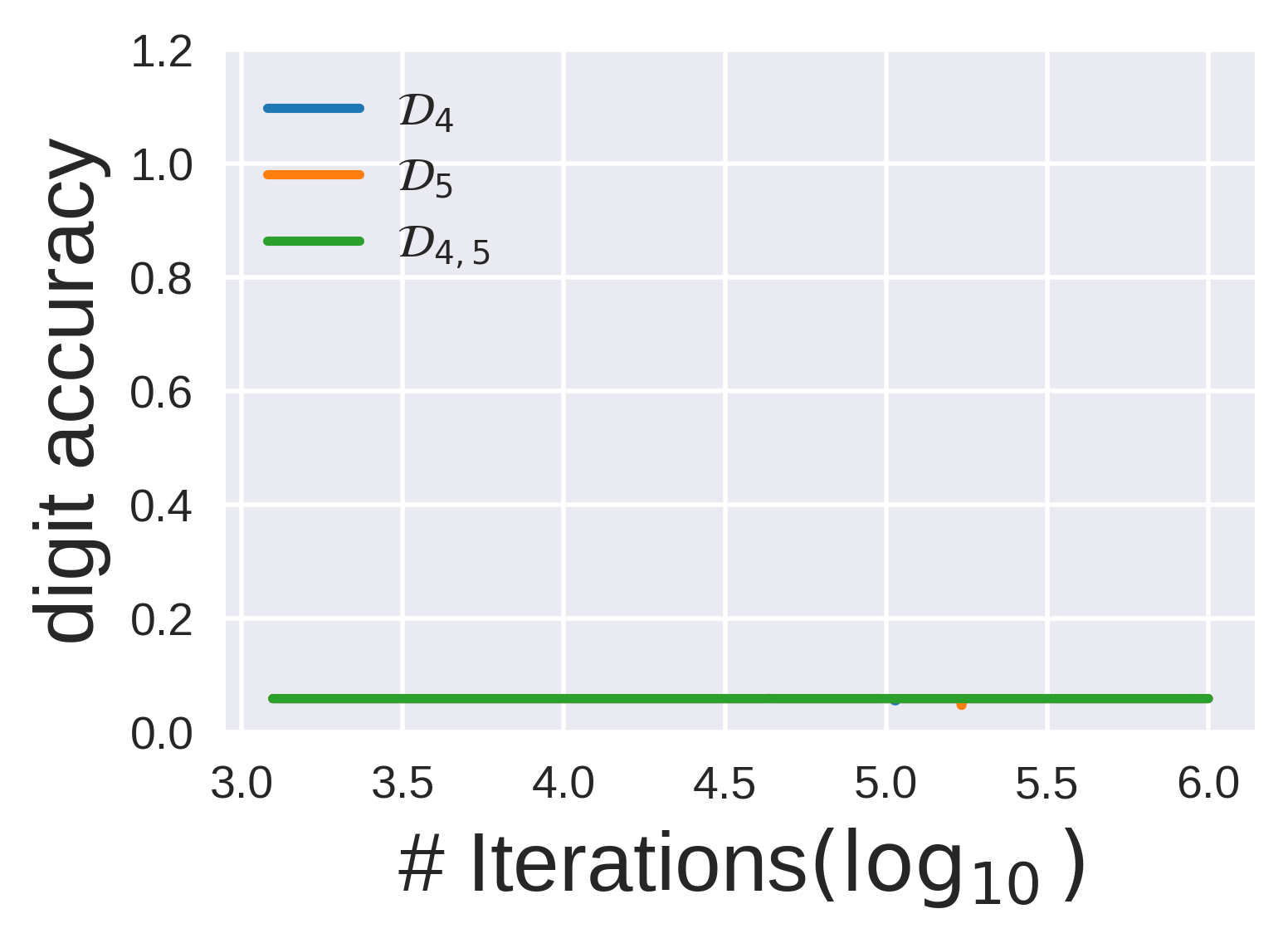}
  \caption*{8th digit}
\end{subfigure}
\hfill
\begin{subfigure}{0.3\textwidth}
  \includegraphics[width=\textwidth,height=0.75\textwidth]{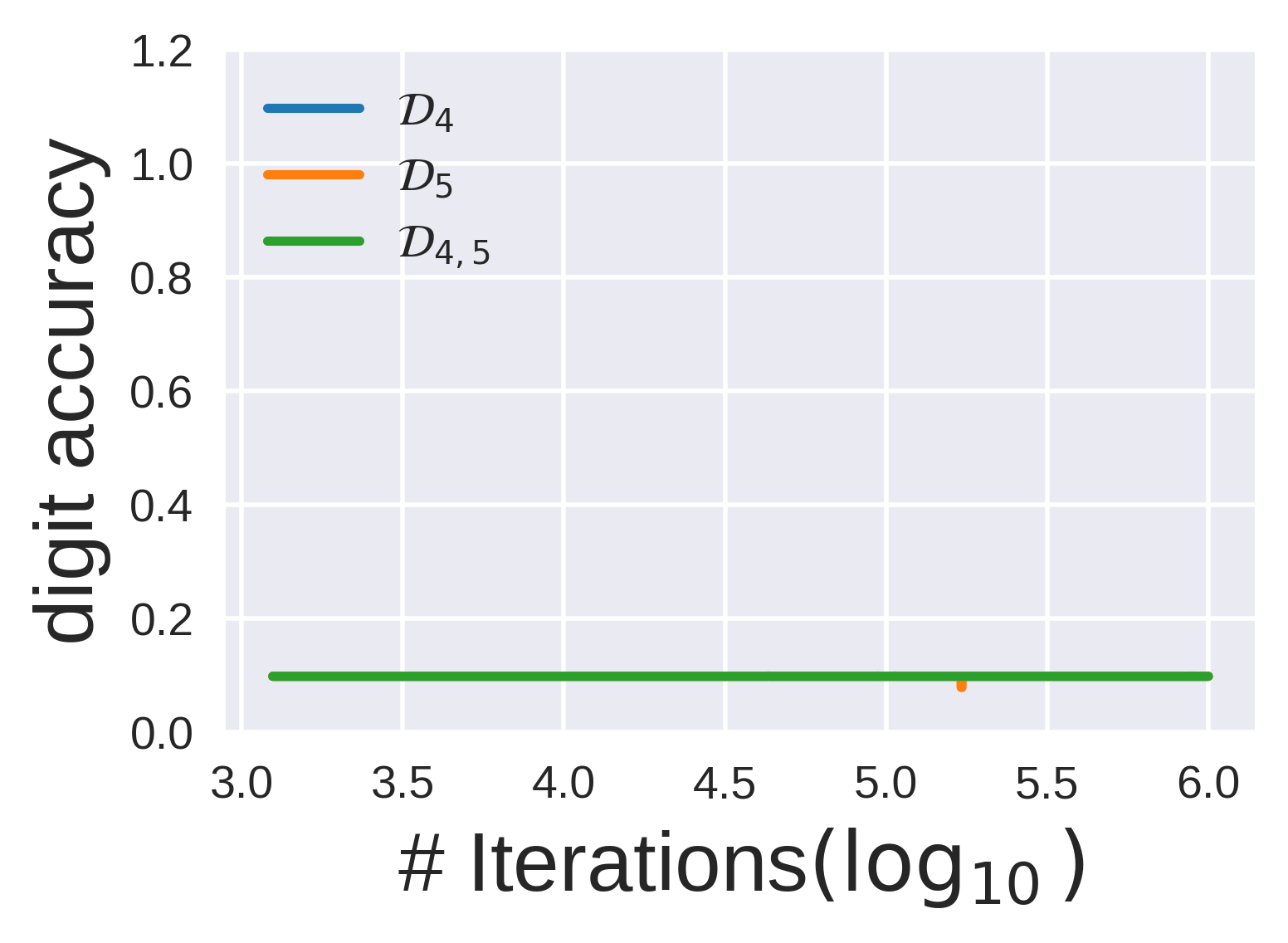}
  \caption*{9th digit}
\end{subfigure}
\end{minipage}
\caption{Digit-Wise Test Accuracy of Transformer Models with APE for Addition Tasks}
\caption*{\textit{Note:} In this figure, we present the results of three different experiments using distinct training datasets. For all experiments, we employ the MiniGPT model equipped with a learned APE. In the legend, the label $\mathcal{D}_4$ indicates that the MiniGPT model is trained on a random sample from dataset $\mathcal{D}_4$. The label $\mathcal{D}_5$ denotes training on a random sample from dataset $\mathcal{D}_5$, while $\mathcal{D}_{4,5}$ signifies training on a combined subset from $\mathcal{D}_4$ and $\mathcal{D}_5$. Each subfigure illustrates the digit-wise test accuracy on a combined random sample sets $\mathcal{D}_{\leq 9}$ for these models throughout the training process.}
\label{fig:addition_acc_digits}
\end{figure*}

\subsubsection{How Digits are Learned During Training?}

\begin{figure*}[ht]
\centering
\begin{minipage}{0.85\textwidth}
\centering
\begin{subfigure}{0.3\textwidth}
  \includegraphics[width=\textwidth,height=0.75\textwidth]{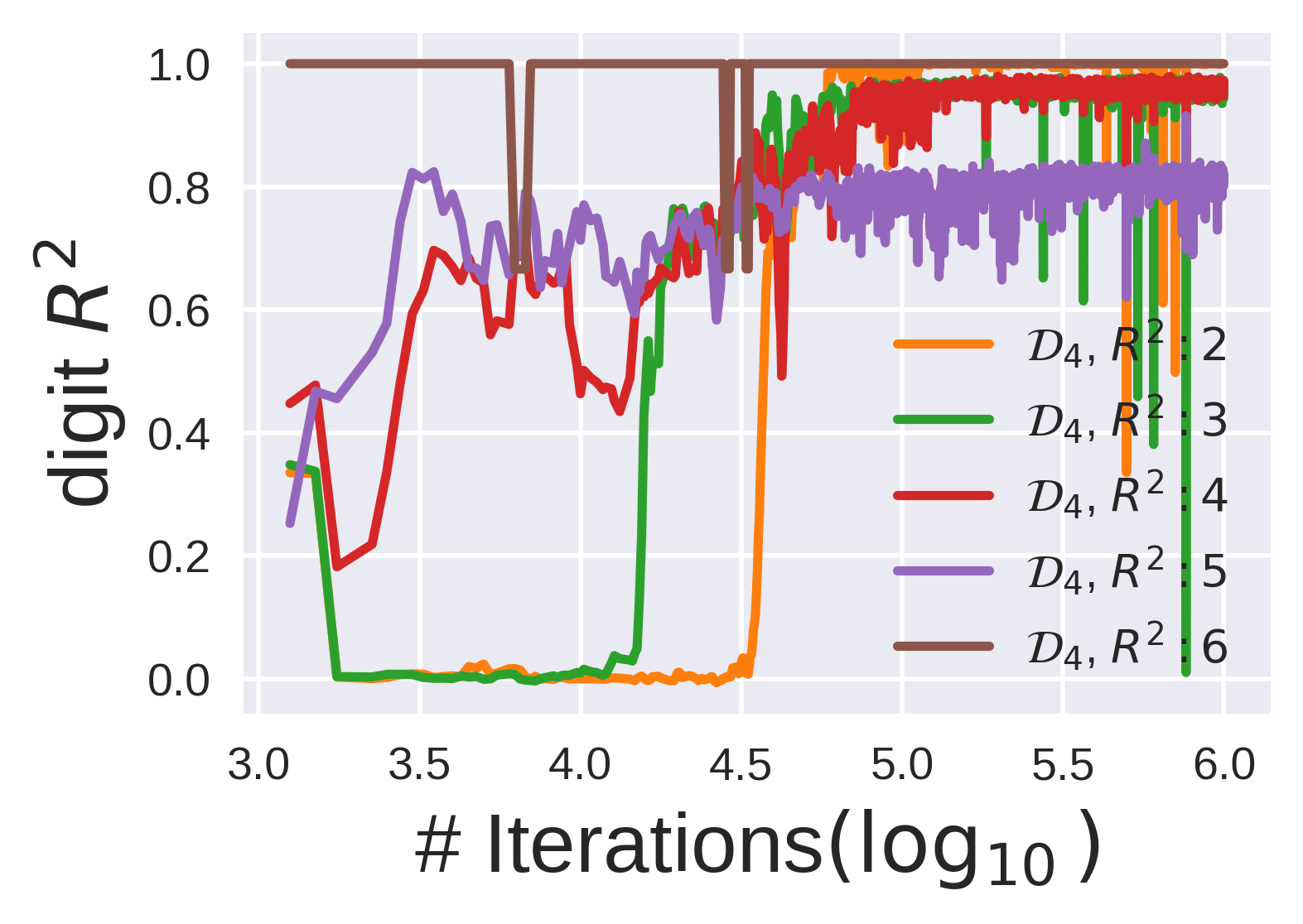}
  \caption*{$\mathcal{D}_{4}$}
\end{subfigure}
\hfill
\begin{subfigure}{0.3\textwidth}
  \includegraphics[width=\textwidth,height=0.75\textwidth]{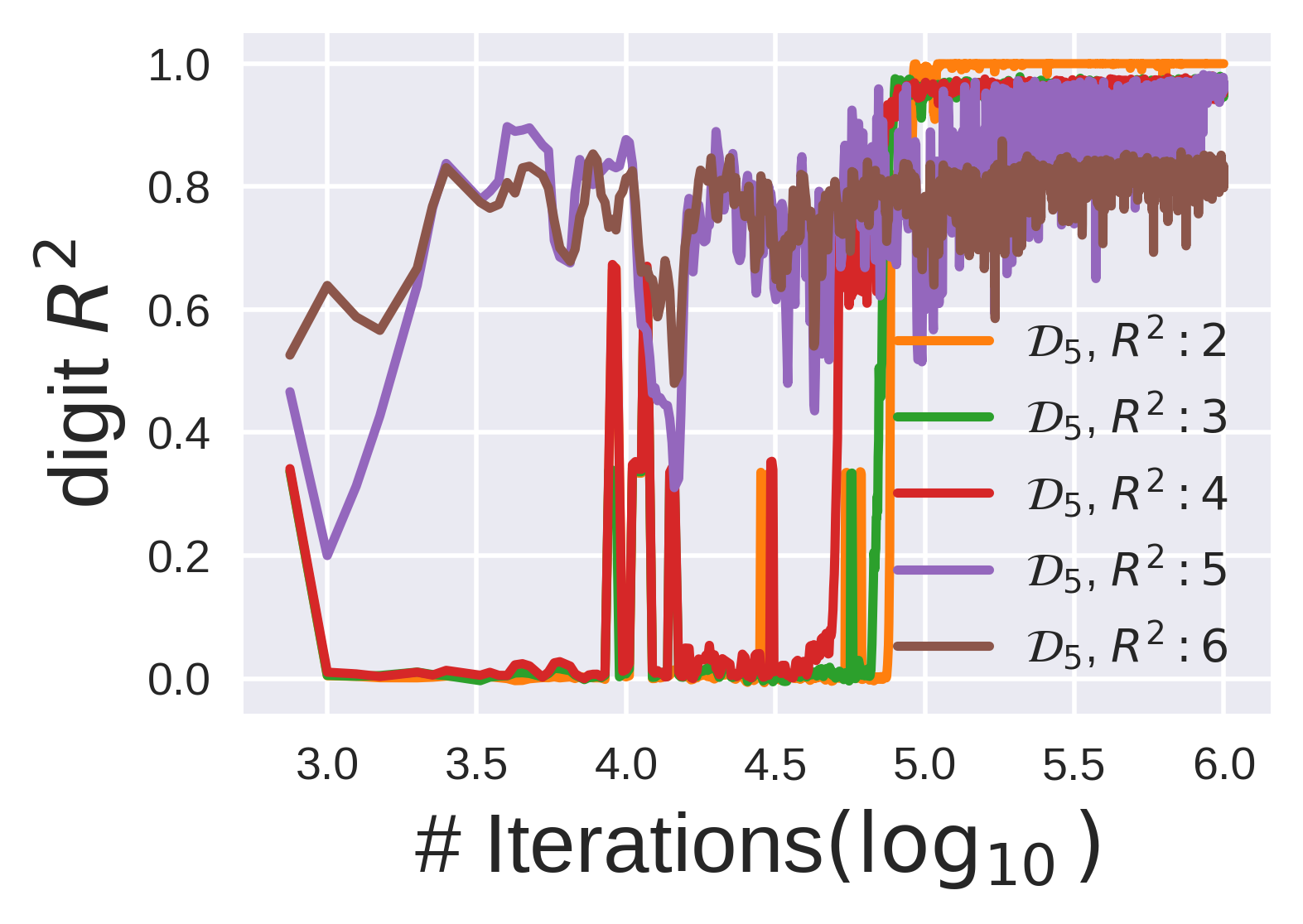}
  \caption*{$\mathcal{D}_{5}$}
\end{subfigure}
\hfill
\begin{subfigure}{0.3\textwidth}
  \includegraphics[width=\textwidth,height=0.75\textwidth]{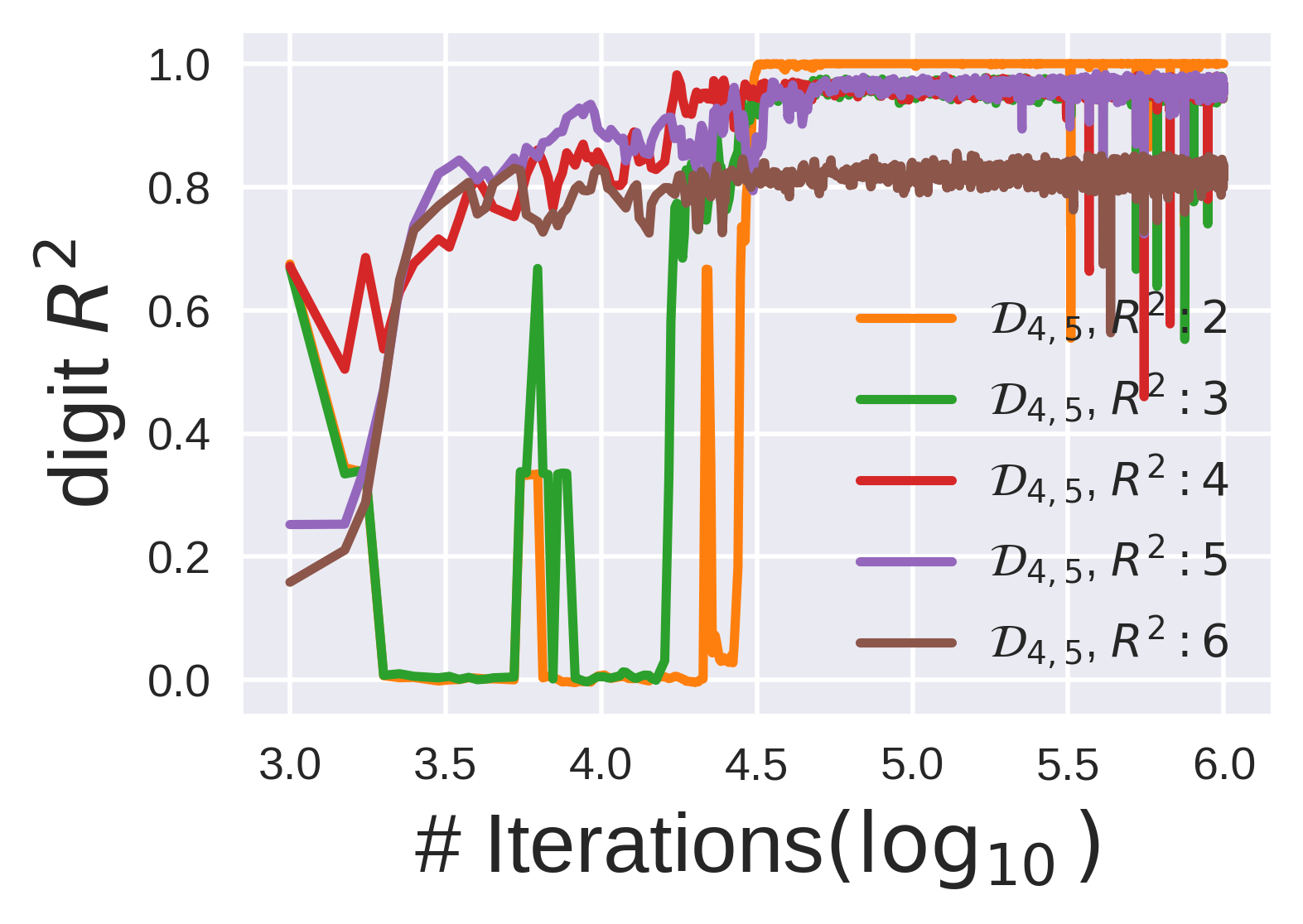}
  \caption*{$\mathcal{D}_{4,5}$}
\end{subfigure}
\end{minipage}
\caption{Learning Dynamics of Each Function $c_i = \zeta(a_i+b_i+c_{i-1}^\chi)$ for Addition}
\caption*{\textit{Note:} $\mathcal{D}_4$ is trained on two number addition task with at least one number to be a four digit number, $\mathcal{D}_5$ is trained on two number addition task with at least one number to be a 5 digit number, $\mathcal{D}_{4,5}$ is trained on the combined training dataset of $\mathcal{D}_4$ and $\mathcal{D}_5$. for $\mathcal{D}_{4}$ the model never quite learn to approximate $c_i = \zeta(a_i+b_i+c_{i-1}^\chi)$ yet only learn to output 0 as in its training data there's only $0$ at its 6th digit.}
\label{fig:addition_dynamic_r2}
\end{figure*}

The experiment results depicted in Figure~\ref{fig:addition_dynamic_r2} illustrate the learning dynamics of each function \( c_i \) during the training of Transformer models, using DecisionTreeRegressor to approximate these functions. The \( R^2 \) values, which measure how well the model's predictions fit the actual data, indicate that the models effectively learn lower-order digits with high accuracy, achieving \( R^2 \) values close to 1. However, higher-order digits present more challenges, resulting in lower and less stable \( R^2 \) values. Furthermore, at the early stages of training, the models first learn the higher-order digits (with higher \( R^2 \) values) and then proceed to learn the lower-order digits.

From Figure~\ref{fig:addition_dynamic_r2}, it is evident that the Transformer model trained on \(\mathcal{D}_4\) initially focuses on learning the digits at positions 4 and 5 before addressing positions lower than 4. Here, position 6 is trivial since it always equals zero. The Transformer model trained on \(\mathcal{D}_5\) first attempts to learn the digits at positions 5 and 6, then proceeds to positions lower than 5. The Transformer model trained on \(\mathcal{D}_{4,5}\) starts by learning the digits at positions 4, 5, and 6, and then moves to positions lower than 4. In our theoretical analysis, the most challenging parts are \( c_n \) and \( c_{n+1} \) when training the model with data in \(\mathcal{D}_n\), since these positions never encounter \( a_n = b_n = 0 \) and require OOD generalization. The models prioritize learning the hardest positions first, followed by the easier positions in these experiments.

\begin{figure*}[htbp]
\centering
\begin{subfigure}{0.35\textwidth}
  \includegraphics[width=\textwidth,height=0.75\textwidth]{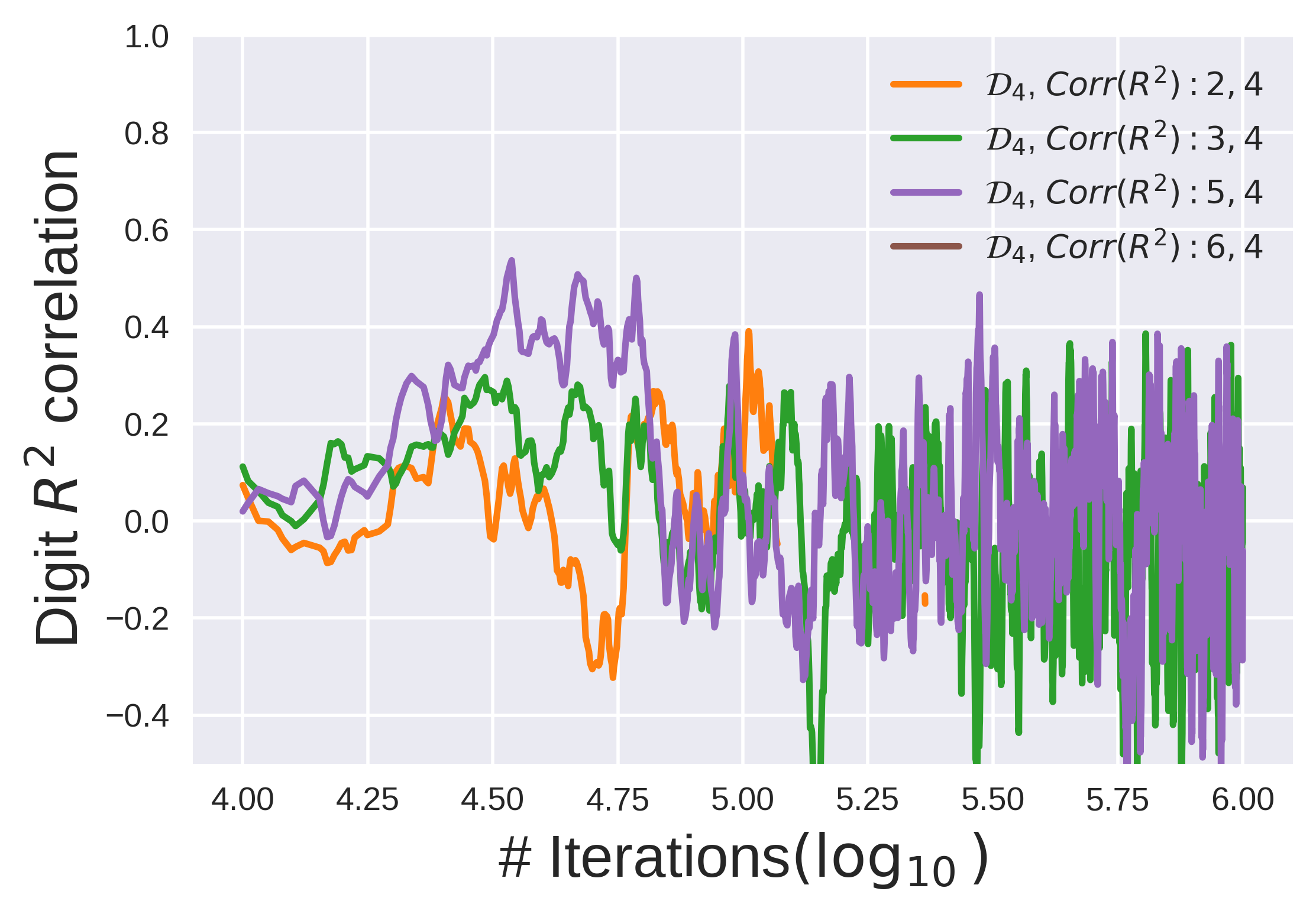}
  \caption*{$\mathcal{D}_{4}$, Ref digit 4}
\end{subfigure}
\hspace{1em} % Adjust the space as needed % \hfill
\begin{subfigure}{0.35\textwidth}
  \includegraphics[width=\textwidth,height=0.75\textwidth]{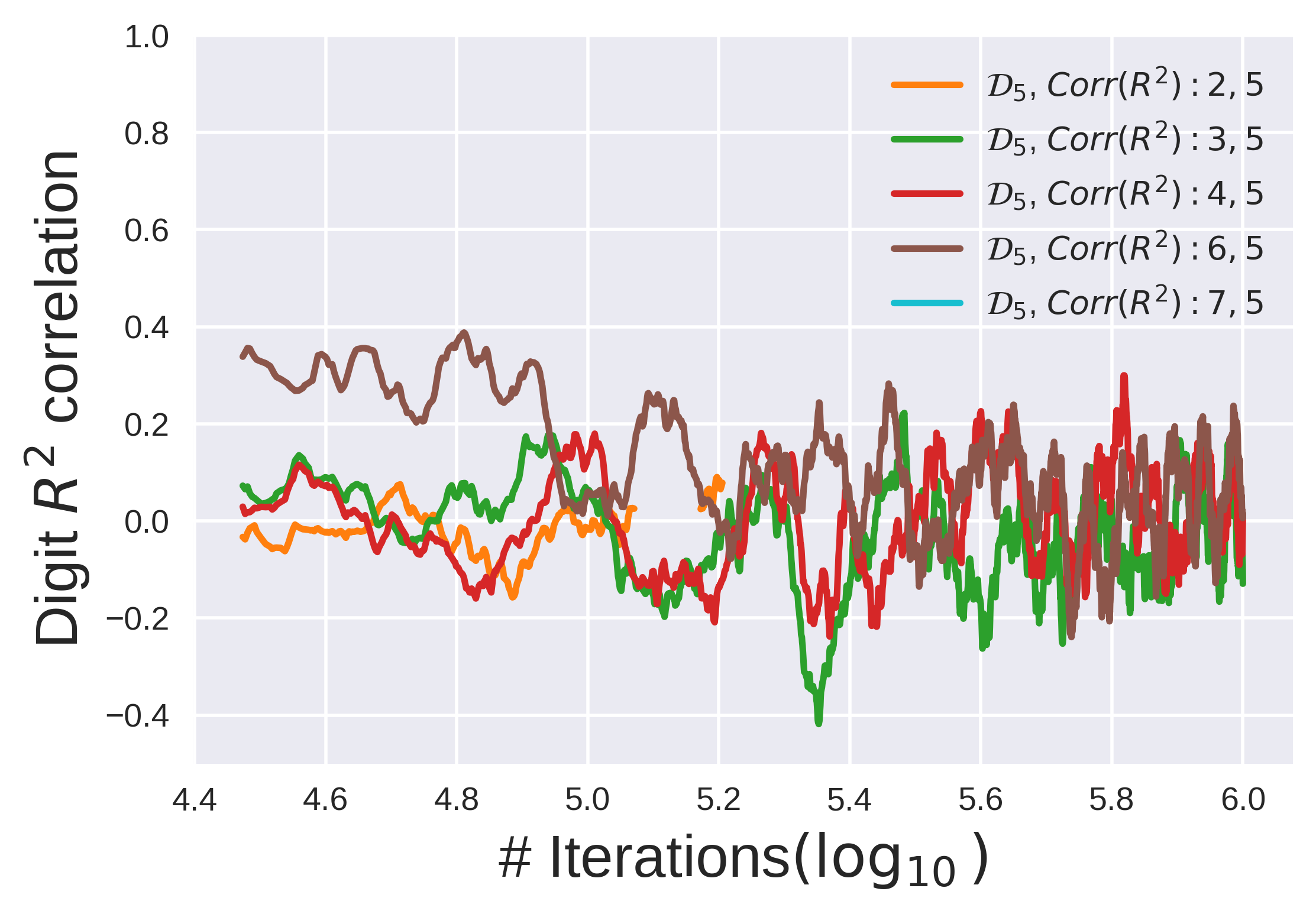}
  \caption*{$\mathcal{D}_{5}$, Ref digit 5}
\end{subfigure}
\\
\begin{subfigure}{0.35\textwidth}
  \includegraphics[width=\textwidth,height=0.75\textwidth]{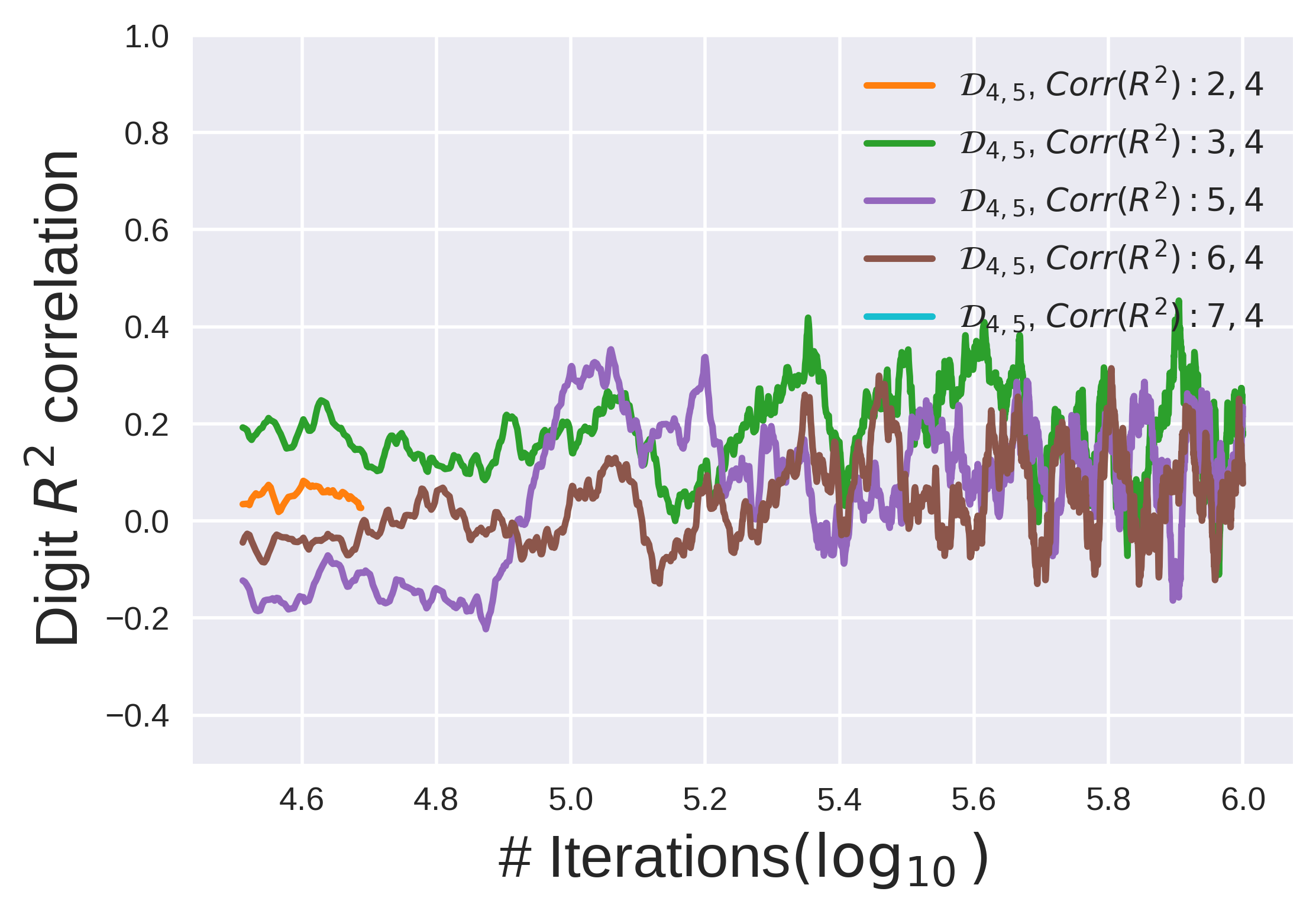}
  \caption*{$\mathcal{D}_{4,5}$, Ref digit 4}
\end{subfigure}
\hspace{1em} % Adjust the space as needed % \hfill
\begin{subfigure}{0.35\textwidth}
  \includegraphics[width=\textwidth,height=0.75\textwidth]{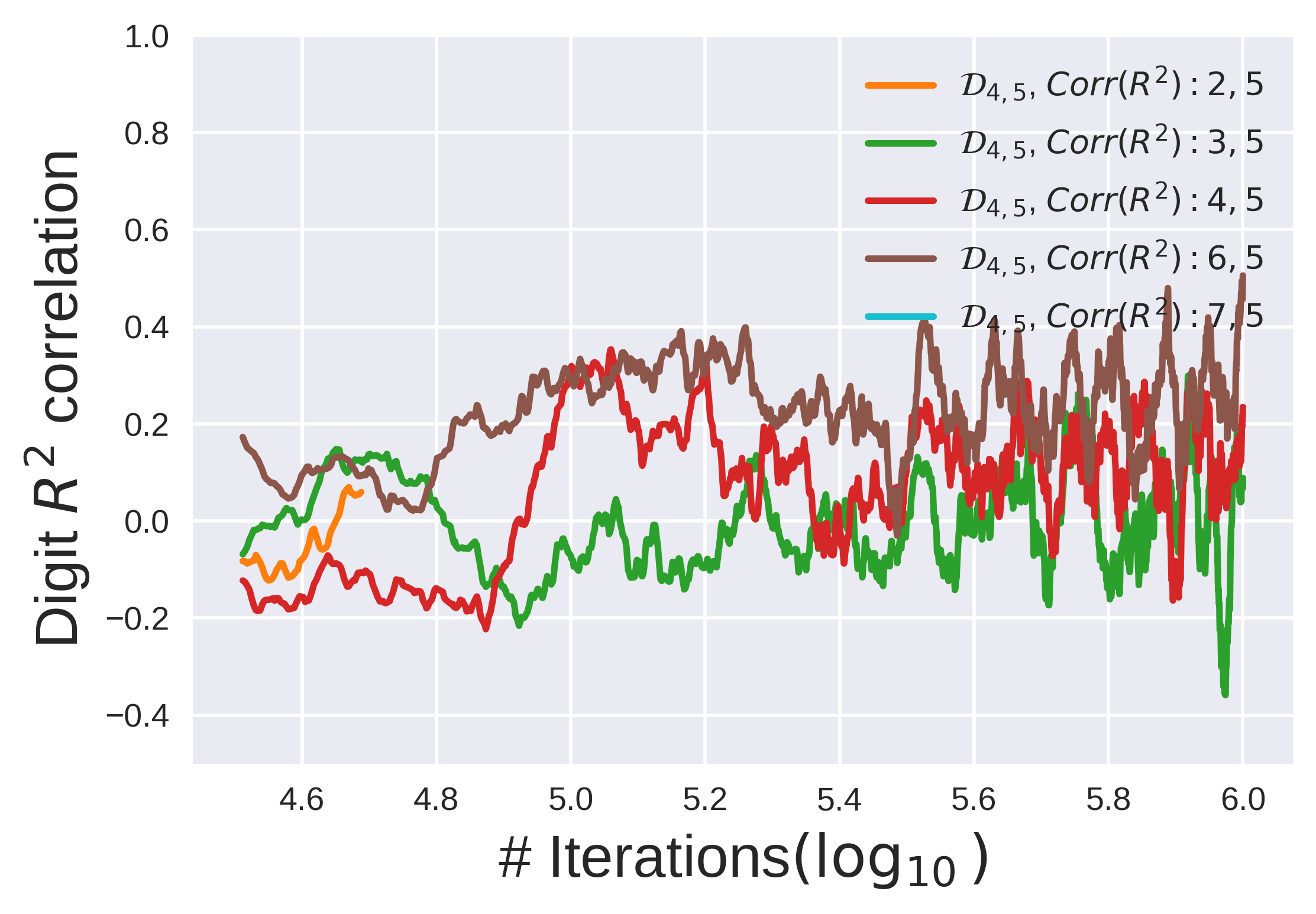}
  \caption*{$\mathcal{D}_{4,5}$, Ref digit 5}
\end{subfigure}
\caption{Correlation Between Digit Pairs of Learning $c_i$ and $c_j$ for Addition}
\caption*{\textit{Note:} $\mathcal{D}_4$ is trained on two number addition task with at least one number to be a four digit number, $\mathcal{D}_5$ is trained on two number addition task with at least one number to be a 5 digit number, $\mathcal{D}_{4,5}$ is trained on the combined training dataset of $\mathcal{D}_4$ and $\mathcal{D}_5$.}
\label{fig:addition_dynamic_r2_corr}
\end{figure*}

Another notable result from the experiments is that the correlation of \( R^2 \) values between different digit pairs is around zero (see Figure~\ref{fig:addition_dynamic_r2_corr} in this Appendix). %~\ref{appendix_results_addition}). 
This indicates that changes in the approximation for one position have little impact on other positions. This finding suggests that the Transformer model is flexible enough to handle different tokens independently, even though they share parameters.

% Addition: Learning dynamics r2_corr

\subsubsection{Learning Addition Under Relative Positional Embedding}

% Relative/Abacus Positional Embedding: Addition
\begin{figure}[htbp]
\centering
\includegraphics[width=0.45\textwidth]{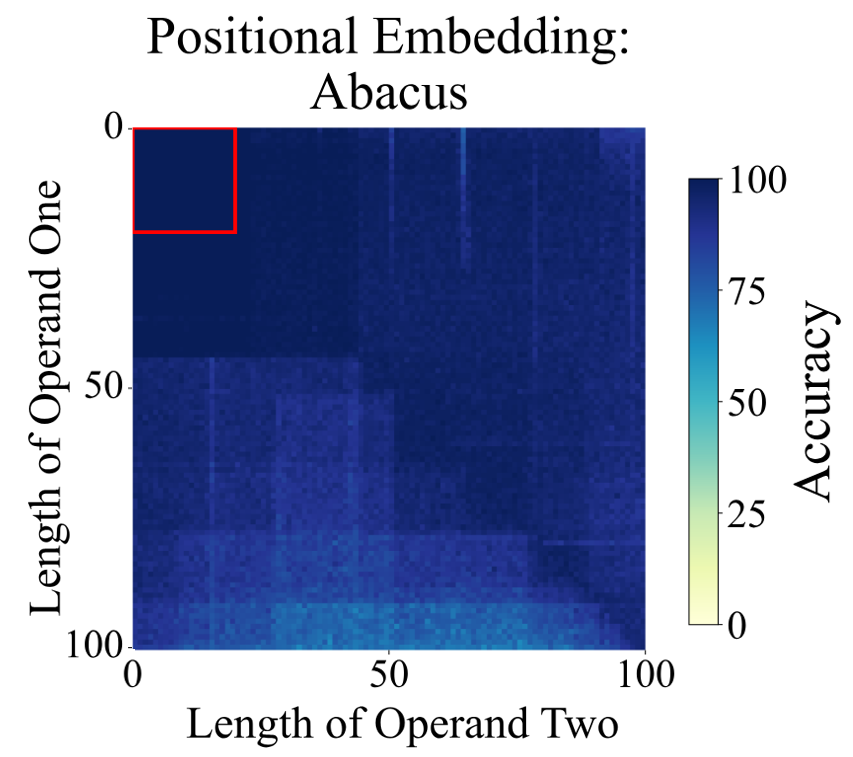}
\caption{Test Accuracy on Addition When Training Short and Testing Long using a 16-Layer Transformer (Decoder only) Model with Abacus Positional Embedding.} 
\caption*{\textit{Note:} The image is extracted from the work~\citet{mcleish2024transformers} and is a screenshot of their Figure 1. The interior of the red box represents the training data domain $\mathcal{D}_{\leq 20}$. Code to reproduce the result can be found on the GitHub: \url{https://github.com/mcleish7/arithmetic}. The obtained result constitutes empirical evidence that validates our Theorem~\ref{thm:addition-rpe}. The result is very clear. We will not repeat the same procedures. Use this as a reference in the present context.
}
\label{fig:addition-abacus}
\end{figure}

\newpage
\subsection{Further Results on Modular Addition}\label{appendix_results_modular_addition}

% Modular Addition
\begin{figure}[htbp]
\centering
\begin{minipage}{0.85\textwidth}
\centering
\begin{subfigure}{0.45\textwidth}
  \includegraphics[width=\textwidth,height=0.75\textwidth]{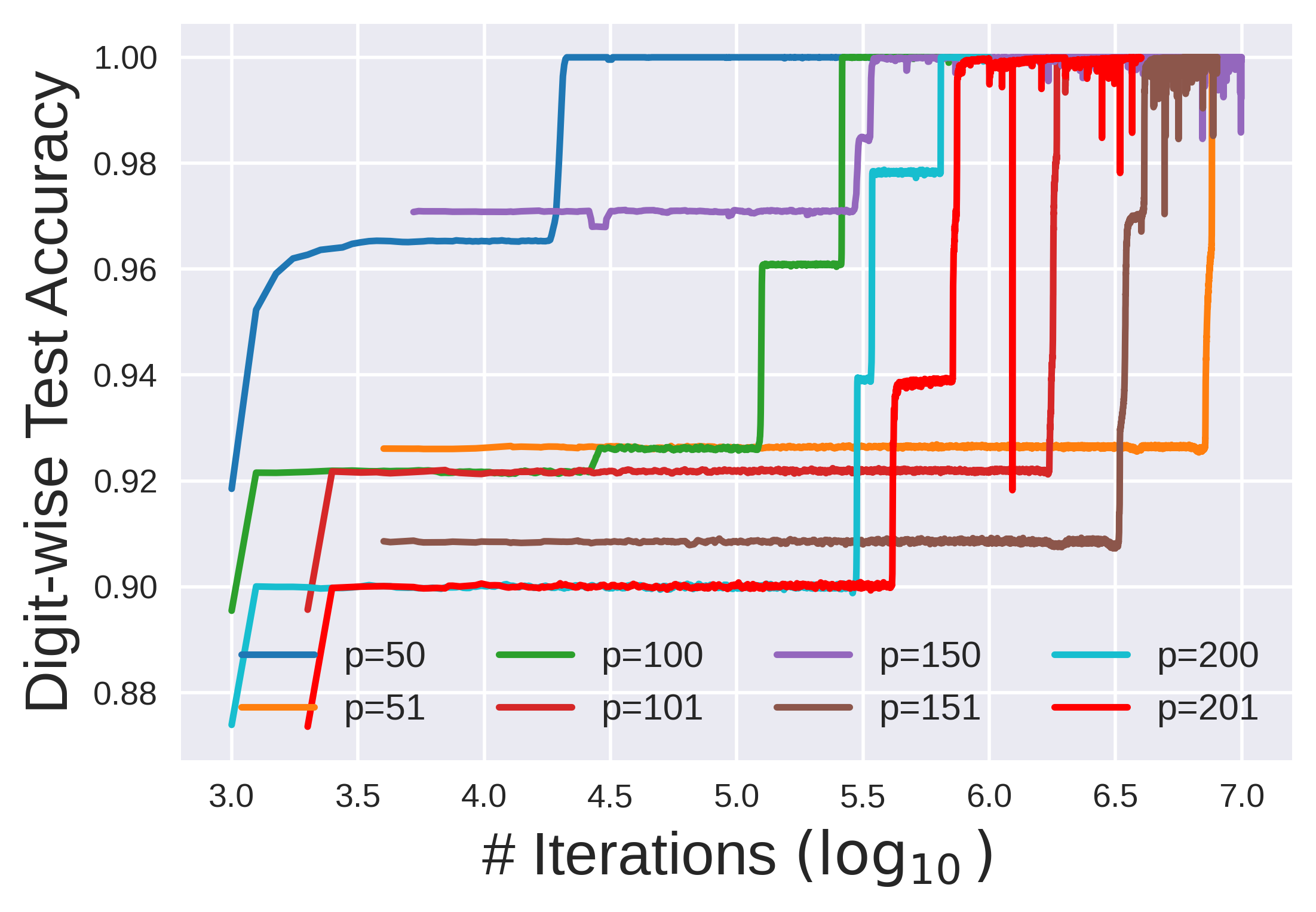}
  \caption*{Digit-wise Accuracy}
\end{subfigure}
\hfill
\begin{subfigure}{0.45\textwidth}
  \includegraphics[width=\textwidth,height=0.75\textwidth]{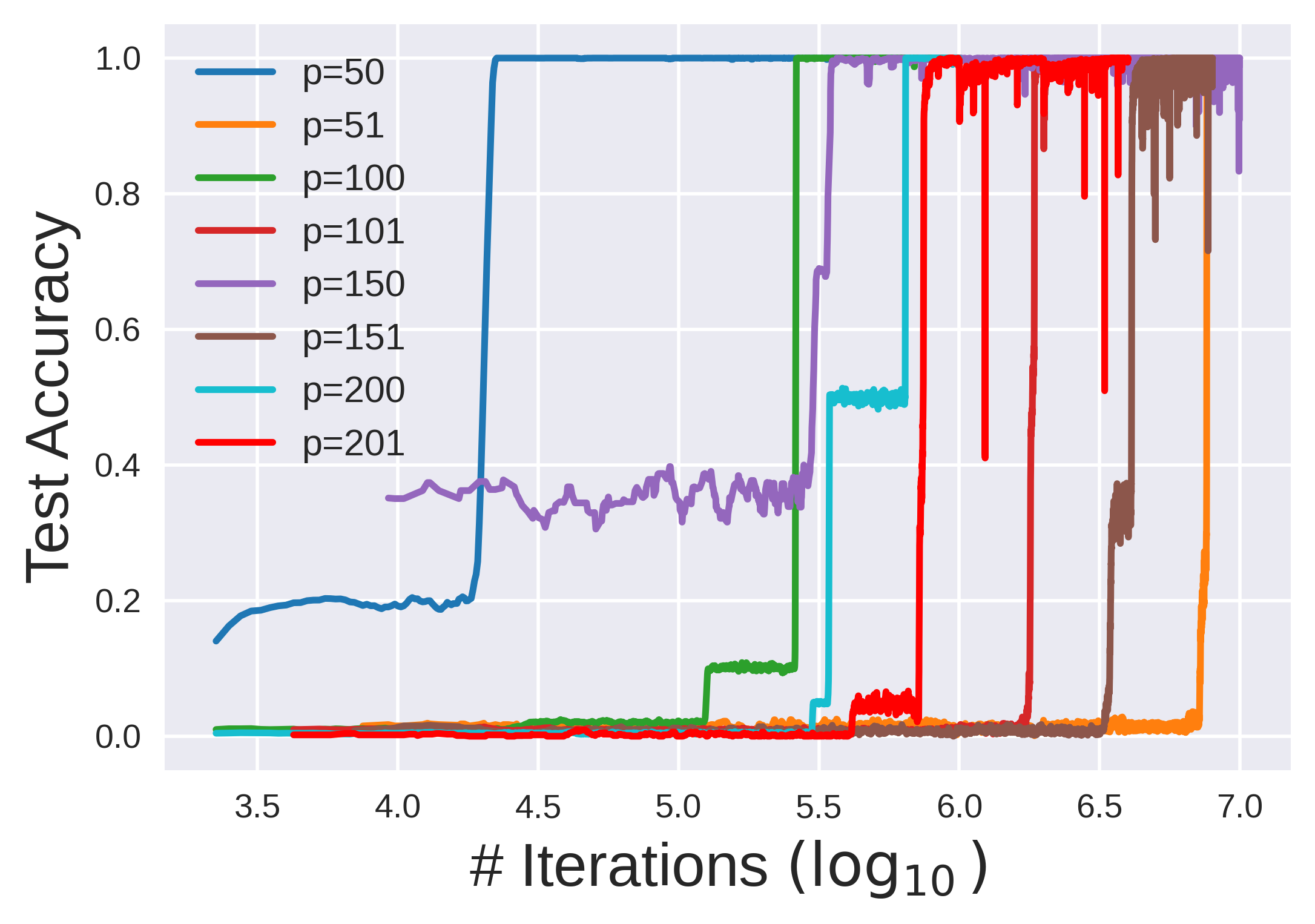}
  \caption*{In Sample Accuracy}
\end{subfigure}
\end{minipage}
\caption{Digit-wise In-Distribution Test Accuracy \&  Total Accuracy for Modular addition}
\caption*{\textit{Note:} These results correspond to modular addition tasks with the modulus $p$ taking values in the set $\{50, 51, 100, 101, 150, 151, 200, 201\}$. Each model is trained using the MiniGPT model with a sample drawn from the domain $\mathcal{D}_4$ (except $p=150$, which is on $\mathcal{D}_{\leq 4}$).}
\end{figure}

% Modular Addition: Modular Accuracy
\begin{table*}[ht]
\centering
\begin{tabular}{lrrrrrrrrr}
\toprule
\multicolumn{1}{c}{} & \multicolumn{9}{c}{Test Accuracy (\%) w.r.t. the Modular Truth on the Domain $\widetilde{\mathcal{D}}_i$} \\
Modulus & 1 & 2 & 3 & 4 & 5 & 6 & 7 & 8 & 9\\
\midrule
$p=50$ & 100 & 100 & 100 & 100 & 99.3 & 92.0 & 93.1 & 95.2 & 91.4 \\
$p=51$ & 100 & 98.5 & 99.9 & 99.3 & 95.1 & 94.4 & 92.6 & 91.3 & 92.4 \\
$p=100$ & 100 & 100 & 100 & 100 & 100 & 100 & 100 & 100 & 100 \\
$p=101$ & 100 & 100 & 100 & 100 & 100 & 100 & 100 & 100 & 100 \\
$p=150$ & 100 & 100 & 100 & 100 & 100 & 100 & 100 & 99.8 & 99.7 \\
$p=151$ & 100 & 99.9 & 99.9 & 100 & 99.9 & 99.7 & 99.6 & 99.1 & 99.2 \\
$p=200$ & 100 & 100 & 100 & 100 & 99.8 & 98.9 & 93.7 & 94.1 & 93.5 \\
$p=201$ & 100 & 100 & 99.9 & 99.9 & 96.4 & 96.6 & 95.7 & 90.4 & 91.2 \\
\bottomrule
\end{tabular}
\caption{Modular Addition: Test Accuracy w.r.t. the Modular Truth $\hat{f}^p(a,b)=\overline{\overline{a}^{10^n}+\overline{b}^{10^n}}^p$ on the Domain $\widetilde{\mathcal{D}}_i$ for $i=1,2\cdots,9$. % The models and test methods are as indicated in the above table.
}
\caption*{\textit{Note:} All the Transformer models in above experiments are instances of MiniGPT, which have been trained on a random sample drawn from $\mathcal{D}_4$ (except $p=150$). The accuracy is tested on 10,000 random test samples (when \( n > 2 \)), otherwise on the entire dataset. The outputs of models are generated using maximum probability sampling.}
\label{table:modular_addition_modular_acc}
\end{table*}

\newpage
\subsection{Further Results on Multiplication}\label{appendix_results_multiplication}

% Multiplication
\begin{figure}[htbp]
\centering
\begin{minipage}{0.85\textwidth}
\centering
\begin{subfigure}{0.45\textwidth}
  \includegraphics[width=\textwidth,height=0.75\textwidth]{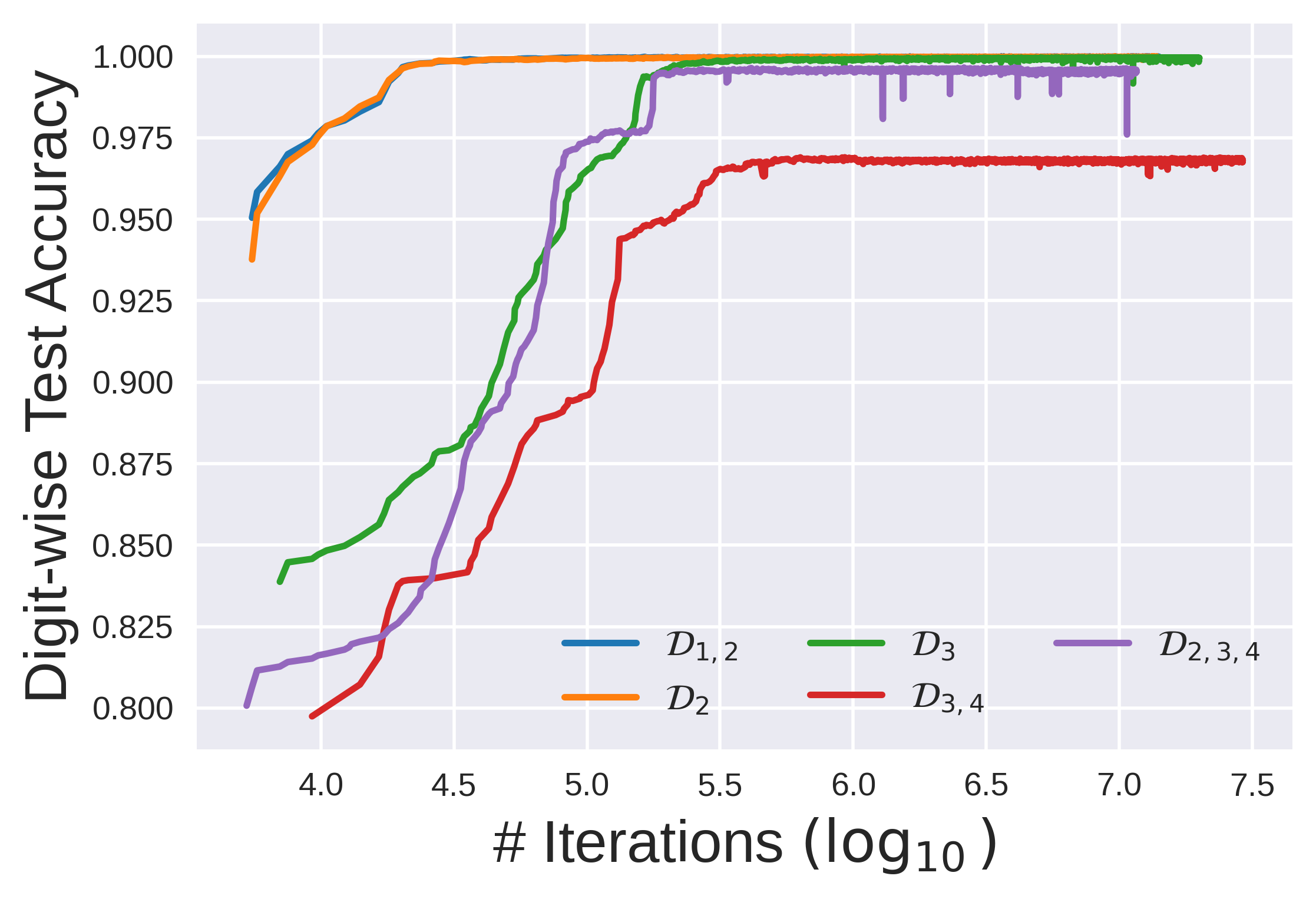}
  \caption*{Digit-wise Accuracy}
\end{subfigure}
\hfill
\begin{subfigure}{0.45\textwidth}
  \includegraphics[width=\textwidth,height=0.75\textwidth]{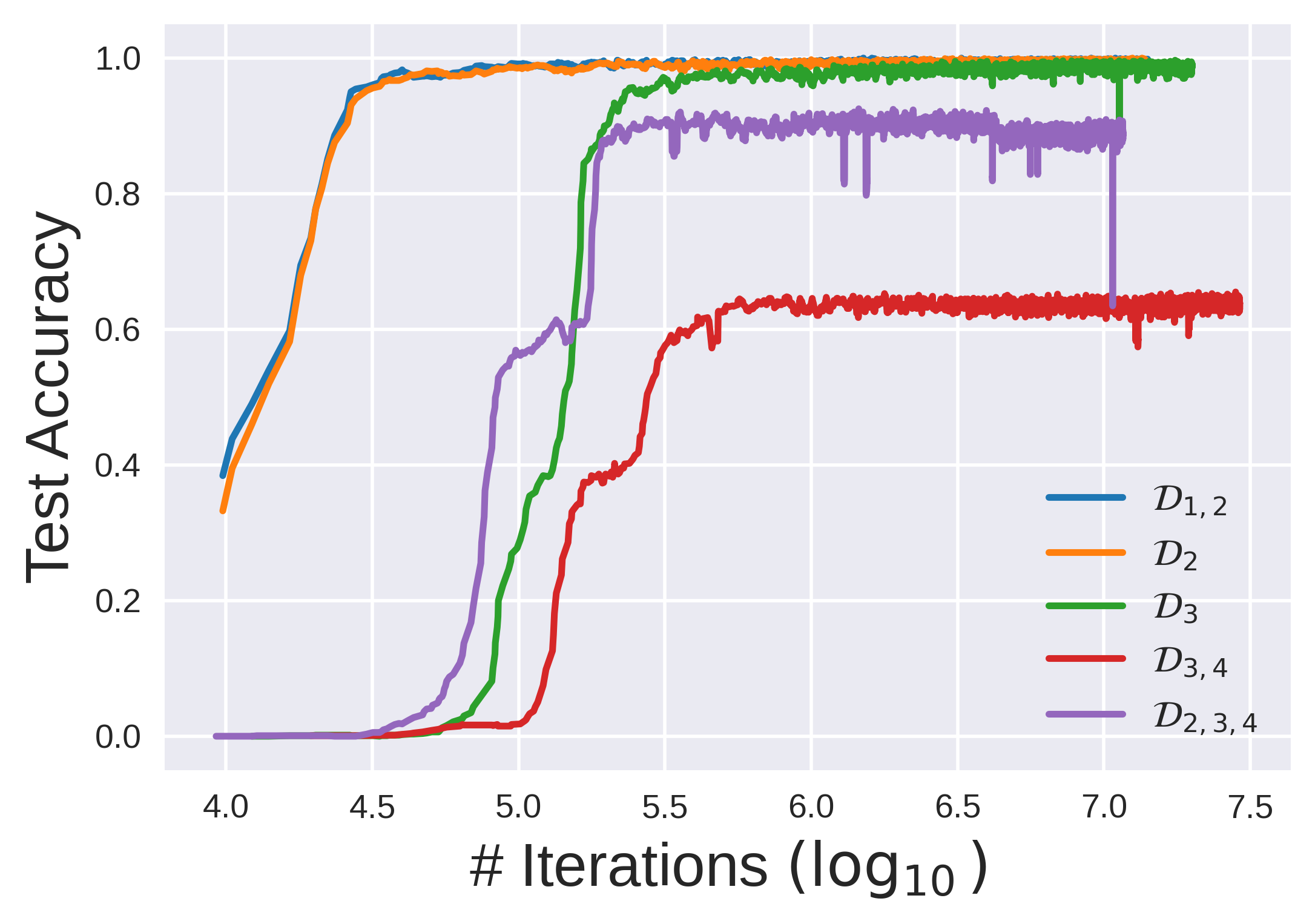}
  \caption*{In Sample Accuracy}
\end{subfigure}
\end{minipage}
\caption{Digit-wise In-Distribution Test Accuracy \&  Total Accuracy for Multiplication}
\caption*{\textit{Note:} These results correspond to multiplication tasks. The models trained on $\mathcal{D}_{1,2}$ and $\mathcal{D}_{2}$ are instances of MicroGPT, while others are of MiniGPT.}
\end{figure}

% Standard Multiplication: Ground Truth Accuracy
\begin{table*}[ht]
\centering
\begin{tabular}{lrrrrrrrrr}
\toprule
\multicolumn{1}{c}{} & \multicolumn{9}{c}{Test Accuracy (\%) w.r.t. the Ground Truth on $\mathcal{D}_i$} \\
Training Data & 1 & 2 & 3 & 4 & 5 & 6 & 7 & 8 & 9\\
\midrule
$\mathcal{D}_{1,2}$ & 100 & 100 & 0.1 & 0 & 0 & 0 & 0 & 0 & 0 \\
$\mathcal{D}_{2}$ & 80.0 & 99.4 & 0.1 & 0 & 0 & 0 & 0 & 0 & 0 \\
$\mathcal{D}_{3}$ & 100 & 96.4 & 99.0 & 0 & 0 & 0 & 0 & 0 & 0 \\
$\mathcal{D}_{2,3,4}$ & 100 & 100 & 98.9 & 80.5 & 0 & 0 & 0 & 0 & 0 \\
\bottomrule
\end{tabular}
\caption{Standard Multiplication: Test Accuracy w.r.t. the Ground Truth $f(a,b)=a\cdot b$ on the Domain $\mathcal{D}_i$ for $i=1,2\cdots,9$. The models trained on $\mathcal{D}_{1,2}$ and $\mathcal{D}_{2}$ are instances of MicroGPT, while others are of MiniGPT. The accuracy is tested on 10,000 random test samples (when \( n > 2 \)), otherwise on the entire dataset. The outputs of models are generated using maximum probability sampling.}
\label{table:multiplication_true_acc}
\end{table*}

% Standard Multiplication: Modular Truth Accuracy
\begin{table*}[ht]
\centering
\begin{tabular}{lrrrrrrrrr}
\toprule
\multicolumn{1}{c}{} & \multicolumn{9}{c}{Test Accuracy (\%) w.r.t. the Modular Truth on $\mathcal{D}_i$} \\
Training Data & 1 & 2 & 3 & 4 & 5 & 6 & 7 & 8 & 9\\
\midrule
$\mathcal{D}_{1,2}$ & 100 & 99.9 & 93.0 & 90.1 & 86.0 & 82.6 & 80.6 & 78.2 & 77.7 \\
$\mathcal{D}_{2}$ & 85.0 & 99.4 & 98.1 & 96.7 & 89.0 & 88.9 & 88.4 & 89.8 & 88.7 \\
$\mathcal{D}_{3}$ & 100 & 96.2 & 98.8 & 98.9 & 99.0 & 97.9 & 97.9 & 97.2 & 97.1 \\
% $\mathcal{D}_{3,4}$ & 85.0 & 10.3 & 96.9 & 1.4 & 0.9 & 0.8 & 0.8 & 0.7 & 0.2 \\
$\mathcal{D}_{2,3,4}$ & 100 & 100 & 98.9 & 81.0 & 75.6 & 76.2 & 73.8 & 67.5 & 66.9 \\
\bottomrule
\end{tabular}
\caption{Standard Multiplication: Test Accuracy w.r.t. the Modular Truth $\hat{f}(a,b)=\overline{a}^{10^n}\cdot\overline{b}^{10^n}$ on the Domain $\mathcal{D}_i$ for $i=1,2\cdots,9$. The models and test methods are indicated as above.}
\label{table:multiplication_modular_acc}
\end{table*}

% Relative/Abacus Positional Embedding: Multiplication
\begin{figure}[htbp]
\centering
\includegraphics[width=0.45\textwidth]{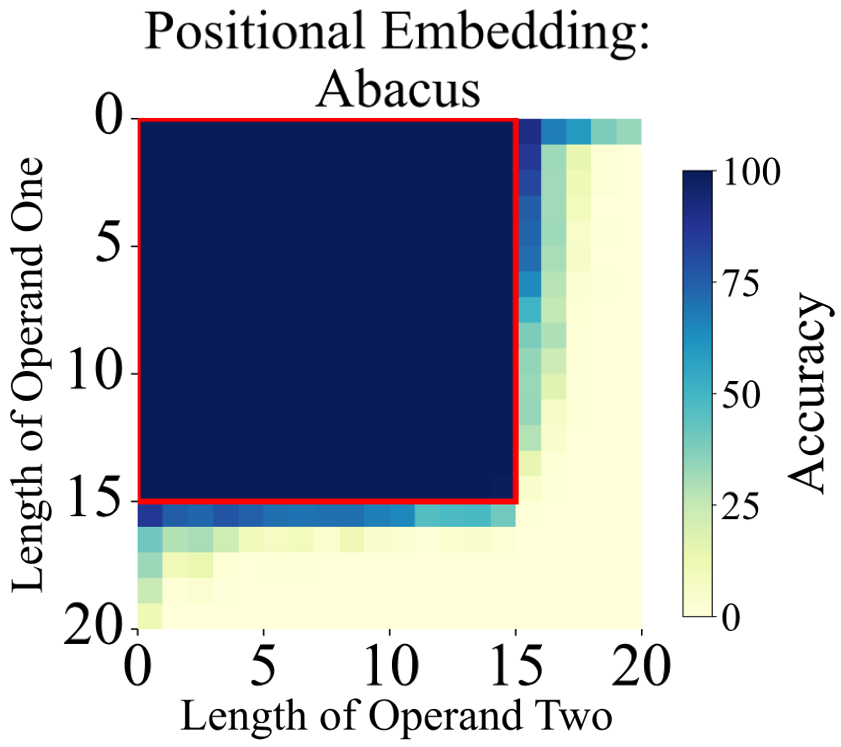}
\caption{Test Accuracy on Multiplication When Training Short and Testing Long using a Looped Transformer Models with Abacus Positional Embedding.}
\caption*{\textit{Note:} The image is extracted from the work~\citet{mcleish2024transformers} and is a screenshot of their Figure 5. The interior of the red box represents the training data domain $\mathcal{D}_{\leq 15}$.
}
\label{fig:multiplication-abacus}
\end{figure}

\newpage
\subsection{Further Results on Modular Multiplication}\label{appendix_results_modular_multiplication}

% Modular Multiplication
\begin{figure}[htbp]
\centering
\begin{minipage}{0.85\textwidth}
\centering
\begin{subfigure}{0.45\textwidth}
  \includegraphics[width=\textwidth,height=0.75\textwidth]{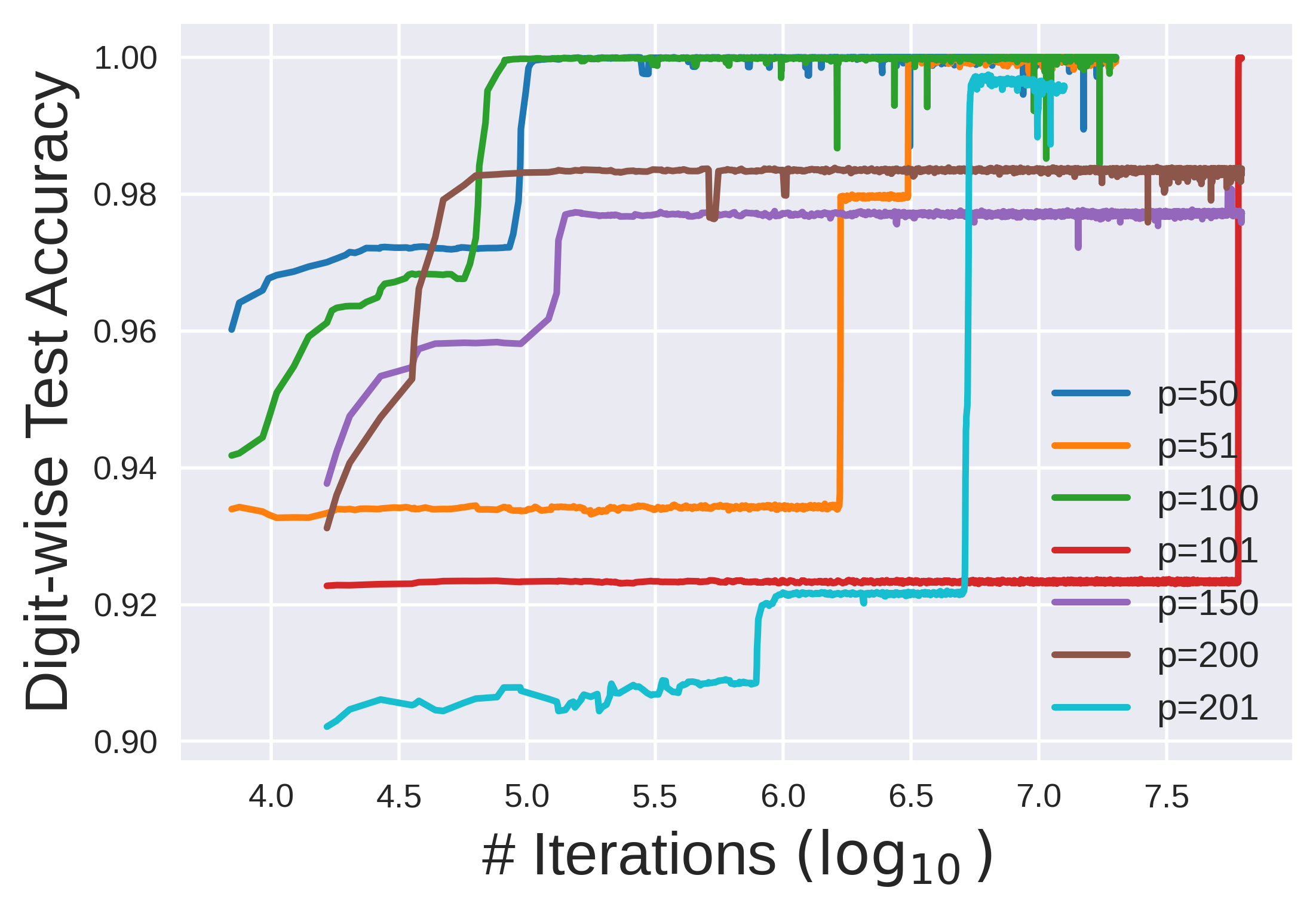}
  \caption*{Digit-wise Accuracy}
\end{subfigure}
\hfill
\begin{subfigure}{0.45\textwidth}
  \includegraphics[width=\textwidth,height=0.75\textwidth]{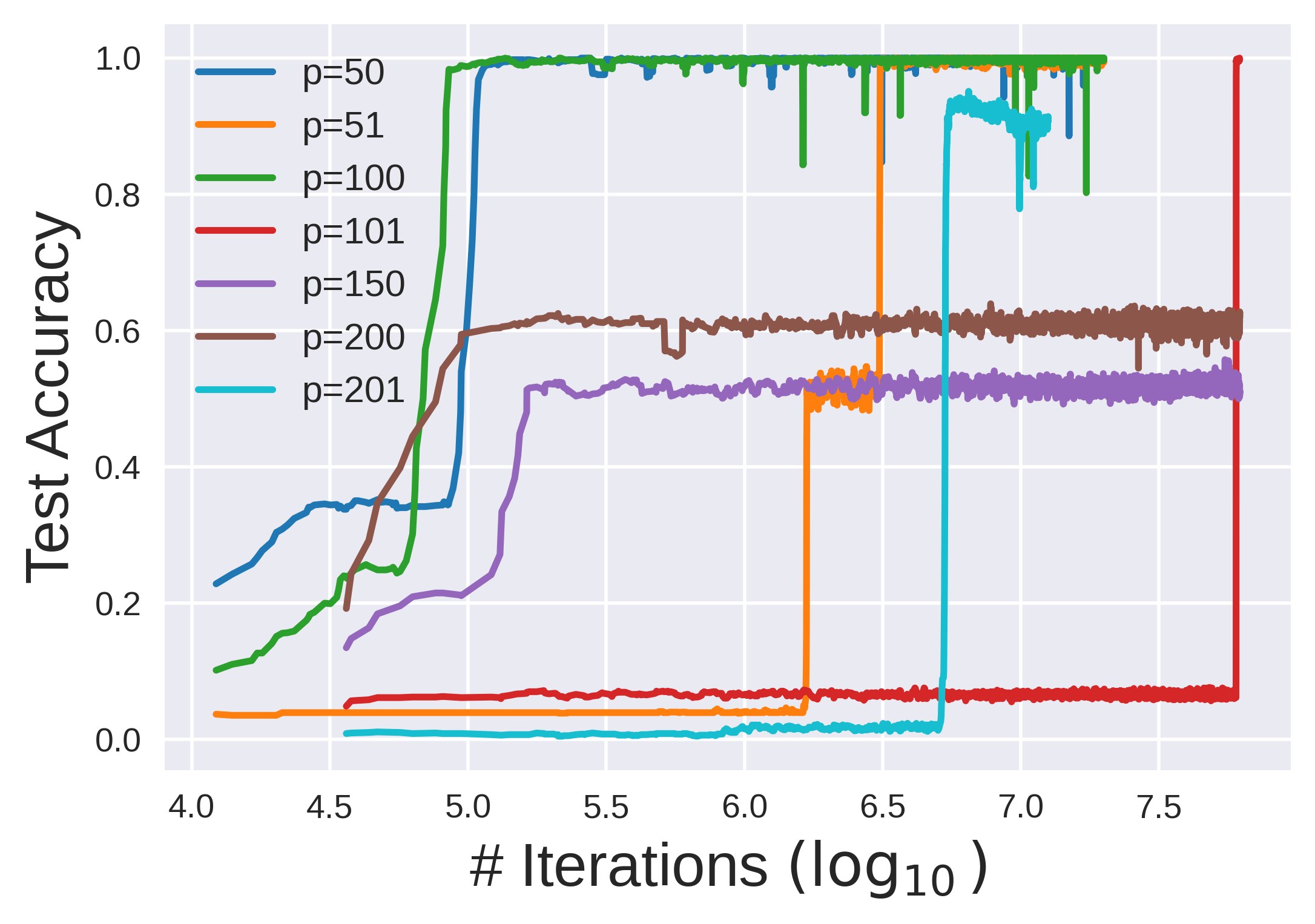}
  \caption*{In Sample Accuracy}
\end{subfigure}
\end{minipage}
\caption{Digit-wise In-Distribution Test Accuracy \&  Total Accuracy for Modular Multiplication}
\caption*{\textit{Note:} These results correspond to modular multiplication tasks. The models are instances of MiniGPT and trained on $\mathcal{D}_{3}$.}
\end{figure}

% Modular Multiplication: Ground Truth Accuracy
\begin{table*}[ht]
\centering
\begin{tabular}{lrrrrrrrrr|r}
\toprule
\multicolumn{1}{c}{} & \multicolumn{9}{c}{Test Accuracy (\%) w.r.t. the Ground Truth on the Domain $\widetilde{\mathcal{D}}_i$} & \multicolumn{1}{c}{Theor.} \\
Modulus & 1 & 2 & 3 & 4 & 5 & 6 & 7 & 8 & 9 & Acc.\\
\midrule
$p=50$ & 100 & 100 & 100 & \textcolor{red}{100} & \textcolor{red}{100} & \textcolor{red}{100} & \textcolor{red}{100} & \textcolor{red}{100} & \textcolor{red}{100} & 100 \\
$p=51$ & 100 & 100 & 99.7 & \textcolor{gray}{2.6} & \textcolor{gray}{2.5} & \textcolor{gray}{2.8} & \textcolor{gray}{2.4} & \textcolor{gray}{2.5} & \textcolor{gray}{3.2} & 2.4 \\
$p=100$ & 100 & 100 & 100 & \textcolor{red}{100} & \textcolor{red}{100} & \textcolor{red}{100} & \textcolor{red}{100} & \textcolor{red}{100} & \textcolor{red}{100} & 100 \\
$p=101$ & 100 & 100 & 100 & \textcolor{gray}{1.1} & \textcolor{gray}{1.0} & \textcolor{gray}{1.2} & \textcolor{gray}{0.9} & \textcolor{gray}{1.1} & \textcolor{gray}{1.0} & 1.0 \\
$p=150$ & 30.0 & 56.4 & 55.5 & \textcolor{blue}{46.9} & \textcolor{blue}{46.5} & \textcolor{blue}{46.3} & \textcolor{blue}{47.4} & \textcolor{blue}{46.9} & \textcolor{blue}{47.0} & 40.8 \\
$p=200$ & 100 & 63.3 & 61.8 & \textcolor{red}{62.1} & \textcolor{red}{62.6} & \textcolor{red}{62.9} & \textcolor{red}{62.4} & \textcolor{red}{61.7} & \textcolor{red}{62.6} & 100 \\
$p=201$ & 80.0 & 78.3 & 92.2 & \textcolor{gray}{0.7} & \textcolor{gray}{0.6} & \textcolor{gray}{0.5} & \textcolor{gray}{0.6} & \textcolor{gray}{0.6} & \textcolor{gray}{0.6} & 0.6 \\
\bottomrule
\end{tabular}
\caption{Modular Multiplication: Test Accuracy w.r.t. the Ground Truth $f^p(a,b)=\overline{a\cdot b}^p$ on $\widetilde{\mathcal{D}}_i$}
\caption*{\textit{Note:} All the Transformer models in above experiments are instances of MiniGPT, which have been trained on a random sample drawn from $\mathcal{D}_3$. The accuracy is tested on 10,000 random test samples (when \( i > 2 \)), otherwise on the entire dataset. The outputs of models are generated using maximum probability sampling. 
When $p=150$ and $p=200$, there is a significant difference between the experimental accuracy and the theoretical accuracy, which is \textit{due to the fact that these two models have not yet achieved sufficient training on the training set, or in other words, they are under-trained}. This can be observed from the test accuracy in columns 1, 2, and 3 of the table above.
}
\label{table:modular_multiplication_true_acc}
\end{table*}

% Modular Multiplication: Modular Truth Accuracy
\begin{table*}[ht]
\centering
\begin{tabular}{lrrrrrrrrr}
\toprule
\multicolumn{1}{c}{} & \multicolumn{9}{c}{Test Accuracy (\%) w.r.t. the Modular Truth on $\widetilde{\mathcal{D}}_i$} \\
Training Data & 1 & 2 & 3 & 4 & 5 & 6 & 7 & 8 & 9\\
\midrule
$p=50$ & 100 & 100 & 100 & 100 & 100 & 100 & 100 & 100 & 100 \\
$p=51$ & 100 & 100 & 99.7 & 99.8 & 98.4 & 84.4 & 81.9 & 68.6 & 57.2 \\
$p=100$ & 100 & 100 & 100 & 100 & 100 & 100 & 100 & 100 & 100 \\
$p=101$ & 100 & 100 & 100 & 86.6 & 73.6 & 71.7 & 68.1 & 65.7 & 54.5 \\
$p=150$ & 42.0 & 55.7 & 56.0 & 51.0 & 51.2 & 50.0 & 50.0 & 50.3 & 50.1 \\
$p=200$ & 100 & 62.6 & 62.2 & 62.7 & 62.3 & 62.4 & 62.7 & 62.3 & 61.9 \\
$p=201$ & 71.0 & 79.5 & 92.1 & 90.9 & 90.7 & 90.5 & 88.7 & 87.9 & 85.0 \\
\bottomrule
\end{tabular}
\caption{Modular Multiplication: Test Accuracy w.r.t. the Modular Truth $\hat{f}^p(a,b)=\overline{\overline{a}^{10^n}\cdot \overline{b}^{10^n}}^p$ on the Domain $\widetilde{\mathcal{D}}_i$ for $i=1,2\cdots,9$. The models and test methods are as indicated in the above table.}
\label{table:modular_multiplication_modular_acc}
\end{table*}

\end{document}